\documentclass[journal]{IEEEtran}
\usepackage{amsmath,amsfonts}
\usepackage{algorithmic}
\usepackage{array}
\usepackage[caption=false,font=normalsize,labelfont=sf,textfont=sf]{subfig}
\usepackage{textcomp}
\usepackage{stfloats}
\usepackage{url}
\usepackage{hyperref}
\usepackage{verbatim}
\usepackage{graphicx}
\usepackage{cite}
\usepackage{multirow}
\usepackage{makecell}
\usepackage{booktabs}
\usepackage{tabularx}
\usepackage{amssymb}
\usepackage{ulem}
\usepackage[table]{xcolor}
\usepackage[ruled,linesnumbered]{algorithm2e}
\def\BibTeX{{\rm B\kern-.05em{\sc i\kern-.025em b}\kern-.08em
    T\kern-.1667em\lower.7ex\hbox{E}\kern-.125emX}}
\usepackage{balance}

\begin{document}
\normalem
\title{MEGO: Learning Mixture-of-Experts for General-Purpose Binary Optimization}

\author{Shengcai Liu,~\IEEEmembership{Member,~IEEE}, Zhiyuan~Wang,~\IEEEmembership{Student Member,~IEEE},
\\
Yew-Soon Ong,~\IEEEmembership{Fellow,~IEEE},
Xin Yao,~\IEEEmembership{Fellow,~IEEE},
\and Ke Tang,~\IEEEmembership{Fellow,~IEEE}
\thanks{Shengcai Liu, Zhiyuan Wang, and Ke Tang are with the Guangdong Provincial Key Laboratory of Brain-inspired Intelligent Computation, Department of Computer Science and Engineering, Southern University of Science and Technology, Shenzhen 518055, China (email: liusc3@sustech.edu.cn; wangzy2020@mail.sustech.edu.cn; tangk3@sustech.edu.cn)}
\thanks{Yew-Soon Ong is with the College of Computing \& Data Science, Nanyang Technological University, Singapore 639798 (email: asysong@ntu.edu.sg)}
\thanks{Xin Yao is with the School of Data Science, Lingnan University, Hong Kong, China (email: xinyao@ln.edu.hk)}
\thanks{Corresponding author: Ke Tang.}
}

\markboth{Journal of \LaTeX\ Class Files,~Vol.~14, No.~8, August~2021}%
{Shell \MakeLowercase{\textit{et al.}}: A Sample Article Using IEEEtran.cls for IEEE Journals}


\maketitle

\begin{abstract}
Discrete optimization is ubiquitous in science and engineering.
The vast array of existing discrete optimization problems, coupled with the continuous emergence of new ones, necessitates off-the-shelf optimizers capable of generating high-quality solutions for a large variety of optimization problems.
This article introduces MEGO, a novel general-purpose neural optimizer for binary optimization {\color{black}under the black-box setting, intended for broad applicability across diverse binary optimization problem classes with minimal problem-specific customization.}
MEGO comprises a mixture-of-experts trained without domain knowledge.
When presented with a new problem instance to solve, it employs a routing policy to dynamically activate the most relevant expert models to generate high-quality solutions.
The strong generalization capability of MEGO is demonstrated on six problem classes from different disciplines, including classic problems and real-world applications.
Trained solely on classic problems, MEGO effectively generalizes to unseen and complex real-world problem classes, significantly outperforming widely-used
general-purpose optimizers in both solution quality and efficiency.
Furthermore, MEGO provides a computational approach for quantifying similarity between optimization problems and classifying them, which is fundamentally different from the conventional analysis-based problem classification.

\end{abstract}


\section{Introduction}
\label{sec:intro}

\IEEEPARstart{D}{iscrete} optimization, an important branch of applied mathematics and computer science, focuses on solving optimization problems with discrete decision variables.
This type of problems represents a foundational challenge that spans across scientific and engineering disciplines, such as in 
synthetic biology~\cite{naseri2020application},
urban planning~\cite{tang2025searching},
social network analysis~\cite{kempe2003maximizing,yu2024recognizing},
camera imaging~\cite{hruby2022learning, wang2022efficient},
and compiler configuration~\cite{jiang2021smartest}.
Traditionally, these problems are addressed through a problem-specific approach, where the inherent structures of the specific problem class are analyzed, and tailored optimizers are designed to exploit these characteristics.
For example, for the Influence Maximization Problem (IMP) on social network~\cite{kempe2003maximizing}, the submodularity and monotonicity of the problem have been leveraged to achieve theoretically optimal approximation ratios.
On the other hand, although such problem-specific approach can achieve excellent performance, it requires substantial domain expertise and human effort for each problem class~\cite{liu2023good}.
The significant time and resources required to develop each specialized optimizer create a critical bottleneck.
Consequently, this problem-specific approach struggles to keep pace with the vast and rapidly expanding landscape of optimization problems~\cite{pan2023surveybinary}, particularly for optimization problems where domain knowledge is not yet available or is difficult to acquire~\cite{rios2013derivative}.

\begin{figure*}[tbp]
    \centering
    \includegraphics[width=0.8\textwidth]{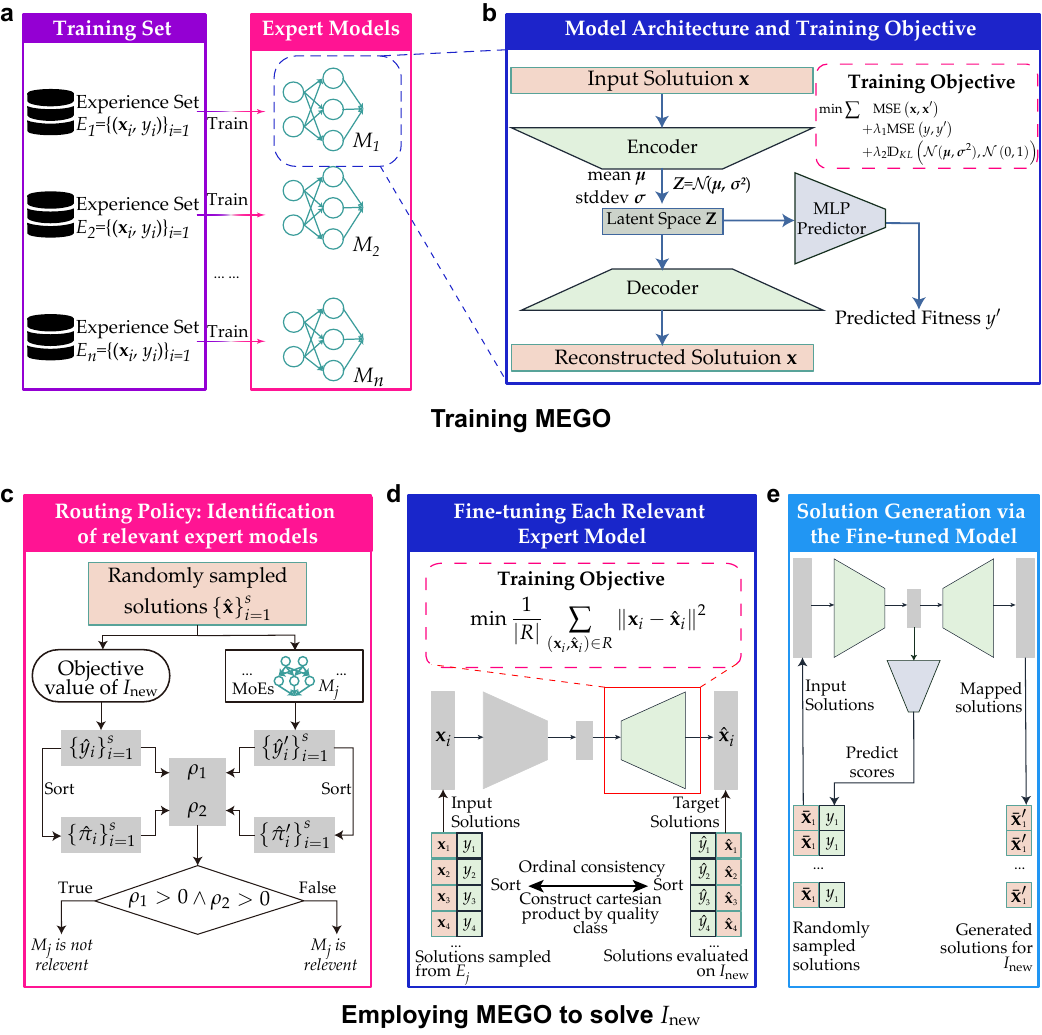}
    \caption{\textbf{An overview of MEGO}. \textbf{a-b}, the training process of MEGO, where the experience sets gained from solving training problem instances are abstracted into a MoEs; each expert model \(M_j\) consists of an encoder, a decoder, and a latent score predictor. \textbf{c-e}, the three steps of employing MEGO to solve a new problem instance \(I_{new}\): 1) the routing policy identifies relevant expert models based on the correlation coefficients between \(I_{new}\) and each \(M_j\); 2) each relevant \(M_j\) is fine-tuned to establish a transformation from the solution space of \(I_j\) to that of \(I_{new}\); 3) high-quality solutions to \(I_j\) are mapped into the solution space of \(I_{new}\).}
    \label{fig:mego_main}
\end{figure*}

Given the above challenge, there is a pressing need for \textit{high-performing general-purpose optimizers that can serve as off-the-shelf tools for a large variety of optimization problems}.~\footnote{We focus on problems that arise in practical contexts rather than completely arbitrary random binary functions. Even when treated as black-box problems, such practical problems often exhibit non-arbitrary regularities in their solution-quality landscapes, for example in variable interactions or the relative importance of variables~\cite{li2023machine,schrijver2003combinatorial,wang2016bayesian}.
	Such regularities may make the transfer of solving experience across instances, and potentially across problem classes, more plausible.
	{\color{black}Detailed discussions on the non-arbitrary regularities can be found in the supplementary material.}}
Specifically, such optimizers should deliver high-quality (if not optimal) solutions across diverse problem classes without requiring problem-specific adjustments.
Actually, the pursuit of general-purpose optimizers dates back several decades, gradually giving rise to optimizers including 
Evolutionary Algorithms (EAs)~\cite{holland1992genetic},
Simulated Annealing (SA)~\cite{kirkpatrick1983optimization},
Estimation of Distribution Algorithms (EDAs)~\cite{larranaga2001estimation},
and Bayesian optimization (BO)~\cite{jones1998efficient}.
However, the performance of these optimizers often falls short of satisfactory levels without problem-specific customization, such as meticulous parameter tuning~\cite{lopez2016irace} or the selection of specialized search operators and kernel~\cite{malkomes2018automating}.
This dependency on expert-driven, often time-consuming customization presents a significant barrier to their practical utility.
Consequently, a critical gap persists in the availability of high-performing, general-purpose optimizers that can be readily deployed.

This article aims to bridge the above gap in the realm of discrete optimization with binary decision variables (hereafter referred to as binary optimization).
Such problems are widespread in practice, e.g., appearing in machine learning~\cite{elsken2019neuralarch} and certain graph problems~\cite{lu2015competition}.
{\color{black}
	In this work, we study these problems under the black-box setting, where the optimizer accesses a problem instance only through function evaluations.
	This setting fits naturally our goal of minimizing problem-specific customization, since it avoids reliance on handcrafted structures, explicit formulations, or other domain knowledge.
	Moreover, the black-box setting is also practically relevant, since in many real-world problems~\cite{jiang2021smartest} such information is unavailable or difficult to exploit effectively.
Accordingly, our use of ``general-purpose'' refers to broad applicability across diverse binary problem classes with minimal problem-specific customization, rather than a universal performance guarantee for all binary optimization problems.}
Specifically, we develop a novel general-purpose neural optimizer for binary optimization under this black-box setting through the learning-to-optimize (L2O) paradigm~\cite{tang2024learn}.
The core principle of L2O involves training optimizers to effectively handle unseen problem instances by drawing upon experiences from solving training problem instances.
A prominent approach within this paradigm is to train neural networks (called neural optimizers) to solve optimization problems.
However, prior research in this area has predominantly relied on domain knowledge to train neural optimizers for particular problem classes~\cite{vinyals2015pointer,bello2016neural,bengio2021machine,liu2023good,wang2024asp,kortus2023towards}.
In contrast, this work introduces a domain-agnostic neural optimizer that is trained using only sampled solutions and their corresponding objective values.

The resulting neural optimizer, named \underline{M}ixture-of-\underline{E}xperts as \underline{G}eneral-Purpose \underline{O}ptimizers (MEGO), can be viewed as a foundation model for binary optimization.
Fig.~\ref{fig:mego_main} presents an overview of MEGO.
At its core, MEGO comprises a collection of distinct neural networks, with each network acting as an ``expert'' model.
The architecture is inspired by the classical concept of Mixture-of-Experts (MoEs)~\cite{shazeer2017outrageously}, which refers to a collaborative architecture that adaptively activates specific experts for a given input.
Unlike many contemporary MoEs frameworks~\cite{cai2025surveymoe} where the router is typically a module jointly and end-to-end trained with the experts within a unified neural network, MEGO is designed as an ensemble of neural networks (expert models) governed by a dedicated routing policy.
These expert models are trained on experiences (i.e., sampled solutions and their objective values) gathered from solving a set of training problem instances (Figs.1a-1b).
When applied to new, unseen problem instances (Figs. 1c-1e), the routing policy automatically activates relevant expert(s) to generate high-quality solutions.
The entire process, from training to application, operates without the need for domain knowledge or human intervention.
This domain-agnostic design allows MEGO to effectively address diverse problem classes.

The strong generalization capability of MEGO across different problem classes is demonstrated through extensive experiments, highlighting two significant findings.
First, to the best of our knowledge, MEGO represents the first neural optimizer 
trained on a specific set of binary optimization problem classes that successfully generalizes to entirely different and unseen binary optimization problem classes.
We validate this across six problem classes from diverse disciplines, including three classic problem classes and three challenging problem classes arising from real-world applications.
Experimental results {\color{black} in our testbed} show that MEGO, trained solely on the three classic problem classes, consistently outperforms widely-used general-purpose optimizers such as GA~\cite{holland1992genetic}, Hill Climbing with random restart (HC)~\cite{russell2016artificial}, and BO~\cite{lindauer2022smac3} across all six problem classes in both solution quality and efficiency.
Specifically, these competitors require at least 3.6 times the number of function evaluations (\#FEs) to match MEGO's solution quality, with MEGO's advantage holding across the tested problem dimensions.
Furthermore, when used as an initial solution generator, MEGO can significantly enhance the solution quality of existing general-purpose optimizers.
Indeed, on certain problem classes, this enhancement enables a general-purpose optimizer to surpass the performance of specialized optimizers.

Second, MEGO introduces a new perspective on the relationships between binary optimization problems through its vector-based representation of problem instances.
Such representation facilitates the measurement of similarity between problem instances, leading to an intriguing problem clustering (classification) that diverges from conventional, analysis-based classification.
This data-driven perspective may offer new insights into the landscape of optimization problems.

The contributions of this article are summarized below.
\begin{enumerate}
\item A novel general-purpose neural optimizer, MEGO, is proposed to address diverse binary optimization problem classes {\color{black}under the black-box setting}.
\item Extensive experiments demonstrate that MEGO effectively generalizes to unseen and complex real-world problem classes {\color{black}in our testbed}, significantly outperforming widely-adopted general-purpose optimizers in both solution quality and efficiency.
Moreover, as an initial solution generator, MEGO can also boost the performance of existing optimizers.
\item MEGO provides a computational approach to classify optimization problems based on the vector-based representations, offering a new, data-driven perspective on the relationships between different problem classes.
\end{enumerate}

The remainder of this article is organized as follows.
Section~\ref{sec:related_work} presents the preliminaries and reviews the related work. 
Section~\ref{sec:method} details the architecture of MEGO, along with its training and application procedures.
Experiments are presented in Section~\ref{sec:exp}.
Section~\ref{sec:conclusion} concludes the article with discussions.

\section{Preliminaries and Related Work}
\label{sec:related_work}

Binary optimization refers to a broad category of optimization problems where the decision variables are restricted to binary values, i.e., $\mathbf{x} \in \{0, 1\}^d$.
The goal is to find a binary vector $\mathbf{x}^*$ that minimizes or maximizes an objective function $F(\mathbf{x})$, potentially subject to a set of constraints.
These problems are ubiquitous in scientific and engineering domains, including graph problems~\cite{lu2015competition} and machine learning~\cite{elsken2019neuralarch,cai2018featureselection}.
It is important to distinguish between a problem class and a problem instance.
A problem class, such as the Knapsack Problem (KP)~\cite{martello1990knapsack}, shares a common mathematical structure, including its objective function and constraints.
In contrast, a problem instance is a specific realization of that class with concrete parameters, e.g., item weights for a KP instance.

For many well-structured problem classes, specialized optimizers are developed by exploiting domain knowledge and the underlying mathematical properties.
A classic example is the greedy algorithm for the IMP, which leverages the submodularity and monotonicity of the objective function to provide theoretical guarantees~\cite{kempe2003maximizing}.
It is conceivable that such specialized optimizers often yield superior performance on their target problems;
however, designing a specialized optimizer for each problem class is exceptionally costly and not scalable. 
This inherent limitation motivates the exploration of a general-purpose optimizer presented in this work.
{\color{black}Specifically, we consider general-purpose binary optimization under the black-box setting, where optimizers access each problem instance only through function evaluations and aim to work across diverse binary problem classes with minimal problem-specific customization.
Below we briefly review existing related approaches under this scope.}

\subsection{General-Purpose Optimizers for Binary Optimization}
Existing general-purpose optimizers can be broadly categorized into search heuristics and model-based methods.
Search heuristics directly explore the solution space through iterative search procedures.
A prominent example is SA, which mimics the physical annealing process by occasionally accepting inferior solutions to escape local optima.
Variations of SA often incorporate momentum~\cite{keikha2011improved} or employ multi-point parallel search~\cite{mahfoud1995parallelsimulated}.
Another key family is EAs, particularly Genetic Algorithms (GAs), which are inspired by natural selection and use operators like crossover and mutation to evolve a population of solutions.
Different GAs may alter the solution encoding~\cite{londe2025brkga} or use self-adaptive operators to fit specific problems~\cite{srinivasa2007selfadaga}.
In contrast, model-based methods construct a surrogate model of the objective function to intelligently guide the search.
BO is a popular model-based approach where different surrogate models have been explored, ranging from sparse Bayesian linear models~\cite{baptista2018bocs} to non-parametric Gaussian Processes (GPs)~\cite{oh2019combo}.
Some methods~\cite{lindauer2022smac3} also utilize tree-based models, such as random forests, as effective surrogates.
Estimation of Distribution Algorithms (EDAs)~\cite{pelikan2006hboa} represent another direction, building a probabilistic model of high-performing solutions and then sampling from it to generate new candidates.
However, there has been much evidence showing that to achieve satisfactory performance on a specific problem class, these general-purpose optimizers often require problem-specific customization.
This can involve meticulous parameter tuning~\cite{lopez2016irace}~\cite{huang20surveymetahuristics} or selecting specialized operators for search heuristics~\cite{li2014operatorselection}, and choosing an appropriate kernel for model-based methods~\cite{malkomes2018automating,malkomes2016bayeskernel}.

\subsection{Neural Optimizers}
In recent years, deep learning has emerged as a powerful paradigm for developing optimizers~\cite{bengio2021machine}.
These methods learn neural optimizers from data, aiming to generalize across different problem instances.
This research can be broadly viewed from two perspectives based on the problem structures they address.
One major line of research focuses on permutation-based problems, such as the Traveling Salesman Problem and Vehicle Routing Problem.
The seminal Pointer Network~\cite{vinyals2015pointer} laid the foundation by using a sequence-to-sequence architecture to construct solutions autoregressively.
This paradigm was later advanced by using reinforcement learning~\cite{bello2016neural} for policy training and leveraging the power of Transformer-based architectures~\cite{kool2019attention} to capture complex dependencies between decision steps.
Recent efforts have further improved performance by explicitly handling solution symmetries~\cite{kwon2020pomo} and using policy ensembles~\cite{gao2024towardsgeneralizable}.
Another direction utilizes Graph Neural Networks (GNNs) to handle optimization problems naturally defined on graphs.
A common approach involves using a GNN to learn node embeddings that encode structural information about the problem instance.
These embeddings can then be used to guide a constructive heuristic~\cite{khalil2017learningcombinatorial} or to predict the probability of each node or edge being part of the final optimal solution~\cite{joshi2019efficientgcntsp}, which is often refined with a subsequent search algorithm like beam search.
While powerful, these neural optimizers are typically specialized for problems with permutation or graph structure, limiting their applicability to general-purpose binary optimization.

\subsection{VAE-Based Optimization and Learning-based Solver Selection}
A closely related line of research uses VAEs or related generative models to learn continuous latent spaces for optimization over discrete or structured domains.
Representative examples include molecule design~\cite{gomez2018automatic}, neural architecture optimization~\cite{luo2018neural}, and routing~\cite{hottung2021latent}.
More recent studies have further combined such latent representations with Bayesian optimization or domain-specific design strategies~\cite{maus2022local,lee2023advancing,hu2023glso,moss2025cowboys}.
Despite their differences, these methods generally focus on optimization within a specific structured domain by searching in a learned latent space.
MEGO differs from this line of work in both scope and mechanism.
First, it targets domain-agnostic binary black-box optimization rather than optimization within a single structured application domain.
Second, it does not rely on a single global latent search space shared across problem instances. Instead, each expert is trained only from sampled binary solutions and their objective values for one training instance, without using domain-specific representations, handcrafted structures, or explicit problem formulations.
Third, the VAE in MEGO is not used to define a latent search space for within-domain optimization; rather, it serves as a transferable expert in a MoEs framework, where selected experts are adapted to a new problem instance through fine-tuning.

Another related direction is learning-based algorithm or solver selection~\cite{gao2025neural,HeS025,kostovska2023psaas,wu2025robustness}.
These studies aim to select, from a portfolio of complete algorithms or solvers, the one(s) most suitable for a given instance, often based on explicit instance features, hand-crafted landscape features, meta-representations, or problem-specific structured encodings.
For example, recent neural solver selection methods for combinatorial optimization~\cite{gao2025neural} employ graph encoders tailored to routing problems.
He et al.~\cite{HeS025} study algorithm selection in black-box continuous optimization using Exploratory Landscape Analysis (ELA) features and neural-network selection models.
In contrast, MEGO does not select among external full solvers, but activates internal expert models.
Moreover, it does not require explicit problem features, graph-structured encoders, or customized feature extractors.
Its routing policy is based only on a small number of sampled evaluations on the target instance and the alignment between actual and predicted scores.
Finally, unlike conventional solver selection methods, MEGO further adapts the selected experts through decoder fine-tuning, enabling instance-wise transfer of optimization experience beyond simple solver choice.
Therefore, the above-mentioned methods are not directly applicable to the feature-free black-box setting considered in this work.

\section{MoEs as General-Purpose Optimizers (MEGO)}
\label{sec:method}
This section details our proposed neural optimizer, MEGO.
We first introduce its architecture, and then describe how to train MEGO and apply it to solve new problem instances.
Finally, different modes in which MEGO can be utilized are also discussed.

\subsection{Model Architecture and Training Procedure}
We aim to learn a general-purpose neural optimizer for a wide range of binary optimization problem classes.
To train such an optimizer, a training set \(T\) is constructed by collecting experiences from solving $n$ training problem instances \(I_j\left(1\leq j\leq n\right)\) from different problem classes.
The experience for each \(I_j\) is a dataset \(E_j=\left\{\left(\mathbf{x}_i,y_i \right)\right\}_{i=1}\), where \(\mathbf{x}_i\in\left\{0,1\right\}^{d_j}\)(\(d_j\) is the problem dimension of \(I_j\)) is a sampled solution and \(y_i\in\mathbb{R}\) is its objective value.
Note that \(E_j\) does not contain any explicit knowledge of \(I_j\), such as the formulation of its objective function.
Based on the training set \(T=\left\{E_j|1\leq j\leq n\right\}\), the goal is to train an optimizer that generalizes well to an unseen test set \(T^*\)  comprising problem instances from diverse problem classes and with dimensions that may differ from those in $T$.

Our neural optimizer, MEGO, is a MoEs (an ensemble of expert models) where each expert \(M_j\) is trained on an experience set \(E_j\).
As a whole, the MoEs collectively captures diverse structural patterns in the training data, adhering to the categorical modularization paradigm~\cite{darwen1997speciation} where different expert models specialize in solving different problems.
To obtain \(M_j\), classical machine learning methods can be used here, e.g., training a neural network based on \(E_j\) to model \(I_j\)'s unknown objective function \(F:\left\{0,1\right\}^{d_j}\rightarrow \mathbb{R}\).
However, there are two main limitations to this approach.
First, small changes in the discrete input, e.g., flipping several bits, can lead to substantial changes in the objective value, potentially causing \(M_j\) to overfit to local data points and hindering its ability to capture overall structural characteristics.
Second, this approach poses challenges when adapting \(M_j\) to new problem instances, as it is difficult to balance leveraging prior experience with incorporating new information.

\begin{algorithm}[tbp]
	\small
	\caption{Training Procedure of MEGO}
	\label{alg:train_mego}
	\SetKwInput{KwData}{Input}
	\SetKwInput{KwResult}{Output}
	\KwData{Training set \(T=\{E_1,E_2,\cdots,E_n\}\)}
	\KwResult{Expert models \(M_1,M_2,\cdots,M_n\)}
	\SetKw{To}{to}
	\SetKw{Append}{append}
	\SetKw{Into}{into}
	\SetKw{Break}{break}
	\SetKw{Return}{return}
	\SetKwProg{Fn}{Function}{:}{end}
	\SetKwFunction{Sort}{sort}
	\SetKwFunction{Normalize}{normalize}
	\SetKwFunction{Distance}{distance}
	\SetKwFunction{MSE}{MSE}
	\SetKwFunction{MLP}{MLP}
	\SetKwFunction{InsFeature}{ins\_feature}
	\SetKwComment{EmptyLine}{ }{ }
	\SetNoFillComment
	\For{\(j\leftarrow 1,2,\cdots,n\)}{
    \(y_{max} \leftarrow \max\{y_i | (\mathbf{x}_i, y_i) \in E_j\}\)\;
    \(y_{\min} \leftarrow \min \{y_i | (\mathbf{x}_i, y_i) \in E_j\}\)\;
    Create normalized dataset \(E'_j = \{(\mathbf{x}_i, \frac{y_i - y_{min}}{y_{max} - y_{min}}) | (\mathbf{x}_i, y_i) \in E_j\}\)\;
    Initialize parameters \(\theta, \phi, \omega\) for \(M_j\)'s encoder, decoder, and score predictor, respectively\;
    Train \(\theta, \phi, \omega\) on \(E'_j\) using stochastic gradient descent to minimize the loss in Eq.~(\ref{eq:loss})\;
   
	}
    \Return \(M_1,M_2,\cdots,M_n\)
\end{algorithm}

To address these limitations, we propose a decoupled architecture for the expert model based on variational autoencoder (VAE)~\cite{kingma2014autoencoding}.
The VAE maps a discrete solution into a compact, continuous latent representation.
By predicting objective values from within this regularized latent space, the model is compelled to learn the problem instance's global structural patterns, thereby mitigating overfitting to local patterns in the discrete domain.
Furthermore, the model's encoder-decoder architecture facilitates fine-tuning on new instances to solve them, as detailed in Sec.~\ref{sec:apply_mego}.
The model's architecture and training procedure are illustrated in Figs. 1a-1b, and are detailed below.

Specifically, each expert model \(M_j\) consists of an encoder \(f_\theta\), a decoder \(g_\phi\), and a latent score predictor \(h_\omega\), where $\theta$, $\phi$, and $\omega$ are trainable parameters.
The encoder-decoder is implemented as a VAE that explicitly regularizes the latent space to be smooth and compact.
Given an input solution \(\mathbf{x}_i\in\left\{0,1\right\}^{d_j}\), the encoder \(f_{\theta}\) predicts means \(\boldsymbol{\mu}\in\mathbb{R}^{d_z}\) and standard deviations \(\boldsymbol{\sigma}\in\mathbb{R}^{d_z}\) of a \(d_z\)-dimensional multivariate Gaussian distribution \(\mathcal{N}\left(\boldsymbol{\mu},\boldsymbol{\sigma}^2\boldsymbol{I}\right)\), from which a vector $\boldsymbol{z} \in \mathbb{R}^{d_z}$ is sampled.
Based on $\boldsymbol{z}$, The decoder \(g_\phi\left(\mathbf{z}\right)=\mathbf{x}_i^\prime\) reconstructs the input, while the score predictor \(h_\omega\left(\mathbf{z}\right)=y_i^\prime\) estimates the objective value.
Conceptually, \(M_j\) is a VAE augmented with a predictor \(h_\omega\) that introduces semantic meaning (i.e., the supervision signal from the objective values) into the latent space.
Once well trained, this latent space becomes a smooth and compact transformation of \(I_j\)'s original discrete solution space that also aligns well with its objective function. 

\begin{algorithm}[tbp]
	\small
	\caption{Application Procedure of MEGO}
    \label{alg:apply_mego}
	\SetKwInput{KwData}{Input}
	\SetKwInput{KwResult}{Output}
	\KwData{Training set \(T=\{E_1,E_2,\cdots,E_n\}\), expert models \(M_1,M_2,\cdots,M_n\), new problem instance \(I_{new}\).}
	\KwResult{High quality solution set \(X\) for \(I_{new}\).}
	\SetKw{To}{to}
	\SetKw{Append}{append}
	\SetKw{Into}{into}
	\SetKw{Break}{break}
	\SetKw{Return}{return}
	\SetKwProg{Fn}{Function}{:}{end}
	\SetKwFunction{Sort}{sort}
	\SetKwFunction{Normalize}{normalize}
	\SetKwFunction{Distance}{distance}
	\SetKwFunction{MSE}{MSE}
	\SetKwFunction{MLP}{MLP}
	\SetKwFunction{Pearson}{Pearson}
	\SetKwFunction{Spearman}{Spearman}
	\SetKwFunction{InsFeature}{ins\_feature}
	\SetKwComment{EmptyLine}{ }{ }
	\SetNoFillComment
	\(\left\{\left(\hat{\mathbf{x}}_i, \hat{y}_i\right)\right\}^s_{i=1}\leftarrow\) Randomly sample \(s\) solutions and evaluating them on the objective function of $I_{new}$\;
	\(X\leftarrow \varnothing\)\;
	\For{\(j\leftarrow 1,2,\cdots, n\)}{
	\(\left\{\hat{y}^\prime_i\right\}_{i=1}^s\leftarrow\) Predict the scores of \(\left\{\hat{\mathbf{x}}^\prime_i\right\}_{i=1}^s\) with \(M_j\)\;
    \(\left\{\hat{\pi}_i\right\}^s_{i=1}, \left\{\hat{\pi}_i^\prime\right\}^s_{i=1}\) $\leftarrow$ Sort \(\left\{\hat{y}_i\right\}^s_{i=1}\) and \(\left\{ \hat{y}_i^\prime\right\}^s_{i=1}\) respectively to obtain the rank sequences\;
	\(\rho_1\leftarrow\)\(\texttt{Pearson}\left(\left\{\hat{y}^\prime_i\right\}_{i=1}^s, \left\{\hat{y}_i\right\}_{i=1}^s\right)\)\;
	\(\rho_2\leftarrow\)\(\texttt{Spearman}\left(\left\{\hat{\pi}_i^\prime\right\}^s_{i=1}, \left\{\hat{\pi}_i\right\}^s_{i=1}\right)\)\;
	\If{\(\rho_1>0 \wedge \rho_2>0\)}{
	\(\left\{\mathbf{x}_i, y_i\right\}_{i=1}^{4s}\leftarrow\) Randomly sample \(4s\) solutions and objective values from \(E_j\)\;
	\(\left\{R_1^i\right\}_{i=1}^{m_1}\leftarrow\) Partition \(\left\{\hat{\mathbf{x}}^\prime_i\right\}_{i=1}^s\) into \(m_1\) disjoint subsets by y-values and sort them, where the y-values within the same subset are the same.\;
	\(\left\{R_2^i\right\}_{i=1}^{m_2}\leftarrow\) Partition \(\left\{\mathbf{x}_i\right\}_{i=1}^s\) into \(m_2\) disjoint subsets by y-values and sort them, where the y-values within the same subset are the same.\;
	\(m_{min}\leftarrow \min\left(m_1, m_2\right)\)\;
	\(R\leftarrow \cup_{i=1}^{m_{min}}R_1^i\times R_2^i\), where``\(\times\)'' denotes Cartesian product\;
	Freeze the parameters \(\theta, \omega\) of \(M_j\) and optimize \(\phi\) to minimize the loss in Eq.~(\ref{eq:fine_tune})\;
    \(\left\{\bar{\mathbf{x}}_i\right\}^k_{i=1}\leftarrow\)  Randomly sample \(p\) solutions, predict their scores with \(M_j\), and retain the top-$k$ solutions\;
	\(\left\{\bar{\mathbf{x}}^\prime_i\right\}_{i=1}^k\leftarrow \) Apply the encoder-decoder of $M_j$ to \(\left\{\bar{\mathbf{x}}_i\right\}^k_{i=1}\)\;
    $X \leftarrow X \cup \left\{\bar{\mathbf{x}}^\prime_i\right\}_{i=1}^k $\;
	}
	}
    $X \leftarrow $ Evaluate all solutions in $X$ on the objective function of $I_{new}$ and retain the top-$k$ solutions\;
\Return{X}
\end{algorithm}

The training procedure of MEGO is outlined in Alg.~\ref{alg:train_mego}.
The encoder \(f_\theta\), decoder \(g_\phi\), and score predictor \(h_\omega\) are all neural networks and are jointly trained using normalized samples in \(E_j\) (lines 2-4 in Alg.\ref{alg:train_mego}).
The overall training objective of $M_j$ is to minimize a composite loss function:
\begin{equation}
\label{eq:loss}
    \min_{\theta,\phi,\omega}\frac{1}{\left\vert E_j\right\vert}\sum_{\left(\mathbf{x}_i,y_i\right)\in E_j}{
        \begin{aligned}[t]
        & \left\Vert\mathbf{x}_i-\mathbf{x}_i^\prime\right\Vert^2+\lambda\left\Vert y_i-y_i^\prime\right\Vert^2 \\
             & + \gamma\mathbb{D}_{\textrm{KL}}{\left(\mathcal{N}\left(\boldsymbol{\mu},\boldsymbol{\sigma^2I}\right)\Vert\mathcal{N}\left(\boldsymbol{0},\boldsymbol{I}\right)\right)}
        \end{aligned},
    }
\end{equation}
where \(\boldsymbol{\mu},\boldsymbol{\sigma}=f_\theta\left(\mathbf{x}_i\right)\), \(\mathbf{z}\sim\mathcal{N}\left(\boldsymbol{\mu},\boldsymbol{\sigma^2I}\right)\) , \(\mathbf{x}_i^\prime=g_\phi\left(\mathbf{z}\right)\), and \(y^\prime=h_{\omega}\left(\mathbf{z}\right)\).
Overall, Eq.~(\ref{eq:loss}) consists of three parts:
1) a reconstruction loss measured by square error between \(\mathbf{x}_i\) and \(\mathbf{x}_i^\prime\), encouraging the VAE to capture the essential information of the discrete input;
2) a score prediction loss measured by square error between \(y_i\) and \(y_i^\prime\), encouraging the score predictor to be accurate;
and 3) a regularization loss measured by the Kullback-Leibler (KL) divergence between the learned probability distribution \(\mathcal{N}\left(\boldsymbol{\mu},\boldsymbol{\sigma^2I}\right)\) over the latent space and a predefined prior distribution (the standard normal distribution \(\mathcal{N}\left(\boldsymbol{0},\boldsymbol{I}\right)\)), which promotes a smooth and compact latent space.
Here, \(\lambda\) and \(\gamma\) are two weighting hyper-parameters. 

\subsection{Employing MEGO to Solve a New Problem Instance}
\label{sec:apply_mego}
\begin{figure*}[tbp]
    \centering
    \includegraphics[width=0.85\textwidth]{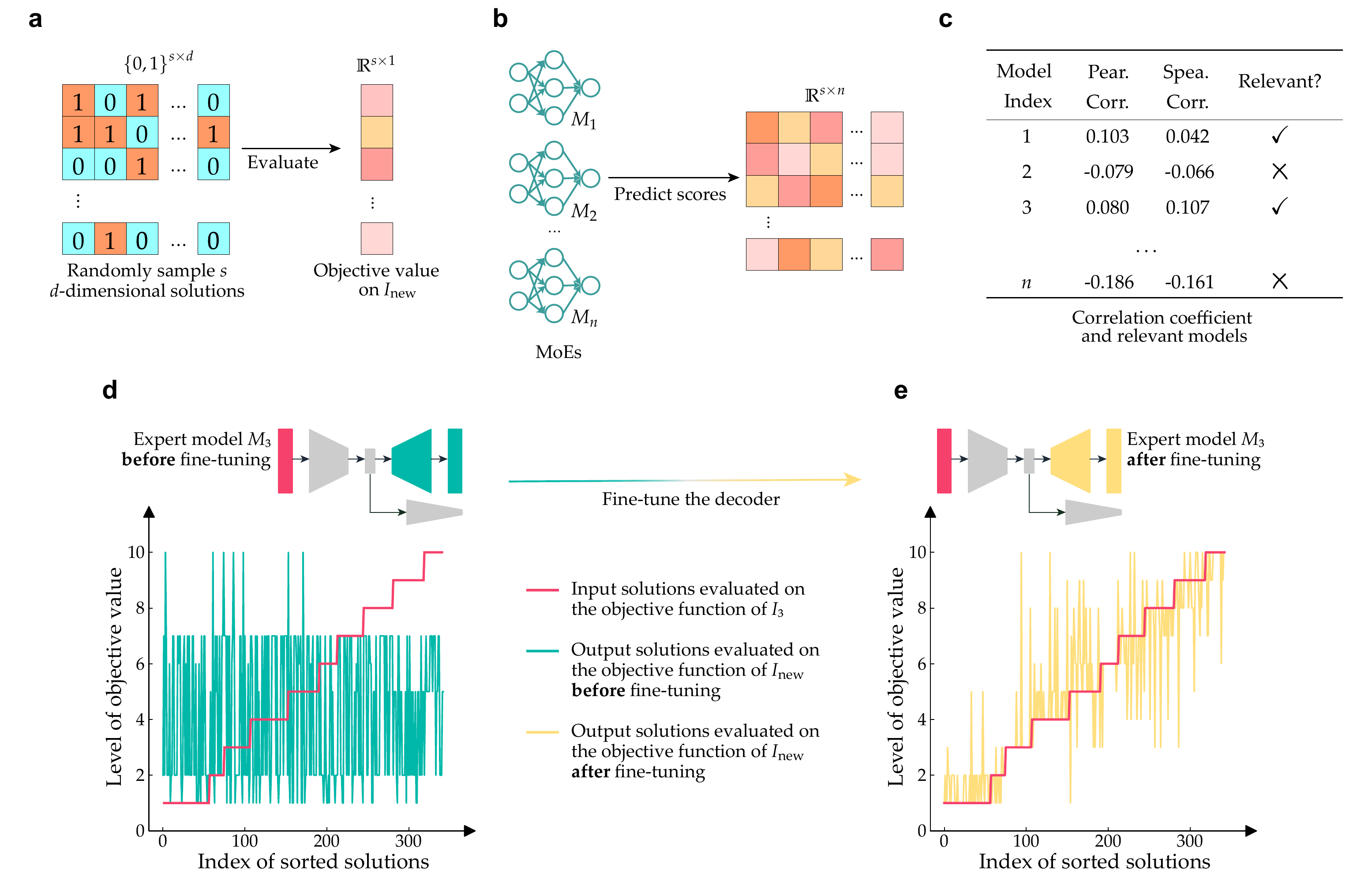}
    \caption{ \textbf{Illustrations of employing MEGO to solve a test instance \(I_{new}\) that belongs to the compiler argument optimization problem class.} \textbf{a}, s randomly sampled \(d\)-dimensional (\(d\) is the problem dimension of \(I_{new}\)) solutions and their objective values evaluated on the objective function of \(I_{new}\). \textbf{b}, each expert model predicts the scores of these solutions. \textbf{c}, based on the predictions, two correlation coefficients are calculated, and an expert model is considered relevant when both coefficients are larger than 0; \(M_3\) is one relevant model. \textbf{d}, before fine-tuning \(M_3\), for the input solution \(\mathbf{x}_i\) of \(M_3\) and its output solution \(\mathbf{x}_i^\prime\), there is almost no correlation between the objective value of \(\mathbf{x}_i\) on \(I_3\) and the objective value of \(\mathbf{x}_i^\prime\) on \(I_{new}\) (here we divide the objective values into 10 levels for visualization). \textbf{e}, after fine-tuning \(M_3\), the objective value of \(\mathbf{x}_i\) on \(I_3\) and the objective value of \(\mathbf{x}_i^\prime\) on \(I_{new}\) are largely aligned, indicating that high-quality solutions of \(I_3\) can be mapped through \(M_3\) to generate high-quality solutions for \(I_{new}\).}
    \label{fig:employing_mego}
\end{figure*}

When presented with a new problem instance $I_{new}$, MEGO follows a three-step procedure: 
1) identifying relevant experts,
2) fine-tuning them,
and 3) generating solutions.
This select-adapt-generate process is designed to efficiently utilize knowledge from previously solved training problem instances for $I_{new}$, requiring only a small budget of function evaluations~(\#FEs) on $I_{new}$.
The application procedure of MEGO is shown in Figs. 1c-1e, with the pseudo-code outlined in Alg.~\ref{alg:apply_mego}.
For a concrete illustration, Fig.~\ref{fig:employing_mego} also details these steps as MEGO solves a test instance of the compiler argument optimization problem from our experiments.

\subsubsection{Relevant Expert Identification}
\label{sec:expert_selection}
This step aims to identify experts whose learned structural patterns correlate strongly with $I_{new}$, ensuring that only the most suitable knowledge is utilized.
It is governed by MEGO's routing policy (Fig 1c), a rule-based mechanism that quantifies the correlation between \(M_j\) and \(I_{new}\), measured by the alignment between the objective values of the sampled solutions on \(I_{new}\) and the objective values predicted by \(M_j\).
Specifically, a small set of \(s\) solutions, \(\left\{\left(\hat{\mathbf{x}}_i, \hat{y}_i\right)\right\}^s_{i=1}\), is obtained by sampling uniformly at random and evaluating on the objective function of \(I_{new}\) (Fig. 2a and line 1 in Alg.~\ref{alg:apply_mego}).
Each expert model \(M_j\) is then applied to \(\left\{\hat{\mathbf{x}}_i\right\}^s_{i=1}\) to predict their objective values.
The results are denoted as  \(\left\{\left(\hat{\mathbf{x}}_i, \hat{y}_i^\prime\right)\right\}^s_{i=1}\), where \(\hat{y}_i^\prime=h_\omega\left(f_\theta\left(\hat{\mathbf{x}}_i\right) \right) \) is the predicted score.
If the dimension of \(I_{new}\) is smaller than an expert's input dimension, the sampled target-instance solutions are padded with zeros during relevant expert identification; if it is larger, truncation is used instead.
These two sets of results, \(\left\{\left(\hat{\mathbf{x}}_i, \hat{y}_i\right)\right\}^s_{i=1}\) and \(\left\{\left(\hat{\mathbf{x}}_i, \hat{y}_i^\prime\right)\right\}^s_{i=1}\), are then sorted separately based on their respective \(y\)-values to obtain their corresponding rank sequences, \(\left\{\hat{\pi}_i\right\}^s_{i=1}\) and  \(\left\{\hat{\pi}_i^\prime\right\}^s_{i=1}\) (Fig. 2b and lines 4-5 in Alg.~\ref{alg:apply_mego}).

To measure the correlation, two coefficients \(\rho_1\) and \(\rho_2\) are calculated (Fig. 2c and lines 6-7 in Alg.~\ref{alg:apply_mego}).
An expert \(M_j\) is considered relevant to \(I_{new}\) if \(\rho_1>0\) and \(\rho_2>0\):
\begin{itemize}
    \item \(\rho_1\): the Pearson correlation between the actual and predicted objective values, i.e., \(\left\{\hat{y}_i\right\}^s_{i=1}\) and \(\left\{\hat{y}_i^\prime\right\}^s_{i=1}\). This captures the linear relationship between the scores.
    \item \(\rho_2\): the Spearman's rank coefficient between the ranks of actual and predicted values, i.e., \(\left\{\hat{\pi}_i\right\}^s_{i=1}\) and  \(\left\{\hat{\pi}_i^\prime\right\}^s_{i=1}\). This measures the ordinal relationship, assessing whether the expert can correctly rank the solutions by quality.
\end{itemize}

\subsubsection{Expert Fine-Tuning}
All relevant models are then efficiently adapted to \(I_{new}\).
For each relevant model \(M_j\), only its decoder \(g_\phi\) is fine-tuned while the other parts (encoder and latent score predictor) are kept fixed (Fig. 1d), allowing the model to incorporate new information while preserving the learned knowledge.
The goal of fine-tuning is to establish a transformation that aligns the solution spaces of \(I_j\) and \(I_{new}\) (Figs. 2d-2e).
This is achieved by training the decoder $g_{\phi}$ on a specially constructed dataset (lines 9-14 in Alg.~\ref{alg:apply_mego}), \(\left\{\left(\mathbf{x}_i,\hat{\mathbf{x}}_i\right)\right\}_{i=1}\).
This dataset pairs solutions from the expert's original experience set, $\mathbf{x}_i \in E_j$, with the newly sampled solutions, $\hat{\mathbf{x}}_i$, from $I_{new}$, enforcing ordinal consistency (i.e., better solutions are mapped to better solutions).

Specifically, to construct such a mapping dataset, we first sort the solutions sampled from \(I_{new}\) according to the objective values, denoted as \(R_1\).
Then we sample \(4s\) solutions from the training set \(E_j\) of \(M_j\) and also sort them, denoted as \(R_2\).
Within each of \(R_1\) and \(R_2\), there may be solutions with the same objective value.
Supposing there are \(m_1\) and \(m_2\) unique objective values within \(R_1\) and \(R_2\), respectively, we partition \(R_1\)  and \(R_2\) into two sequences of \(m_1\) and \(m_2\) disjoint subsets and sort them, respectively, where the y-values within the same subset are the same: \(R_1=\cup_{i=1}^{m_1}R_1^i\)
and \(R_2=\cup_{i=1}^{m_2}R_2^i\).
Finally, from both sequences of subsets, we select the first \(m_{min}=\min\left\{m_1,m_2\right\}\) subsets, constructing \(m_{min}\) pairs \(\left\{\left(R_1^i, R_2^i\right)\right\}_{i=1}^{m_{min}}\).
The Cartesian product of each pair will constitute the mapping dataset , where ``\(\times\)'' denotes Cartesian product.
Given the mapping dataset \(R=\cup_{i=1}^{m_{min}}R_1^i\times R_2^i=\left\{\left(\mathbf{x}_i,\hat{\mathbf{x}}_i\right)\right\}_{i=1}\), the loss function for fine-tuning \({M_j}\) is:
\begin{equation}
\label{eq:fine_tune}
    \min_\phi\frac{1}{\left\vert R\right\vert}\sum_{\left(\mathbf{x}_i,\hat{\mathbf{x}}_i\right)\in R}\left\Vert \mathbf{x}_i^\prime-\hat{\mathbf{x}}_i\right\Vert^2,
\end{equation}
where \(\boldsymbol{\mu},\boldsymbol{\sigma}=f_\theta\left(\mathbf{x}_i\right)\), \(\mathbf{z}\sim\mathcal{N}\left(\boldsymbol{\mu},\boldsymbol{\sigma^2I}\right)\), \(\mathbf{x}_i^\prime=g_\phi\left(\mathbf{z}\right)\), and \(\phi\) denotes the trainable parameters of the decoder \(g_\phi\).
During fine-tuning, the final layer of the decoder will be adjusted to match the problem dimension of \(I_{new}\). 

\subsubsection{Solution Generation}
\label{sec:solution_generation}
Each fine-tuned \(M_j\) is used to generate candidate solutions for \(I_{new}\) (Fig. 1e).
Since the encoder-decoder of fine-tuned \(M_j\) serves as an effective transformation from the solution space of \(I_j\) to that of \(I_{new}\), high-quality solutions for \(I_j\) can be mapped to generate high-quality solutions for $I_{new}$.
Specifically, a large number of $p$ solutions are randomly sampled in the input space of the encoder of \(M_j\), and their scores, as predicted by \(M_j\), are obtained.\footnote{An alternative is to search directly in the latent space of a selected expert. An empirical comparison with latent-space search is provided in Section~\ref{sec:latent_search}.}
These solutions are sorted based on the scores, and the top-\(k\) unique solutions, denoted as \(\left\{\bar{\mathbf{x}}_i\right\}_{i=1}^k\) (line 15 in Alg.~\ref{alg:apply_mego}), are retained, where \(k\) is a hyper-parameter.
It is worth noting that these retained solutions are high-quality solutions in the source expert's solution space, rather than in the solution space of \(I_{new}\).
The encoder-decoder of \(M_j\) is then applied to map \(\left\{\bar{\mathbf{x}}_i\right\}_{i=1}^k\) to a new set of solutions \(\left\{\bar{\mathbf{x}}^\prime_i\right\}_{i=1}^k\) tailored for \(I_{new}\) (line 16 in Alg.~\ref{alg:apply_mego}).
Solutions generated from all fine-tuned models are aggregated, evaluated on \(I_{new}\), and the final top-\(k\) solutions are returned (line 20 in Alg.~\ref{alg:apply_mego}).

The total \#FEs used by MEGO to generate solutions for \(I_{new}\) is \(s+k\times m\), where \(m\) is the number of relevant expert models.
In the experiments, $s$ and $k$ are set to small values ($s=64$ and $k=4$), while \(m\) can vary across different problem instances but is generally of moderate size.
Typically, MEGO consumes around 100 FEs in total to solve a problem instance.

\subsection{Constraint Handling and Application Modes of MEGO}
\label{sec:discuss_mego}

MEGO does not incorporate a built-in mechanism for handling constraints, as constraints in optimization problems are often highly problem-dependent.
Integrating problem-specific constraint-handling mechanism into the architecture would compromise the model's generality, conflicting with our primary goal of creating a general-purpose optimizer.
To address constrained problems, MEGO uses simple, external constraint-handling techniques that operate as a wrapper around it.
During training phase, any sampled solution that violates constraints is first transformed into a feasible one by a repair heuristic.
Similarly, when MEGO generates solutions for a constrained problem instance, the same technique is applied to ensure the final outputs are feasible.
This straightforward approach allows MEGO to maintain its general-purpose characteristic while effectively solving constrained optimization problems.

MEGO can be employed in two modes:
\begin{itemize}
\item \textbf{As a standalone optimizer:} MEGO can be used directly to find high-quality solutions for a new problem instance within a small budget of \#FEs.
\item \textbf{As an initial solution generator:} The solutions produced by MEGO can provide a set of high-quality starting points to warm-start other optimizers. 
As demonstrated in the experiments, MEGO's output can be used to initialize the population of GA, provide effective starting points for HC, or serve as the initial design for BO, thereby boosting their performance.
\end{itemize}

\section{Experiments}
\label{sec:exp}
The experiments mainly aim to answer the following key questions.
First, as a general-purpose optimizer that targets at diverse binary problem classes, the most important aspect to evaluate is its generalization ability across problem classes.
Specifically, can MEGO effectively solve problem instances from classes and dimensionalities that are not part of its training data  (\textbf{RQ1})?
Second, for the problem classes that are included in the training set, how well does MEGO generalize to new instances within those same classes~(\textbf{RQ2})?

Beyond evaluating performance, the experiments also explore a novel application of MEGO.
A computational approach to problem classification is developed based on vector representations of problem instances derived from MEGO's fine-tuning procedure (see Sec.~\ref{sec:new_classification}).
This approach allows for measuring the similarity between different problem instances, leading to a data-driven problem classification.
This gives rise to the third key question: How does this learned classification compare with conventional, analysis-based problem classification, and what new relationships between problems might it reveal (\textbf{RQ3})?

To address these questions, a comprehensive experimental setup is established across six problem classes from various disciplines.
This includes three classic problems---the generalized One-Max Problem (OM)~\cite{eshelman1991crossover}, the Knapsack Problem (KP)~\cite{martello1990knapsack}, and the Max-Cut Problem (MC)~\cite{garey1979computers}---as well as three real-world applications: Compiler Arguments Optimization (CA)~\cite{jiang2021smartest}, Complementary Influence Maximization (CIM) on social networks~\cite{lu2015competition}, and Anchor Selection (AS) for pose estimation in camera imaging~\cite{hruby2022learning}.

The training set is composed of 27 instances exclusively from the three classic problem classes (OM, KP, and MC).
For each class, three instances are generated for each of the dimensions 30, 35, and 40.
To generate the training data, each of these instances is associated with an experience set of 20,000 pairs of (solution, objective value), where solutions are sampled uniformly at random.
In contrast, the test set comprises 72 instances from all six problem classes, including the three real-world applications (CA, CIM, and AS) unseen during training.
The dimensions of the test instances are also higher---specifically 40, 60, 80, and 100---with three instances per dimension for each class.
This experimental setup is motivated by a practical consideration: data for classic problems is often abundant or easy to generate, whereas data for real-world applications can be scarce and expensive to obtain.
Therefore, demonstrating that an optimizer trained on accessible classic problems can generalize effectively to challenging real-world applications strongly supports its practical utility.
Since  the black-box setting is considered here, in the experiments, all the general-purpose optimizers interact with the problem instances solely through function evaluations.
The Python code of MEGO, problem instances, baselines, and the scripts for repeating our experiments are open-sourced at \url{https://github.com/MetaronWang/MEGO}.

\subsection{Problem Classes, Instances, and Constraint Handling}
During the six problem classes, KP, MC, CIM, and AS involve constraints.
As noted in Sec.~\ref{sec:discuss_mego}, MEGO does not handle constraints directly.
Instead, external constraint-handling techniques are applied.
Details on each problem class, including their objective functions, constraint-handling techniques, and problem instances, are provided below.

\subsubsection{One-Max Problem (OM)}
The OM problem~\cite{eshelman1991crossover} is a fundamental binary optimization task, often serving as a benchmark for analyzing the theoretical performance of search-based optimizer.
This work employs a generalized version of OM to create a diverse set of instances.
Specifically, the objective of OM is to find a binary vector $\mathbf{x}$ that minimizes the Hamming distance to a predefined reference vector $\hat{\mathbf{x}}$.
This is equivalent to maximizing the number of matching bits.
The original OM problem is a special case where the reference vector consists of all ones.

For a $d$-dimensional OM instance with a reference vector \(\hat{\mathbf{x}} \in \{0,1\}^d\), the objective function is:
\begin{equation}
\text{maximize} \quad F\left(\mathbf{x}\right)=d-D_{hamming}\left(\mathbf{x},\hat{\mathbf{x}}\right).
\end{equation}

In the experiments, OM instances are generated by constructing the refenrece vector \(\hat{\mathbf{x}}\).
Each element of \(\hat{\mathbf{x}}\) is sampled independently from a uniform distribution over $\{0, 1\}$.

\subsubsection{Knapsack Problem (KP)}
The KP~\cite{martello1990knapsack} is a classic combinatorial optimization problem.
The goal is to select a subset of items to maximize their total value, subject to a total weight constraint.
For a KP instance with $d$ items, where each item $i$ has a value $v_i$ and a weight $w_i$, a solution is represented by a binary vector $\mathbf{x} \in \{0,1\}^d$.
Here, $x_i=1$ indicates that item $i$ is selected.
The objective function is:
\begin{equation}
\begin{aligned}
\text{maximize} \quad & F(\mathbf{x}) = \sum_{i=1}^d v_i x_i \\
\text{subject to} \quad & \sum_{i=1}^d w_i x_i \leq w_{\text{max}}
\end{aligned},
\end{equation}
where $w_{\text{max}}$ is the weight capacity.

In the experiments, KP instances are generated as follows.
Both the value vector $\mathbf{v}$ and the weight vector $\mathbf{w}$ are populated with elements sampled uniformly from $[0, 1]$.
To introduce a correlation between values and weights, the two vectors are sorted in tandem, ensuring that a higher value corresponds to a higher weight.
The weight capacity $w_{\text{max}}$ is then set as a fraction of the total possible weight, calculated as $w_{\text{max}} = r \sum_{i=1}^d w_i$, where $r$ is sampled uniformly from $[0.2, 0.8]$.

Given a KP instance, infeasible solutions are repaired using a simple heuristic.
It processes the solution vector $\mathbf{x}$ sequentially from index $i=1$ to $d$.
For each selected item $i$ ($x_i=1$), it checks if adding its weight would cause the solution to be infeasible.
If so, the selection is reversed by setting $x_i$ to 0.
This check is performed for all items in order, guaranteeing that the repaired solution is feasible.

\subsubsection{Max-Cut Problem (MC)}
The MC problem~\cite{garey1979computers} aims to partition the vertices of a graph into two disjoint sets to maximize the number of edges connecting the two sets.
This work focuses on a constrained variant where one of the two partitions is limited to a maximum size.
Given an undirected graph with $d$ vertices, represented by its adjacency matrix $A \in \{0, 1\}^{d \times d}$, a partition can be defined by a binary vector $\mathbf{x} \in \{0,1\}^d$, where $x_i=1$ assigns vertex $i$ to the first partition, and $x_i=0$ assigns it to the second.
The objective function is:
\begin{equation}
\label{eq:mc}
\begin{aligned}
\text{maximize} \quad & F(\mathbf{x}) = \sum_{i=1}^d \sum_{j=i+1}^d A_{ij} |x_i - x_j| \\
\text{subject to} \quad & \sum_{i=1}^d x_i \leq b
\end{aligned},
\end{equation}
where $b$ is the maximum allowed size of the first partition.

In the experiments, problem instances are generated by creating graph structure and the partition size limit. 
Undirected graphs with $d$ vertices are generated using the \texttt{NetworkX} library~\cite{hagberg2008exploring}.
The number of edges is set to $r d^2$, where $r$ is sampled uniformly from $[0.2, 0.4]$.
To ensure a meaningful problem structure, only connected graphs are used; any disconnected graphs are discarded and then regenerated.
The maximum size for the first partition, i.e., $b$, is set to $\lfloor r' d \rfloor$, with $r'$ is sampled uniformly from $[0.2, 0.4]$.

A repair heuristic, analogous to the one for KP, handles solutions violating the constraint.
The heuristic iterates through the solution vector $\mathbf{x}$ sequentially from $i=1$ to $d$.
If it encounters a selected vertex ($x_i=1$) after the size of the first partition has already reached the limit $b$, the selection is reversed ($x_i$ is set to 0).

\subsubsection{Compiler Arguments Optimization Problem (CA)}
The CA problem~\cite{jiang2021smartest} seeks to identify the optimal set of compiler arguments for a given source code.
The goal is to minimize the size of the generated executable file.
This is a practical black-box optimization problem, as the objective function involves an actual compilation process.
Given a source code file $G$ and a set of $d$ available compiler arguments, a solution $\mathbf{x} \in \{0, 1\}^d$ represents a compiler configuration where $x_i=1$ enables the $i$-th argument.
The objective function is:
\begin{equation}
\text{minimize} \quad F(\mathbf{x}) = \textrm{Executable\_Size}(G, \mathbf{x}).
\end{equation}

In the experiments, CA instances are generated using source files randomly selected from the \texttt{cbench} and \texttt{polybench-cpu} benchmarks~\cite{fursin2009collective}.
The \texttt{GCC} compiler is used, and $d$ arguments are randomly selected from its 186 available arguments to generate a $d$-dimensional CA instance.

\subsubsection{Complementary Influence Maximization Problem (CIM)}
The CIM problem~\cite{lu2015competition} extends the classic IMP~\cite{kempe2003maximizing} by introducing complementary users.
The goal is to select a set of ``seed'' nodes $S_B$ for one opinion $B$ to maximize its spread in a social network, given that a complementary opinion $A$ is already being propagated from a predefined seed set $S_A$.
Formally, given a social network $\mathcal{G}$ and $S_A$, CIM aims to find a node set $S_B$ from a candidate set of $d$ nodes to maximize the expected number of nodes activated by opinion $B$, subject to a constraint $b$ on the size of $S_B$:
\begin{equation}
\begin{gathered}
\text{maximize} \quad F(\mathbf{x}) = \textrm{ActiveNum}(\mathcal{G}, S_A, \mathbf{x}) \\
\text{subject to} \quad \sum_{i=1}^{d} x_i \leq b
\end{gathered},
\end{equation}
where $\mathbf{x} \in \{0,1\}^d$ and $x_i=1$ indicates the $i$-th node is included into $S_B$.
The interaction between opinions $A$ and $B$ are governed by parameters $q_{A|\emptyset}$, $q_{A|B}$, $q_{B|\emptyset}$, and $q_{B|A}$, as detailed in~\cite{lu2015competition}. 

In the experiments, CIM instances are generated using real-world social networks.
Specifically, the social network $\mathcal{G}$ is randomly chosen from the \texttt{Wiki}~\cite{leskovec2010predicting} and \texttt{Facebook} datasets~\cite{mcauley2012learning}.
The candidate node set and $S_A$ are formed by randomly selecting nodes from the network.
The budget $b$ is set to $\lfloor r d \rfloor$, with $r$ sampled uniformly from $[0.2, 0.4]$.
The interaction parameters between opinions are set following~\cite{lu2015competition}: $q_{A|\emptyset}=0.5$, $q_{A|B}=0.7$, $q_{B|\emptyset}=0.5$, and $q_{B|A}=0.5$.

A repair heuristic similar to the one for KP is employed to handle infeasible solutions.
If a solution vector violates the size constraint, it iterates through the solution $\mathbf{x}$ sequentially from $i=1$ to $d$, and excess selections are disabled by setting their corresponding bits to 0.

\subsubsection{Anchor Selection Problem (AS)}
The AS Problem arises in selecting robust initial points for numerical solvers in camera imaging~\cite{hruby2022learning}.
The goal is to select a small subset of ``anchor points'' (of size at most $b$) from a larger candidate set, such that these anchors collectively ``cover'' or correctly solve the maximum number of problems from a given problem set.
This task can be framed as the Maximum Coverage problem.
Let $M = \{m_1, \dots, m_d\}$ be a candidate set of $d$ anchor points and $Q$ be a set of problems.
For each point $m_i \in M$, let $Q_i \subseteq Q$ be the subset of problems it can solve.
A solution is represented by a binary vector $\mathbf{x} \in \{0, 1\}^d$, where $x_i$ indicates $m_i$ is selected.
The objective function is:
\begin{equation}
\begin{gathered}
\text{maximize} \quad F(x) = \left| \bigcup_{i=1, x_i=1}^d Q_i \right| \\
\text{subject to} \quad \sum_{i=1}^d x_i \leq b
\end{gathered}
\end{equation}

In the experiments, AS instances are generated by randomly selecting scenes from the \texttt{ETH3D} dataset~\cite{schoeps2017eth3dstereodataset}, following~\cite{hruby2022learning}.
A problem set $Q$ of size 100,000 is generated from a selected scene.
The candidate anchor set $M$ of size $d$ is created by sampling points from $Q$.
The size limit $b$ for the anchor set is set to $\lfloor r d \rfloor$, where $r$ 
is uniformly sampled from $[0.1, 0.6]$.

A repair heuristic same as the one for CIM is employed.
If a solution vector violates the size constraint, it iterates through the solution and excess selections are disabled.

\begin{table}[tbp]
  \centering
  \caption{Hyper-parameter Settings of MEGO}
    \begin{tabular}{ccc}
    \toprule
    \multicolumn{2}{c}{Hyper-parameters} & Values \\
    \midrule
    {\multirow{4}[14]{*}{VAE}} & \makecell{\#hidden units of\\encoder} & [64, 128, 128, 64] \\
\cmidrule{2-3}          & \makecell{\#hidden units of\\decoder} & [64, 128, 128, 64] \\
\cmidrule{2-3}          & \#dims of latent space & 4\(\times\) input dimension\\
\cmidrule{2-3}          & \makecell{activation function \&\\ batch normalization} & \makecell{LeakyReLU in every layer\\except HardTanh in the last\\layer of decoder;\\ Batch normalization used} \\
    \midrule
    {\multirow{2}[4]{*}{\makecell{Latent score\\predictor}}} & \#hidden units & \makecell{[128, 256, 512, 1024,\\512, 256, 128]} \\
\cmidrule{2-3}          & \makecell{activation function \&\\ batch normalization} & \makecell{ReLU in the last layer;\\Batch normalization used} \\
    \midrule
    {\multirow{2}[3]{*}{\makecell{Weighting\\loss function}}} & \(\lambda\) & {1} \\
\cmidrule{2-3}          & \(\gamma\) & {0.0025} \\
    \midrule
    {\multirow{2}[3]{*}{Training}} & learning rate & \makecell{0.0005, Adam Optimizer~\cite{kingma2015adam}\\without weight decay} \\
\cmidrule{2-3}          & batch size & {1024} \\
    \midrule
    {\multirow{3}[6]{*}{Fine-tuning}} & \(s\) & {64} \\
\cmidrule{2-3}          & learning rate & \makecell{0.001, Adam Optimizer~\cite{kingma2015adam}\\without weight decay} \\
\cmidrule{2-3}          & batch size & {1024} \\
    \midrule
    {\multirow{2}[3]{*}{\makecell{Solution\\generation}}} & \(k\) & {4} \\
\cmidrule{2-3}          & \(p\) & \(2\times10^6\) \\
    \bottomrule
    \end{tabular}%
  \label{tab:param}%
\end{table}%

\begin{figure*}[tbp]
	\centering
	\includegraphics[width=0.9\textwidth]{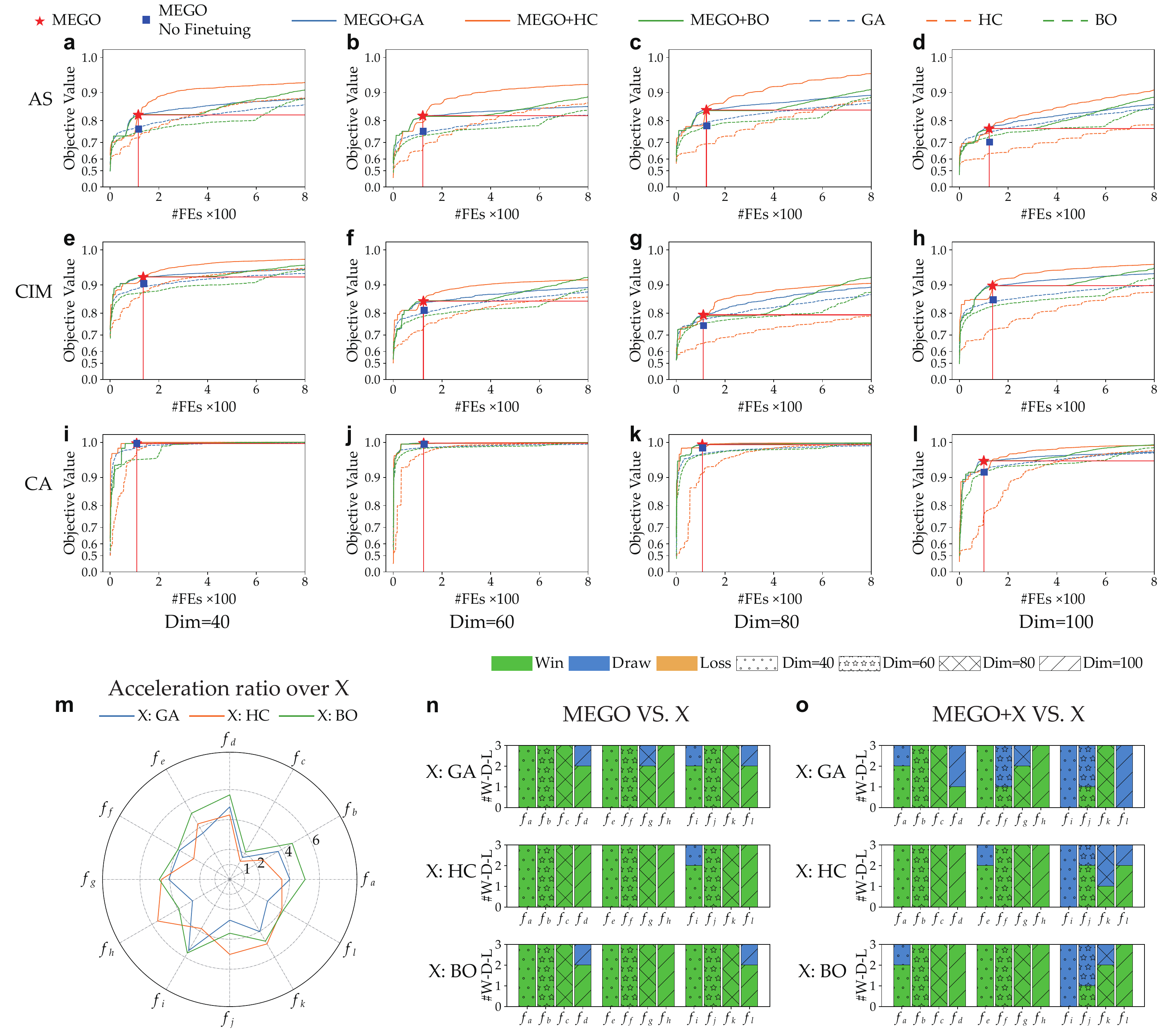}
	\caption{\textbf{Testing performance on the real-world problem classes beyond the training set}. \textbf{a-l}, averaged convergence curves of 30 independent runs across 3 test instances. \textbf{m}, acceleration ratios of MEGO compared to GA, HC, and BO. Here, \(f_a\)- \(f_l\) indicate the test instances in fig. a – fig. l, respectively. \textbf{n}, W-D-L counts derived from statistical results of comparing MEGO vs. X (X: GA, HC and BO) under the first stopping criterion. \textbf{o}, W-D-L counts derived from statistical results of comparing MEGO+X vs. X (X: GA, HC and BO) under the second stopping criterion.}
	\label{fig:mego_test_domain}
\end{figure*}

\subsection{Compared Optimizers and Experimental Protocol}
MEGO is compared with widely-adopted general-purpose optimizers for binary optimization, including GA~\cite{holland1992genetic}, Hill Climbing with random restart (HC)~\cite{russell2016artificial}, and BO~\cite{lindauer2022smac3}.
The parameters of these baselines are tuned or set according to their original publications.
Specifically, GA employs the elitism mechanism, the single-point crossover operator and random flip mutation operator.
Its population size is 32, the number of elites is 1, and the mutation rate is set to $1/d$, where $d$ is the problem dimension.
HC is based on the bit-flip operator and has no parameters.
BO is implemented via the SMAC3 package~\cite{lindauer2022smac3}, following the suggested settings in its official documentation suitable for various scales of discrete optimization problems.
Specifically, random forest is used as the surrogate model and log expected improvement is adopted as the acquisition function.
The initial design is generated using a scrambled Sobol sequence, with the number of initial \#FEs limited to 25\% of the total budget.
Finally, a comparison is also made with SMARTEST~\cite{jiang2021smartest}, a specialized optimizer for the CA problem.
The parameters of SMARTEST are set following its original paper~\cite{jiang2021smartest}, using a population size of 100, a crossover rate of 0.8, and an elitism rate of 0.1.
We also include a variant of MEGO to examine the effect of target-specific adaptation.
Compared with MEGO, the variant removes the decoder fine-tuning step (line 14 in Alg.~\ref{alg:apply_mego}) after expert selection, while keeping the total FE budget the same as in MEGO for a fair comparison.

The hyper-parameter settings for MEGO are summarized in Table~\ref{tab:param}.
Its structural hyper-parameters (e.g., VAE architecture), learning hyper-parameters (e.g., learning rate), and weighting hyper-parameters are manually tuned for stable training.
Details of the manual tuning process are also provided in the supplementary material.
To ensure MEGO consumes within a small budget (around 100) of \#FEs, $s$ and $k$ are set to 64 and 4, respectively.
In terms of wall-clock time, the fine-tuning step of MEGO uses 20000 epochs and takes about 120-180 seconds, while sampling and scoring $2\times 10^6$ source-space solutions takes about 3 seconds on our reference machine.
As aforementioned in Sec.~\ref{sec:discuss_mego}, MEGO can be employed in two modes.
Both modes are evaluated in the experiments.
First, MEGO is assessed as a standalone optimizer and compared directly against the baselines.
Second, MEGO is utilized as an initial solution generator to warm-start these optimizers.
In the second mode, the combined optimizer is denoted as MEGO+X (e.g., MEGO+GA).
For MEGO+GA, the solutions generated by MEGO will be directly inserted into its initial population.
For MEGO+HC, the generated solutions will serve as candidates for the starting point at each restart.
For MEGO+BO, the generated solutions are used to initialize the surrogate model.

\begin{figure*}[tbp]
	\centering
	\includegraphics[width=0.9\textwidth]{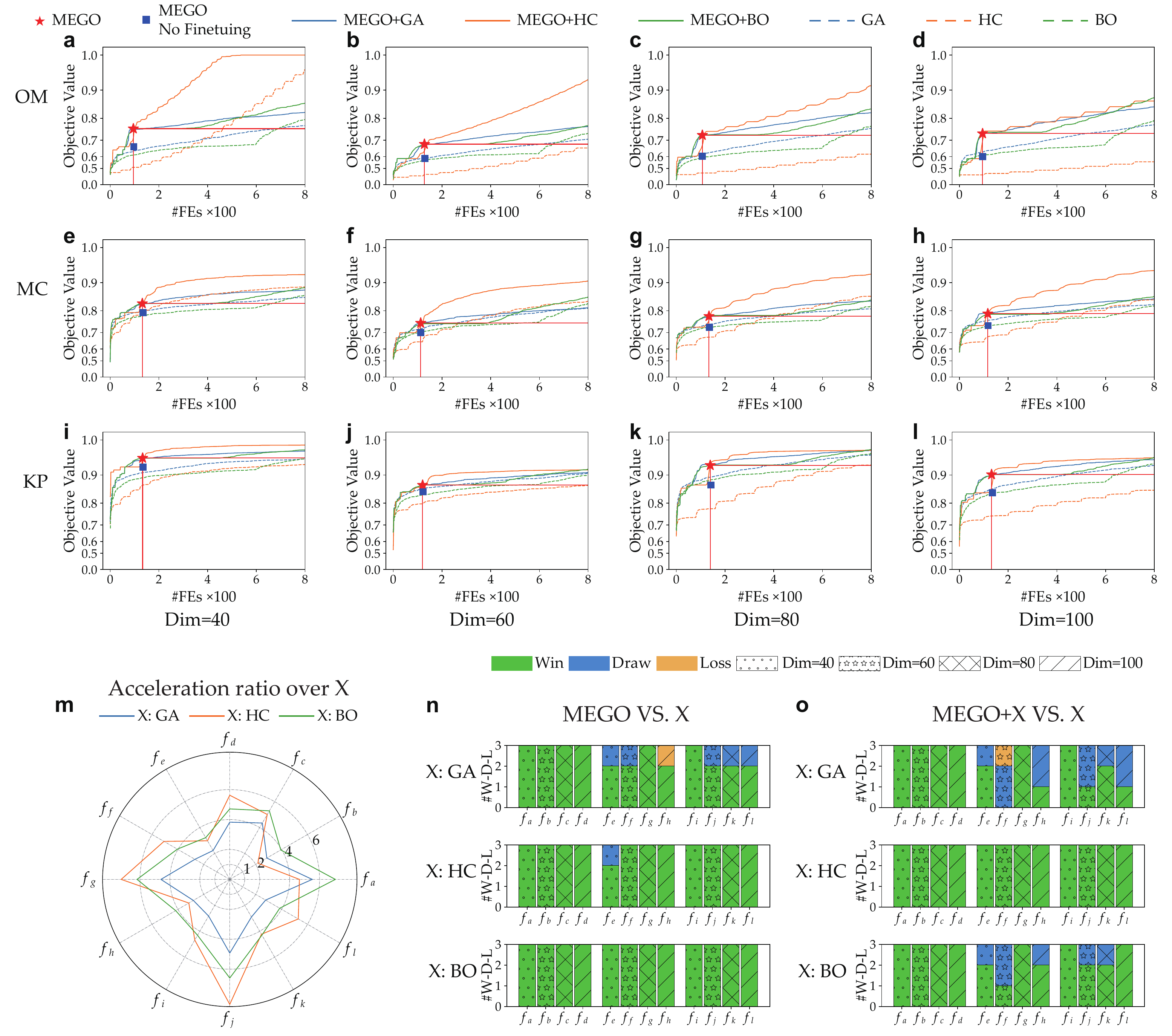}
	\caption{\textbf{Testing performance on the three classic problem classes that appeared in the training set}. \textbf{a-l}, averaged convergence curves of 30 independent trials across 3 test instances. \textbf{m}, acceleration ratios of MEGO compared to GA, HC, and BO. Here, \(f_a\)- \(f_l\) indicate the test instance in fig. a – fig. l, respectively. \textbf{n},  W-D-L counts derived from statistical results of comparing MEGO vs. X (X: GA, HC and BO) under the first stopping criterion. \textbf{o}, W-D-L counts derived from statistical results of comparing MEGO+X vs. X (X: GA, HC and BO) under the second stopping criterion.}
	\label{fig:mego_train_domain}
\end{figure*}

For a comprehensive performance evaluation, the compared optimizers are assessed under two distinct stopping criteria:
\begin{enumerate}
    \item A short budget criterion: the optimizers use the same \#FEs as MEGO, which typically represents the scenario with high demands for solution efficiency.
    \item A long-budget criterion: the optimizers run for a longer time until a budget of 800 FEs is consumed.
\end{enumerate}
To ensure a fair comparison, the same repair procedure is applied to all compared methods under the same evaluation protocol.
Each optimizer is applied 30 times independently on every test instance, and the average results from these runs are compared.
For the ease of comparison, all objective values are normalized and reported, with higher values indicating better performance.
The detailed solution quality results for all compared optimizers are available in the supplementary.
Furthermore, Wilcoxon rank-sum tests with significance level 0.05 are employed to determine the significance of the observed performance differences between the optimizers.

\begin{table*}[tbp]
  \centering
  \caption{\textbf{Comparing MEGO+BO with SMARTEST, a recent optimizer specifically designed for optimizing compiler arguments.} In this application, the goal is to minimize the size of the compiled code. Average ± standard deviation of raw results in terms of code size (bytes) across 30 repeated runs are reported below. On each problem instance, the best quality is indicated by underline ``\_''. Moreover, ``↑, ↓, →'' represents that MEGO+BO is significantly better, worse, or not significantly different than the corresponding optimizer, respectively.}
  \label{tab: result_caop_smartest}
    \begin{tabular}{ccccccc}
    \toprule
    \multicolumn{3}{c}{\multirow{2}[4]{*}{\textbf{Problem Instances}}} & \multicolumn{3}{c}{\textbf{\#FEs=800}} & \textbf{\#FEs=1600} \\
\cmidrule{4-7}    \multicolumn{3}{c}{}  & \textbf{BO} & {\textbf{MEGO+BO}} & \textbf{SMATEST} & \textbf{SMATEST} \\
    \midrule
    \multirow{12}[24]{*}{\makecell{Compiler\\Arguments\\Optimization (CA)}} & \multicolumn{1}{c}{\multirow{3}[6]{*}{Dim=40}} & {ins1} & 5568.27±1.44→ & \underline{5568.00±0.00} & 5569.60±5.99→ & 5569.60±5.99→ \\
\cmidrule{3-7}    &       & {ins2} & \underline{6848.00±0.00→} & \underline{6848.00±0.00} & 6849.33±2.98→ & 6849.33±2.98→ \\
\cmidrule{3-7}    &       & {ins3} & 5672.00±0.00→ & {5672.00±0.00} & 5672.00±0.00→ & 5672.00±0.00→ \\
\cmidrule{2-7}    & \multicolumn{1}{c}{\multirow{3}[6]{*}{Dim=60}} & {ins1} & 6367.20±12.28→ & \underline{6363.73±3.99} & 6366.13±10.47→ & 6365.33±9.98→ \\
\cmidrule{3-7}    &       & {ins2} & 9572.27±3.99↓ & \underline{9569.33±2.98} & 9573.60±6.25↓ & 9573.33±6.31↓ \\
\cmidrule{3-7}    &       & {ins3} & \underline{6264.00±0.00→} & \underline{6264.00±0.00} & 6265.07±3.99→ & 6264.53±2.87→ \\
\cmidrule{2-7}    & \multicolumn{1}{c}{\multirow{3}[6]{*}{Dim=80}} & {ins1} & 5230.93±13.82↓ & \underline{5225.33±4.17} & 5236.00±14.57↓ & 5234.40±14.33↓ \\
\cmidrule{3-7}    &       & {ins2} & 6137.33±2.98→ & \underline{6136.00±0.00} & 6138.40±9.04→ & 6138.40±9.04→ \\
\cmidrule{3-7}    &       & {ins3} & 9134.13±9.39↓ & \underline{9120.27±4.84} & 9132.00±12.18↓ & 9130.13±11.30↓ \\
\cmidrule{2-7}    & \multicolumn{1}{c}{\multirow{3}[6]{*}{Dim=100}} & {ins1} & 5512.53±12.72↓ & \underline{5501.07±7.00} & 5508.80±13.32↓ & 5508.27±13.34↓ \\
\cmidrule{3-7}    &       & {ins2} & 4240.00±12.73↓ & \underline{4213.60±12.76} & 4224.53±19.81↓ & 4222.93±19.13↓ \\
\cmidrule{3-7}    &       & {ins3} & 60583.73±63.98↓ & \underline{60519.73±45.79} & 60618.13±182.73→ & 60610.93±175.61→ \\
    \midrule
    \multicolumn{2}{c}{W-D-L} &       & \textbf{6-6-0} &       & \textbf{5-7-0} & \textbf{5-7-0} \\
    \bottomrule
    \end{tabular}%
\end{table*}%

\subsection{Performance on Entirely Unseen Problem Classes}

The most intriguing findings of this work emerge from the performance of MEGO on real-world problem classes beyond the training set.
Fig.~\ref{fig:mego_test_domain} presents these results.
Figs. 3a-3l show the averaged convergence curves across the three test instances for each problem class and dimension.
The \#FEs consumed by MEGO and the solution quality it achieves are indicated with red lines for easy comparison under the first stopping criterion.
Fig. 3m presents the acceleration ratio (detailed results are provided in the supplementary), calculated as the \#FEs required by baselines to reach MEGO's solution quality divided by MEGO's \#FEs.
Finally, on different problem dimensions, statistical results of comparing MEGO with baselines are presented in Figs. 3n-3o, respectively.

There are five primary findings from Fig.~\ref{fig:mego_test_domain}.
First, Figs. 3a-3l show that MEGO generalizes effectively to the unseen problem classes {\color{black}considered here}.
To the best of our knowledge, this is the first neural optimizer trained on a specific set of binary optimization problem classes generalizes well to entirely unseen binary optimization problem classes.
Furthermore, this generalization extends to problem dimensionality.
MEGO was trained on instances with dimensions up to 40, yet it performs well on test instances with dimensions up to 100. 
Second, MEGO surpasses widely-adopted general-purpose optimizers in both solution quality and computational efficiency {\color{black}on the problem classes considered here.}
Specifically, under the first stopping criterion, MEGO obtains much better solutions than the three baselines when using the same \#FEs.
As shown in Fig. 3m, GA, HC, and BO require on average 3.7, 4.0, and 4.4 times the \#FEs of MEGO, respectively, to match MEGO's solution quality, indicating MEGO is much more efficient than these optimizers.
Third, MEGO serves as an effective performance booster for existing optimizers as an initial solution generator.
Specifically, on the CA class, the W-D-L (win-draw-loss) counts of MEGO+X vs. X (X: GA, HC, or BO) are 4-8-0, 5-7-0, and 6-6-0 when X is GA, HC, and BO, respectively.
On the CIM class, these counts are 9-3-0, 11-1-0, and 12-0-0, respectively, and on the AS class, these counts are 9-3-0, 12-0-0, and 11-1-0, respectively.
Fourth, MEGO is consistently better than the No-Finetuning variant, indicating that target-specific adaptation plays an important role in the effectiveness of MEGO.
Finally, from Fig. 4n and Fig. 4o, the advantage of MEGO under both stopping criteria does not change significantly as the problem dimension varies from 40 to 100, i.e., it is not sensitive to problem dimension.

In summary, these findings have affirmatively answered RQ1, thus indicating MEGO's potential for practical applications.
Particularly, MEGO can be trained on well-studied classic problems and then, without adjustments, {\color{black}applied} to unseen, real-world optimization challenges where training data might be scarce or difficult to obtain.

\subsection{Performance on Problem Classes Seen in the Training Set}
On the classic problem classes present in the training set, MEGO's performance advantage is even more pronounced (Figs. 4a-4l).
Under the first stopping criterion (Fig. 4m), MEGO achieves average acceleration ratios of 3.6, 5.1, and 4.8 compared to GA, HC, and BO, respectively.
With an extended budget of 800 FEs, MEGO+X consistently outperforms the three baselines across all dimensions and problem classes.
For example, on the OM class, MEGO+X achieves a perfect 12-0-0 W-D-L record against all three baselines.
Finally, from Figs. 4n-4o, once again MEGO's advantage is not sensitive to problem dimension.

These results further confirm MEGO's superior performance over the considered general-purpose optimizers and its ability to enhance them, thus affirmatively answering RQ2 {\color{black}within our testbed}.
Collectively, Figs.~\ref{fig:mego_test_domain} and~\ref{fig:mego_train_domain} demonstrate MEGO's promising performance as a general-purpose optimizer applicable to diverse binary optimization problem classes.

\subsection{Comparison with Specialized Optimizer on CA Problem}
Table~\ref{tab: result_caop_smartest} compares MEGO+BO with the recently published specialized optimizer SMARTEST~\cite{jiang2021smartest} on the CA problem class.
From Table~\ref{tab: result_caop_smartest}, MEGO+BO obtains the best solution quality (in terms of code size of the generated executable file) on all 12 problem instances, far better than any other compared optimizer.
Statistical results further show that on any of the problem instances, MEGO+BO is not inferior to SMARTEST, and on five of them, the solution quality obtained by MEGO+BO is significantly better than SMARTEST.
The above findings hold even when the \#FEs consumed by SMARTEST is doubled to 1600.
These results once again demonstrate that MEGO can significantly improve a general-purpose optimizer, in some cases enabling it to substantially surpass the performance of specialized optimizers.

\begin{table}[tbp]
	\centering
	\caption{Performance Comparison Between MEGO and Latent-Space Search Variant via Wilcoxon Rank‑Sum Test~(p=0.05): W~(MEGO Stronger), D~(No Significant Difference), L~(Variant Stronger)}
	\label{tab: wdl_search_space}
    \begin{tabularx}{0.7\linewidth}{l *{3}{>{\centering\arraybackslash}X}}
	\toprule
	\multirow{2}[4]{*}{Problem} & \multicolumn{3}{c}{MEGO VS. Latent-Space Search Variant}\\
	\cmidrule(lr){2-4}
	& W     & D     & L \\
	\midrule
	OM  & 10 & 1 & 1\\
	MC  & 5 & 7 & 0\\
	KP  & 10 & 2 & 0\\
	AS  & 10 & 1 & 1\\
	CIM  & 10 & 1 & 1\\
	CA  & 8 & 4 & 0\\
	Total  & 53 & 16 & 3\\
	\bottomrule
	\end{tabularx}
\end{table}

\subsection{Component Analysis}

\subsubsection{Source-Space Sampling versus Latent-Space Search}
\label{sec:latent_search}
Since prior VAE-based optimization methods often perform search directly in the latent space, we further compared the current solution-generation strategy in MEGO with a latent-space search variant.
In MEGO, candidate solutions are generated by sampling in the source instance's solution space, where $p=2000000$ source-space solutions are randomly sampled (see Sec.~\ref{sec:solution_generation}).
In the latent-space variant, 1024 random latent vectors are first initialized, and gradient descent is then applied for 1954 steps to optimize the latent score predictor, resulting in 2000896 evaluated solutions in total.

The comparison results, in terms of aggregated W-D-L counts for each class, are presented in Table~\ref{tab: wdl_search_space}.
Complete per-instance solution-quality and W-D-L results are provided in the supplementary.
It can be seen that the current source-space sampling strategy outperforms the latent-space search variant on 53 out of 72 test instances, ties on 16 instances, and loses on only 3 instances.
A possible explanation is that the fine-tuning stage in MEGO is designed to establish a mapping from the source solution space to that of the target instance. Under this design, source-space sampling produces genuine source-space candidate solutions, for which the score predictor and the fine-tuned decoder are better aligned. By contrast, direct optimization in the latent space may move to regions that are weakly supported by the source training data, where high predicted scores do not necessarily correspond to truly high-quality source-space solutions. As a result, the decoded target solutions may become less reliable.

\begin{table}[tbp]
\centering
\caption{\textbf{Comparison of Different Padding Strategies in MEGO}. ``↑, ↓, →'' represents that the Padding Strategies is significantly better, worse, or not significantly different than Padding Zero, respectively.}
\label{tab: padding_result}
	\resizebox{\linewidth}{!}{{\begin{tabular}{ccccc}
	\toprule
	\multicolumn{2}{c}{Problem Instance} & Padding Zero & Padding One & Random Padding \\
	\midrule
	\multirow{3}[4]{*}{\makecell{CIM\\Dim=20}} & ins1 & 24.57±0.35 & 24.51±0.32→ & 24.49±0.29→ \\
	\cmidrule{2-5}  & ins2 & 7.78±0.14 & 7.68±0.08↓ & 7.70±0.14↓ \\
	\cmidrule{2-5}  & ins3 & 6.74±0.21 & 7.07±0.32↑ & 7.14±0.30↑ \\
	\midrule
	\multirow{3}[4]{*}{\makecell{CA\\Dim=20}} & ins1 & -375349.87±8.75 & -375350.13±10.26→ & -375352.00±10.93→ \\
	\cmidrule{2-5}  & ins2 & -37166.67±24.62 & -37184.80±23.58↓ & -37189.33±21.30↓ \\
	\cmidrule{2-5}  & ins3 & -527777.33±2.98 & -527782.13±30.12→ & -527783.73±29.97→ \\
	\midrule
	\multirow{3}[4]{*}{\makecell{MC\\Dim=20}} & ins1 & 68.20±0.91 & 68.20±1.01→ & 68.50±0.81→ \\
	\cmidrule{2-5}  & ins2 & 34.03±0.60 & 33.83±0.58→ & 33.87±0.34→ \\
	\cmidrule{2-5}  & ins3 & 50.90±0.30 & 50.87±0.34→ & 51.00±0.00→ \\
	\midrule
	\multirow{3}[4]{*}{\makecell{KP\\Dim=20}} & ins1 & 5.13±0.06 & 5.11±0.04→ & 5.13±0.06→ \\
	\cmidrule{2-5}  & ins2 & 2.81±0.08 & 2.83±0.09→ & 2.85±0.11→ \\
	\cmidrule{2-5}  & ins3 & 6.95±0.05 & 6.95±0.05→ & 6.94±0.06→ \\
	\midrule
	\multirow{3}[4]{*}{\makecell{AS\\Dim=20}} & ins1 & 10226.80±173.64 & 10314.77±258.60→ & 10280.57±285.58→ \\
	\cmidrule{2-5}  & ins2 & 16849.53±194.45 & 16968.60±358.84→ & 17007.37±293.38↑ \\
	\cmidrule{2-5}  & ins3 & 4383.47±138.84 & 4354.57±122.69→ & 4358.57±121.53→ \\
	\midrule
	\multirow{3}[4]{*}{\makecell{OM\\Dim=20}} & ins1 & 18.30±0.46 & 18.20±0.60→ & 18.27±0.44→ \\
	\cmidrule{2-5}  & ins2 & 18.07±0.57 & 17.80±0.60→ & 18.17±0.58→ \\
	\cmidrule{2-5}  & ins3 & 17.50±0.81 & 17.63±0.60→ & 17.67±0.70→ \\
	\midrule
\multicolumn{3}{c}{W-D-L}  & 2-15-1 & 2-14-2\\
	\bottomrule
	\end{tabular}}}
\end{table}

\begin{table}[tbp]
	\centering
	\caption{Performance Comparison between MEGO (``MAPPING'') and A Variant of MEGO (``NO MAPPING'') Across Problem Classes via Wilcoxon Rank‑Sum Test (p=0.05): W~(MEGO Stronger), D~(No Significant Difference), L~(No Fine-Tuning Stronger).}
	\label{tab: wdl_mapping}
	\begin{tabularx}{0.7\linewidth}{l *{3}{>{\centering\arraybackslash}X}}
	\toprule
	\multirow{2}[4]{*}{Problem} & \multicolumn{3}{c}{Mapping VS. No Mapping}\\
	\cmidrule(lr){2-4}
	& W     & D     & L\\
	\midrule
	MC  & 11 & 1 & 0\\
	OM  & 12 & 0 & 0\\
	CIM  & 12 & 0 & 0\\
	CA  & 9 & 2 & 1\\
	AS  & 12 & 0 & 0\\
	KP  & 11 & 1 & 0\\
	Total  & 67 & 4 & 1\\
	\bottomrule
	\end{tabularx}
\end{table}

\begin{table*}[tbp]
	\centering
	\caption{Performance Comparison Between MEGO and Different Expert Identification Strategies via Wilcoxon Rank‑Sum Test~(p=0.05): W~(MEGO Stronger), D~(No Significant Difference), L~(Baseline Stronger).}
	\label{tab: wdl_expert_strategy}
	\begin{tabularx}{0.7\linewidth}{l *{9}{>{\centering\arraybackslash}X}}
		\toprule
		\multirow{2}[4]{*}{Problem} & \multicolumn{3}{c}{MEGO VS. MAX} & \multicolumn{3}{c}{MEGO VS. SAMPLE} & \multicolumn{3}{c}{MEGO VS. MEAN} \\
		\cmidrule(lr){2-4}\cmidrule(lr){5-7}\cmidrule(lr){8-10}
		& W     & D     & L     & W     & D     & L     & W     & D     & L \\
		\midrule
		OM  & 0 & 2 & 10 & 4 & 7 & 1 & 12 & 0 & 0 \\
		MC  & 1 & 3 & 8 & 1 & 11 & 0 & 12 & 0 & 0 \\
		KP  & 2 & 3 & 7 & 3 & 9 & 0 & 12 & 0 & 0 \\
		AS  & 1 & 1 & 10 & 1 & 9 & 2 & 11 & 0 & 1 \\
		CIM & 3 & 3 & 6 & 1 & 9 & 2 & 12 & 0 & 0 \\
		CA  & 1 & 7 & 4 & 7 & 4 & 1 & 12 & 0 & 0 \\
		Total  & 8 & 19 & 45 & 17 & 49 & 6 & 71 & 0 & 1 \\
		\bottomrule
	\end{tabularx}
\end{table*}

\begin{figure*}[tbp]
	\centering
\includegraphics[width=0.9\linewidth]{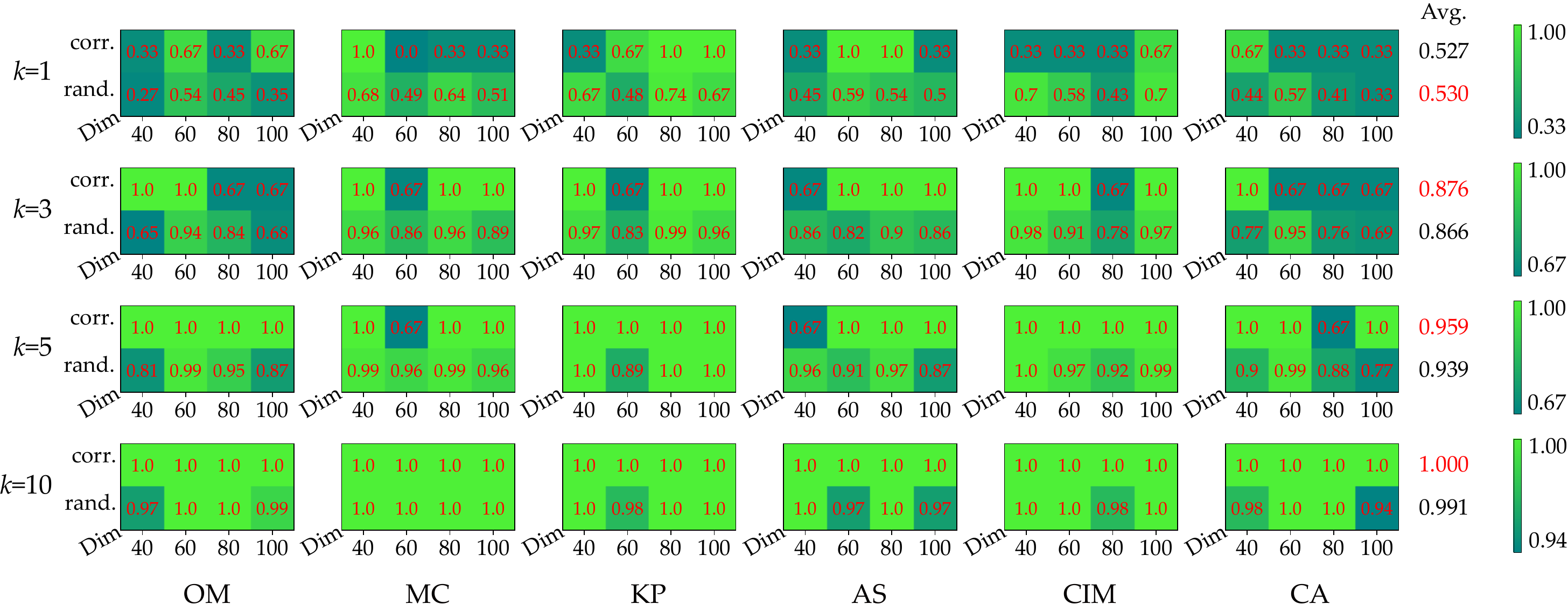}
	\caption{Rates of hitting top-\(k\)~(\(k\)=1, 3, 5, and 10) expert models, achieved by correlation-based selection (corr) and random selection (rand).}
	\label{fig: heatmap}
\end{figure*}
\label{sec:component}
\subsubsection{Padding Strategy}
To examine whether zero padding would introduce bias during the expert identification step of MEGO (see Sec.~\ref{sec:expert_selection}), we compared three padding strategies: zero padding, all-one padding, and random padding.
To ensure that padding is indeed required, we used test instances of dimension 20, while the experts were trained on dimensions 30, 35, and 40.
Each strategy was executed for 30 independent runs on every problem instance.
The test results are presented in Table~\ref{tab: padding_result}.
Across the 18 test instances, zero padding achieves W-D-L counts of 2-15-1 against all-one padding and 2-14-2 against random padding.
These results indicate that all the three strategies are broadly comparable to each other, and no clear systematic bias is observed.

\subsubsection{Source-Target Mapping}
To investigate the effect of the source-target mapping step (line 16 in Alg.~\ref{alg:apply_mego}), we further compared the following two configurations of MEGO.
\begin{itemize}
	\item \textbf{Mapping}: The standard MEGO procedure.
	\item \textbf{No Mapping}: A variant of MEGO where the solution mapping step is removed.
	After experts are selected, new solutions are sampled directly in the source problem instance's solution space.
	Then, these solutions are ranked by their performance on the expert models and the top-\(k\) solutions are selected, and are finally evaluated on the target problem instance.
\end{itemize}
Each method was executed for 30 independent runs on every problem instance.
The test results,  in terms of aggregated W-D-L counts (according to the Wilcoxon rank-sum test at a 0.05 significance level) for each class, are presented in Table~\ref{tab: wdl_mapping}.
Complete per-instance solution-quality and W-D-L results are provided in the supplementary.
Overall, these results show that MEGO (with mapping) outperforms the no-mapping variant on 67 out of 72 test instances, ties on 4 instances, and loses on only 1 instance.
These results indicate that the mapping step is essential for the effectiveness of MEGO.

\subsubsection{Relevant Expert Identification}
The current implementation of MEGO would normally activate multiple experts rather than only one (MEGO selects about 14 experts on average across the experiments).
A natural question is whether activating multiple experts is beneficial, since different experts may contain overlapping knowledge.
To examine the effect of this design, we compared MEGO with three expert-selection baselines:
(i) random selection of the same number of experts as MEGO (denoted as ``SAMPLE''), 
(ii) the average performance obtained by running each expert individually (``MEAN'') that corresponds to randomly selecting only one expert, 
and (iii) the hindsight best individual expert among all experts (``MAX'').
Table~\ref{tab: wdl_expert_strategy} reports aggregated W-D-L results  (according to the Wilcoxon rank-sum test at a 0.05 significance level) at the problem-class level.
Complete results are provided in the supplementary material.

Across all 72 test instances, MEGO achieves W-D-L counts of 19-45-8 against SAMPLE, 71-0-1 against MEAN, and 9-15-48 against MAX. 
These results suggest that activating multiple experts is indeed beneficial, as MEGO substantially outperforms the MEAN baseline (relying on a single randomly chosen expert), and that the correlation-based routing strategy is more effective than selecting the same number of experts at random.
At the same time, it is not surprising that the hindsight best single expert often outperforms MEGO, because that baseline assumes oracle knowledge after exhaustively evaluating all experts.
Overall, these results suggest that overlap among experts does not negate the value of multi-expert activation.
Rather, what matters is activating a relevant subset of experts for the target instance. 

To further examine whether the correlation-based routing policy can identify useful experts, we compared it with random selection in terms of the rates of hitting top-\(k\) relevant experts, where \(k=1,3,5,10\).
The results in Figure~\ref{fig: heatmap} show that using correlation coefficients generally leads to higher average hit rates than random selection, especially for top-3, top-5, and top-10.
This indicates that the routing policy is effective in identifying relevant experts for a target instance.

\subsubsection{Analysis of the Latent Representation }
To further examine whether the latent representations could preserve global structural information, we performed a quantitative analysis of reconstruction fidelity and geometry preservation for the expert models.
Specifically, for an expert model trained on a training instance, we randomly generate 10000 binary solutions \(X\), encode them into the latent space, and then decode them to obtain reconstructed solutions \(X'\).
These input solutions are generated independently of the training and validation sets.
The mean absolute error between \(X\) and \(X'\) is only 0.006, corresponding to a per-dimension reconstruction accuracy of 99.4\% for binary vectors.
In addition, the exact reconstruction accuracy at the whole-solution level reaches 84\%.

\begin{figure*}[tbp]
	\centering
	\includegraphics[width=0.7\linewidth]{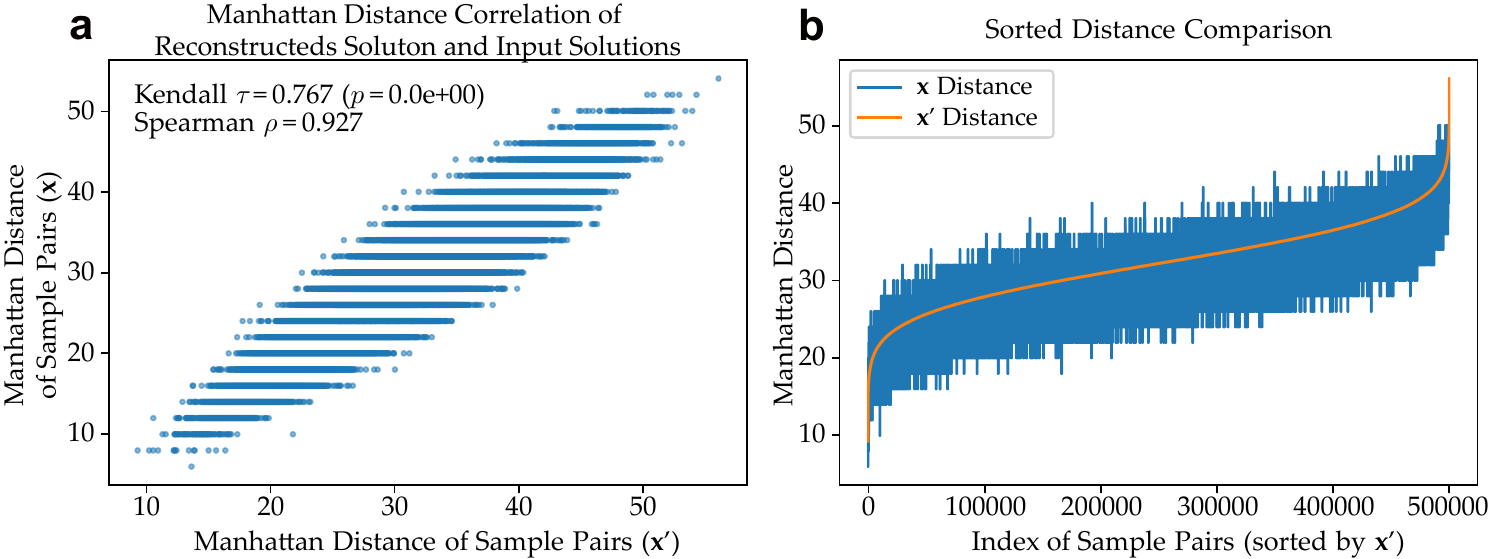}
	\caption{Correspondence between Manhattan Distances of Reconstructed Solution Pairs and their respective Input Pairs. (a) Distance Correlation. (b) Sorted Distance Progression.}
	\label{fig: vae_distance_analysis}
\end{figure*}

We further analyze whether relative relationships among solutions are preserved after passing through the latent representation.
Specifically, we compute the pairwise Manhattan distances among solutions in \(X\) and among reconstructed solutions in \(X'\), and then measure the rank correlation between these two distance matrices.
The results are illustrated in Fig.~\ref{fig: vae_distance_analysis}.
The resulting Kendall's Tau and Spearman correlation coefficients are 0.767 and 0.927, respectively.

These results suggest that the latent representations preserve substantial structural information about the original solution space, both at the level of individual solution reconstruction and at the level of geometric relationships among solutions.

\subsection{A Novel Perspective of Problem Classification}
\label{sec:new_classification}
Based on MEGO, a computational approach is developed to classify optimization problems based on their vector-based representations.
This approach results in a novel, data-driven perspective that diverges from conventional, analysis-based classification.

Specifically, each test instance \(I_{new}\) is represented as a \(27\)-dimensional vector, with the \(j\)-th component quantifies the similarity between $I_{new}$ and the \(j\)-th training problem instance (there are 27 training problem instances in total).
To obtain this vector, the following procedure is applied.
First, each of the 27 expert models in MEGO is adapted to $I_{new}$, using the fine-tuning procedure described in Sec.~\ref{sec:apply_mego}.
Once fine-tuned, the encoder-decoder of each expert model is a mapping from the solution space of the corresponding training problem instance to the solution space of $I_{new}$.
Based on each of these mappings, a large number of solutions are randomly sampled and the four solutions with the highest predicted scores are identified, resulting in a candidate pool of 108 solutions.
These solutions are then evaluated using the objective function of $I_{new}$, and the top four unique solutions from this pool are retained.
This step validates which expert models are genuinely most effective at generating high-quality solutions for $I_{new}$.

The above procedure is repeated for 30 times; therefore a final elite set of 120 solutions (4 solutions × 30 repeat) is collected.
The vector is then derived from this elite set.
A raw 27-dimensional count vector, $\mathbf{a}'$, is first constructed, where its $j$-th component, $a_j$, is simply the number of solutions in the 120-solution elite set that originated from the $j$-th expert model.
This count directly measures the contribution, and thus the relevance, of the $j$-th training problem instance to solving $I_{new}$.
Finally, to ensure the components are on a consistent scale for comparison and clustering, the raw count vector is L2-normalized to produce the final vector: $\mathbf{a}=\frac{\mathbf{a}'}{||\mathbf{a}||_2}$.

\begin{figure*}[htbp]
	\centering
	\includegraphics[width=1.0\textwidth]{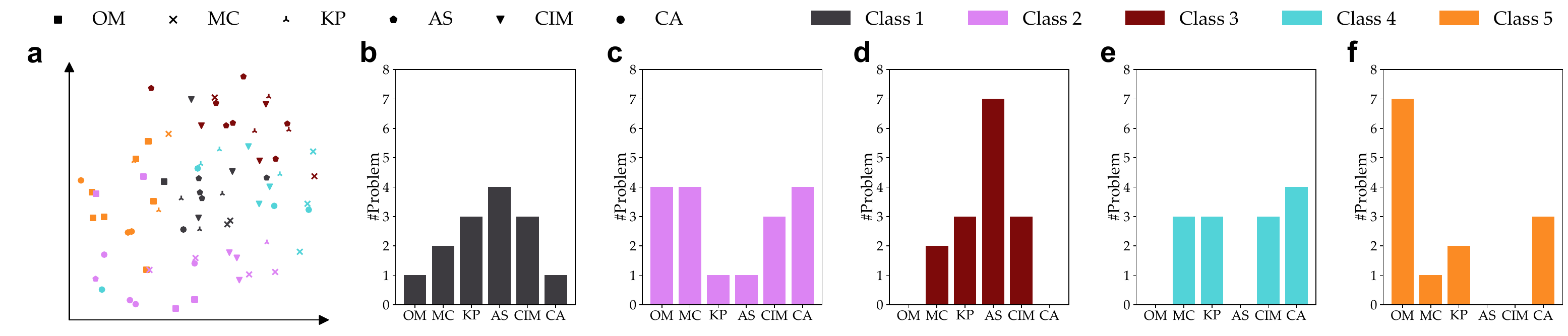}
	\caption{\textbf{Visualization of conventional problem classification and similarity-based problem classification}. \textbf{a}. clustering and 2-dimensional t-SNE visualization results of the test instances. \textbf{b-f}. distributions of instances from different problem classes (according to conventional classification) across different clusters (according to similarity-based classification).}
	\label{fig:mego_vis}
\end{figure*}

Fig.~\ref{fig:mego_vis} illustrates the clustering (using K-means~\cite{lloyd1982least}) and 2-dimensional t-SNE visualization~\cite{van2008visualizing} of the vectors of all 72 test instances (5 clusters).
The resulting classification reveals interesting discrepancies from conventional problem classification.
While some consistency exists (e.g., between OM class and cluster 5, AS class and cluster 3), other problem classes show substantial distribution across multiple clusters.
For instance, the CA and CIM classes are each spread across 4 clusters.
These findings are somewhat counterintuitive.
They indicate that from the perspective of a learned optimizer like MEGO, instances from conventionally different problem classes may exhibit greater similarity than instances within the same problem class.
This observation also suggests that the behavior of MEGO is not determined solely by the manually defined training problem classes.
This insight offers an explanation for MEGO's strong generalization to unseen problem classes.
That is, in the cases studied here, MEGO succeeds not by learning the abstract properties of a problem class, but by identifying that a new instance---regardless of its conventional class label---{\color{black}shares sufficiently aligned solution-quality landscape structure} with some instances in its experience, thereby enabling the transfer of a relevant optimization strategy.

Overall, MEGO introduces a computational lens for examining the relationships between seemingly disparate optimization problems.
This data-driven perspective on problem similarity complements conventional, analysis-based classifications and may offer new insights into the problem landscape.

\section{Conclusion and Discussion}
\label{sec:conclusion}
This work addresses the challenge of building off-the-shelf optimizers for binary optimization.
MEGO, a general-purpose neural optimizer, is introduced.
With minimal problem-specific customization, MEGO shows strong generalization capabilities across the six problem classes considered in this work, including those not seen during training.
Furthermore, MEGO offers a new computational approach for classifying optimization problems, which is fundamentally different from the conventional problem classification approach.
{\color{black}
At the same time, it is important to note that MEGO should not be understood as a universal optimizer for arbitrary binary optimization problems.
Its effectiveness depends on whether the target problem instance shares reusable structure (regularities) with the expert pool.
To further examine this point, we conducted an additional experiment in which the target problem instance is an arbitrary random binary function, and compared MEGO with the baselines under the same \#FEs over 30 independent runs.
The results show that MEGO no longer exhibits a meaningful transfer advantage in that case.
This supports the view that when such reusable structure is weak or absent, transfer is no longer meaningful and negative transfer may occur.
Detailed discussion of these regularities and the additional experiment is provided in the supplementary material.}

There exist many open avenues for future improvement.
	For instance, the current MEGO architecture is intentionally modular rather than fully end-to-end.
	Each expert model is trained independently on one training instance, the routing policy is based on sampled evaluations and correlation-based expert selection, and the selected experts are subsequently adapted to the target instance through decoder fine-tuning.
	As a result, the overall pipeline is better viewed as a select-adapt-generate framework than a conventional end-to-end MoE.

	This design has both advantages and limitations.
	On the one hand, it makes MEGO modular and flexible, allows experts to be learned independently from black-box experience data, and avoids requiring a differentiable routing mechanism tied to a specific problem representation.
	It also naturally supports target-instance-specific adaptation after expert selection.
	On the other hand, because the router and the experts are not jointly optimized, they may not be perfectly synchronized.
	Consequently, the routing rule may fail to activate the most useful experts in some cases, or may activate suboptimal ones.
	
	Developing a more tightly integrated version of MEGO would require a substantially different framework, since the current routing procedure is non-differentiable and the adaptation step depends on target-instance-specific sampled data.
	Future work may therefore investigate differentiable routing, meta-level adaptation across problem instances, or reinforcement-learning-based routing and training strategies.

Furthermore, MEGO does not directly handle constraints.
While it can be paired with simple repair heuristics to solve constrained problems, a notable advancement would be the integration of a built-in, general-purpose constraint-handling mechanism.
Beyond that, several other exciting research directions are envisioned:
\begin{enumerate}
    \item The evaluation of MEGO can be extended to a wider array of problem classes and dimensions.
    	In particular, scaling MEGO to much larger binary dimensions is an important direction for future work, but will likely require improved expert architectures and more efficient adaptation mechanisms for high-dimensional settings.
    \item Another important direction is to investigate more principled ways of automatically constructing the training problem classes and expert pool, thereby mitigating the potential bias of manual selection.
    \item The current solution-generation step uses random sampling for simplicity and efficiency.
    A promising direction is to investigate more sophisticated search strategies in the latent space of selected experts, which may further improve solution quality.
	\item A continuous learning framework could be explored.
	This would allow MEGO to continuously improve by learning from new problem instances encountered after its initial training, but it would also enlarge the expert pool over time.
	In the current implementation, the storage cost grows linearly with the number of experts, with each expert occupying about 5.9 MB.
	At inference time, not all experts are activated for every target instance, and fine-tuning and solution generation for different selected experts are naturally parallelizable.
	The GPU memory usage for one expert is about 750 MB, and on an 8-GPU server with Nvidia MPS or MIG enabled, we can fine-tune and sample from at least 24 experts concurrently with little observed additional latency.
	A key challenge for future work is therefore to manage the growing expert pool efficiently, for example through dynamic model pruning, so as to maintain a balance between performance and computational cost.

    \item Finally, the problem classification approach enable by MEGO could be explored further.
    A valuable next step would be to analyze the structural properties of the resulting problem clusters.
    For instance, by using fitness landscape analysis, one could examine whether the problem instances grouped together share underlying features, such as ruggedness or variable interaction.
    Such an understanding could then be used to develop more suitable optimization strategy for these problem instances.
\end{enumerate}

\section*{Acknowledgments}
This work was supported by National Key Research and Development Program of China under Grant 2022YFA1004102, and in part by the Natural Science Foundation of China under Grants T2495254, 62502192, and 62250710682, and in part by the internal grants of Lingnan University.

\bibliographystyle{IEEEtran}
\bibliography{MEGO}

\begin{IEEEbiographynophoto}{Shengcai Liu}
	\textbf{(Member, IEEE)} received the B.Sc. degree in 2014 and the Ph.D. degree in 2020 from the University of Science and Technology of China, Hefei, China.
	He is currently an Assistant Professor with the Department of Computer Science and Engineering, Southern University of Science and Technology, Shenzhen, China.
	He is mainly interested in Learning to Optimize and has published more than 30 papers in top-tier refereed international conferences and journals.
\end{IEEEbiographynophoto}

\begin{IEEEbiographynophoto}{Zhiyuan Wang}
	\textbf{(Student Member, IEEE)} received the B.Eng. degree in 2020 and the M.Eng. degree in 2023 from the Southern University of Science and Technology (SUSTech), Shenzhen, China. He is currently working toward a Ph.D. 
	degree in the Department of Computer Science and Engineering at SUSTech. His research interests include black-box optimization and machine learning.
\end{IEEEbiographynophoto}

\begin{IEEEbiographynophoto}{Yew-Soon Ong}
	\textbf{(Fellow, IEEE)} received the Ph.D. degree in artificial intelligence in complex design from the University of Southampton, Southampton, U.K., in 2003. He is the Yidan Endowed Professor of Artificial Intelligence at the College of Computing and Data Science, Nanyang Technological University, Singapore, and a Distinguished AI Fellow at Singapore's Agency for Science, Technology and Research. His research interests are in artificial and computational intelligence. Dr. Ong is the Founding Editor-in-Chief of the IEEE Transactions on Emerging Topics in Computational Intelligence, Associate Editor of several IEEE Transactions, and Senior or Area Chair for leading AI conferences, including ICML, ICLR, NeurIPS, AAAI, and IJCAI. He has received several IEEE Outstanding Paper Awards and has been recognized as a Thomson Reuters Highly Cited Researcher and among the World's Most Influential Scientific Minds. He is a Fellow of the IEEE, a Fellow of the Academy of Engineering Singapore, and an IEEE Computational Intelligence Society Distinguished Lecturer.
\end{IEEEbiographynophoto}

\begin{IEEEbiographynophoto}{Xin Yao}
	\textbf{(Fellow, IEEE)} obtained his Ph.D. degree in 1990 from the University of Science and Technology of China (USTC), Hefei, China; M.Sc. in 1985 from North China Institute of Computing Technologies, Beijing, China; and B.Sc. in 1982 from USTC. He is the Vice President (Research and Innovation) and Tong Tin Sun Chair Professor of Machine Learning at Lingnan University, Hong Kong, China. He is a Fellow of IEEE and Hong Kong Academy of Engineering. He served as the President (2014-15) of IEEE Computational Intelligence Society (CIS) and the Editor-in-Chief (2003-08) of IEEE Transactions on Evolutionary Computation. His major research interests include evolutionary computation, machine learning and AI ethics. His work won the 2001 IEEE Donald G. Fink Prize Paper Award; 2010, 2016 and 2017 IEEE Transactions on Evolutionary Computation Outstanding Paper Awards; 2011 IEEE Transactions on Neural Networks Outstanding Paper Award; 2010 BT Gordon Radley Award for Best Author of Innovation (Finalist); and other best paper awards at conferences. He received the 2012 Royal Society Wolfson Research Merit Award, 2013 IEEE CIS Evolutionary Computation Pioneer Award, and the 2020 IEEE Frank Rosenblatt Award.
\end{IEEEbiographynophoto}

\begin{IEEEbiographynophoto}{Ke Tang}
	\textbf{(Fellow, IEEE)} received the Ph.D. degree in computer science from Nanyang Technological University, Singapore, in 2007. He is currently a chair professor at the Department of Computer Science and Engineering, at Southern University of Science and Technology (SUSTech), Shenzhen, China. His research interests mainly include evolutionary computation, machine learning, and their applications. He is a Fellow of IEEE and he was awarded the Newton Advanced Fellowship (Royal Society) and the Changjiang Professorship Ministry of Education (MOE) of China. He was the recipient of the IEEE Computational Intelligence Society Outstanding Early Career Award and the Natural Science Award of MOE of China. He has been actively involved in the research community by serving as an organizer/co-organizer of numerous conferences, an associate editor of the IEEE TRANSACTIONS ON EVOLUTIONARY COMPUTATION, and as a member of editorial boards for a few other journals.
\end{IEEEbiographynophoto}

\clearpage
\def\supplementaryinput{}
\ifdefined\supplementaryinput
\else
\documentclass[journal]{IEEEtran}
\usepackage{amsmath,amsfonts}
\usepackage{algorithmic}
\usepackage{array}
\usepackage[caption=false,font=normalsize,labelfont=sf,textfont=sf]{subfig}
\usepackage{textcomp}
\usepackage{stfloats}
\usepackage{url}
\usepackage{hyperref}
\usepackage{verbatim}
\usepackage{graphicx}
\usepackage{cite}
\usepackage{multirow}
\usepackage{makecell}
\usepackage{booktabs}
\usepackage{tabularx}
\usepackage{amssymb}
\usepackage{ulem}
\usepackage[table, dvipsnames]{xcolor}
\usepackage[ruled,linesnumbered]{algorithm2e}
\hyphenation{op-tical net-works semi-conduc-tor IEEE-Xplore}
\def\BibTeX{{\rm B\kern-.05em{\sc i\kern-.025em b}\kern-.08em
		T\kern-.1667em\lower.7ex\hbox{E}\kern-.125emX}}
\usepackage{balance}

\begin{document}
\fi
	\normalem
	\ifdefined\supplementaryinput
		\twocolumn[
		\begin{center}
			{\LARGE\bfseries Supplementary for the Paper \\
			``MEGO: Learning Mixture-of-Experts for General-Purpose Binary Optimization''\par}
		\end{center}
		\vspace{1em}
		]
	\else
		\title{Supplementary for the Paper \\``MEGO: Learning Mixture-of-Experts for General-Purpose Binary Optimization''}

		\author{}

		
		\maketitle
	\fi

	\ifdefined\supplementaryinput
		\renewcommand{\theHsection}{supplementary.\arabic{section}}
		\renewcommand{\theHsubsection}{supplementary.\arabic{section}.\arabic{subsection}}
		\renewcommand{\theHsubsubsection}{supplementary.\arabic{section}.\arabic{subsection}.\arabic{subsubsection}}
		\renewcommand{\theHtable}{supplementary.\arabic{table}}
		\renewcommand{\theHfigure}{supplementary.\arabic{figure}}
		\renewcommand{\theHequation}{supplementary.\arabic{equation}}
	\fi
	\setcounter{section}{0}
	\setcounter{subsection}{0}
	\setcounter{subsubsection}{0}
	\setcounter{equation}{0}
	\setcounter{table}{2}
	\setcounter{figure}{5}

	\section{Detailed Experimental Results on Problem Classes Seen in the Training Set}
	\subsection{Objective Values of MEGO and the Baselines}
	We compare the objective values achieved by MEGO and the baselines on the training problem classes, while the baselines use the same \#FEs as MEGO.
	Each problem instance is optimized 30 times.
	The results' statistical significance was tested through the Wilcoxon rank-sum statistic.
	Table~\ref{tab: result_om},~\ref{tab: result_kp},~\ref{tab: result_mc} compare the objective values achieved by MEGO and baselines on the OM problem, KP, and MC problem, respectively.
	
\begin{table*}[htbp]
  \centering
  \caption{\textbf{Objective values achieved by MEGO and the Baselines on the OM Problem.} ``↑, ↓, →'' represents that the corresponding method is significantly better, worse, or not significantly different than the MEGO, respectively.}
  \label{tab: result_om}
    {\begin{tabular}{cccccc}
    \toprule
    \multicolumn{2}{p{8.47em}}{Problem Instance} & \multicolumn{1}{c}{MEGO} & GA    & HC    & BO \\
    \midrule
    \multirow{3}[6]{*}{Dim=40} & \multicolumn{1}{c}{ins1} & \multicolumn{1}{c}{31.33±1.01} & 27.40±1.60↓ & 23.30±3.23↓ & 26.50±1.69↓ \\
\cmidrule{2-6}     & \multicolumn{1}{c}{ins2} & \multicolumn{1}{c}{30.83±0.93} & 27.57±1.78↓ & 22.57±3.34↓ & 26.70±1.29↓ \\
\cmidrule{2-6}     & \multicolumn{1}{c}{ins3} & \multicolumn{1}{c}{32.53±1.20} & 27.53±1.52↓ & 22.53±3.31↓ & 26.50±1.48↓ \\
    \midrule
    \multirow{3}[6]{*}{Dim=60} & \multicolumn{1}{c}{ins1} & \multicolumn{1}{c}{44.87±1.18} & 39.63±1.83↓ & 32.57±3.45↓ & 38.73±1.44↓ \\
\cmidrule{2-6}     & \multicolumn{1}{c}{ins2} & \multicolumn{1}{c}{42.90±0.75} & 39.53±1.54↓ & 32.73±3.62↓ & 38.77±1.86↓ \\
\cmidrule{2-6}     & \multicolumn{1}{c}{ins3} & \multicolumn{1}{c}{42.63±1.58} & 39.27±1.61↓ & 31.03±3.49↓ & 38.47±1.93↓ \\
    \midrule
    \multirow{3}[6]{*}{Dim=80} & \multicolumn{1}{c}{ins1} & \multicolumn{1}{c}{54.37±2.36} & 50.23±2.29↓ & 41.83±3.64↓ & 49.40±1.94↓ \\
\cmidrule{2-6}     & \multicolumn{1}{c}{ins2} & \multicolumn{1}{c}{57.03±1.43} & 50.20±2.07↓ & 42.70±4.66↓ & 50.07±2.06↓ \\
\cmidrule{2-6}     & \multicolumn{1}{c}{ins3} & \multicolumn{1}{c}{56.13±1.48} & 50.90±2.45↓ & 43.47±4.83↓ & 49.17±1.44↓ \\
    \midrule
    \multirow{3}[6]{*}{Dim=100} & \multicolumn{1}{c}{ins1} & \multicolumn{1}{c}{69.90±2.44} & 61.53±2.00↓ & 51.07±4.25↓ & 60.30±2.10↓ \\
\cmidrule{2-6}     & \multicolumn{1}{c}{ins2} & \multicolumn{1}{c}{66.67±2.65} & 61.27±3.00↓ & 50.50±3.96↓ & 60.10±1.94↓ \\
\cmidrule{2-6}     & \multicolumn{1}{c}{ins3} & \multicolumn{1}{c}{64.10±1.51} & 61.50±2.01↓ & 50.63±4.85↓ & 60.17±1.93↓ \\
    \midrule
    \multicolumn{3}{c}{W-D-L} & 12-0-0 & 12-0-0 & 12-0-0 \\
    \bottomrule
    \end{tabular}}
\end{table*}%

\begin{table*}[htbp]
  \centering
  \caption{\textbf{Objective values achieved by MEGO with the Baselines on the KP.} ``↑, ↓, →'' represents that the corresponding method is significantly better, worse, or not significantly different than the MEGO, respectively. }
  \label{tab: result_kp}
    {\begin{tabular}{cccccc}
    \toprule
    \multicolumn{2}{c}{Problem Instance} & \multicolumn{1}{c}{MEGO} & GA    & HC    & BO \\
    \midrule
    \multirow{3}[6]{*}{Dim=40} & \multicolumn{1}{c}{ins1} & \multicolumn{1}{c}{7.53±0.09} & 7.23±0.13 ↓ & 6.92±0.25 ↓ & 7.15±0.12 ↓ \\
\cmidrule{2-6}     & \multicolumn{1}{c}{ins2} & \multicolumn{1}{c}{13.13±0.10} & 12.97±0.16 ↓ & 12.57±0.66 ↓ & 12.89±0.17 ↓ \\
\cmidrule{2-6}     & \multicolumn{1}{c}{ins3} & \multicolumn{1}{c}{5.91±0.03} & 5.74±0.08 ↓ & 5.63±0.17 ↓ & 5.65±0.08 ↓ \\
    \midrule
    \multirow{3}[6]{*}{Dim=60} & \multicolumn{1}{c}{ins1} & \multicolumn{1}{c}{8.73±0.09} & 8.72±0.10→ & 8.53±0.18 ↓ & 8.61±0.09 ↓ \\
\cmidrule{2-6}     & \multicolumn{1}{c}{ins2} & \multicolumn{1}{c}{7.41±0.09} & 7.36±0.09 ↓ & 7.17±0.15 ↓ & 7.26±0.13 ↓ \\
\cmidrule{2-6}     & \multicolumn{1}{c}{ins3} & \multicolumn{1}{c}{12.96±0.07} & 12.90±0.07 ↓ & 12.72±0.16 ↓ & 12.86±0.08 ↓ \\
    \midrule
    \multirow{3}[6]{*}{Dim=80} & \multicolumn{1}{c}{ins1} & \multicolumn{1}{c}{19.18±0.07} & 19.07±0.11 ↓ & 18.69±0.40 ↓ & 19.00±0.10 ↓ \\
\cmidrule{2-6}     & \multicolumn{1}{c}{ins2} & \multicolumn{1}{c}{25.89±0.76} & 24.17±1.13 ↓ & 21.07±2.37 ↓ & 23.93±1.19 ↓ \\
\cmidrule{2-6}     & \multicolumn{1}{c}{ins3} & \multicolumn{1}{c}{23.17±0.11} & 23.08±0.15→ & 21.15±1.95 ↓ & 22.93±0.19 ↓ \\
    \midrule
    \multirow{3}[6]{*}{Dim=100} & \multicolumn{1}{c}{ins1} & \multicolumn{1}{c}{18.35±0.03} & 18.37±0.05→ & 18.23±0.14 ↓ & 18.30±0.06 ↓ \\
\cmidrule{2-6}     & \multicolumn{1}{c}{ins2} & \multicolumn{1}{c}{36.55±0.93} & 33.79±1.13 ↓ & 28.18±2.70 ↓ & 33.53±1.08 ↓ \\
\cmidrule{2-6}     & \multicolumn{1}{c}{ins3} & \multicolumn{1}{c}{15.27±0.05} & 15.21±0.07 ↓ & 15.08±0.14 ↓ & 15.16±0.07 ↓ \\
    \midrule
    \multicolumn{3}{c}{W-D-L} & 9-3-0 & 12-0-0 & 12-0-0 \\
    \bottomrule
    \end{tabular}}
\end{table*}%

\begin{table*}[htbp]
  \centering
  \caption{\textbf{Objective values achieved by MEGO and the Baselines on the MC Problem.} ``↑, ↓, →'' represents that the corresponding method is significantly better, worse, or not significantly different than the MEGO, respectively.}
  \label{tab: result_mc}
    {\begin{tabular}{cccccc}
    \toprule
    \multicolumn{2}{c}{Problem Instance} & \multicolumn{1}{c}{MEGO} & \multicolumn{1}{c}{GA} & HC    & BO \\
    \midrule
    \multirow{3}[6]{*}{Dim=40} & \multicolumn{1}{c}{ins1} & \multicolumn{1}{c}{186.03±4.09} & \multicolumn{1}{c}{181.60±3.52↓} & 180.73±3.68↓ & 178.43±2.25↓ \\
\cmidrule{2-6}     & \multicolumn{1}{c}{ins2} & \multicolumn{1}{c}{264.83±2.15} & \multicolumn{1}{c}{264.70±3.21→} & 265.13±3.17→ & 263.13±2.87↓ \\
\cmidrule{2-6}     & \multicolumn{1}{c}{ins3} & \multicolumn{1}{c}{138.13±1.15} & \multicolumn{1}{c}{135.83±2.15↓} & 135.63±2.32↓ & 135.10±2.17↓ \\
    \midrule
    \multirow{3}[6]{*}{Dim=60} & \multicolumn{1}{c}{ins1} & \multicolumn{1}{c}{442.17±2.02} & \multicolumn{1}{c}{439.60±3.78↓} & 435.80±5.20↓ & 439.30±3.48↓ \\
\cmidrule{2-6}     & \multicolumn{1}{c}{ins2} & \multicolumn{1}{c}{622.20±2.21} & \multicolumn{1}{c}{622.23±3.93→} & 613.77±6.78↓ & 619.17±3.34↓ \\
\cmidrule{2-6}     & \multicolumn{1}{c}{ins3} & \multicolumn{1}{c}{465.50±2.42} & \multicolumn{1}{c}{462.60±4.01↓} & 454.83±7.08↓ & 460.20±4.09↓ \\
    \midrule
    \multirow{3}[6]{*}{Dim=80} & \multicolumn{1}{c}{ins1} & \multicolumn{1}{c}{1048.43±6.01} & \multicolumn{1}{c}{1041.27±4.39↓} & 1028.47±11.24↓ & 1039.13±5.85↓ \\
\cmidrule{2-6}     & \multicolumn{1}{c}{ins2} & \multicolumn{1}{c}{1068.33±5.06} & \multicolumn{1}{c}{1062.17±5.09↓} & 1050.97±11.41↓ & 1059.17±4.31↓ \\
\cmidrule{2-6}     & \multicolumn{1}{c}{ins3} & \multicolumn{1}{c}{709.77±5.98} & \multicolumn{1}{c}{700.30±6.00↓} & 688.43±8.81↓ & 698.20±5.96↓ \\
    \midrule
    \multirow{3}[6]{*}{Dim=100} & \multicolumn{1}{c}{ins1} & \multicolumn{1}{c}{1430.50±6.23} & \multicolumn{1}{c}{1427.77±8.34↓} & 1398.80±15.67↓ & 1418.60±8.17↓ \\
\cmidrule{2-6}     & \multicolumn{1}{c}{ins2} & \multicolumn{1}{c}{1111.57±7.10} & \multicolumn{1}{c}{1113.80±7.45↑} & 1091.43±16.70↓ & 1104.33±5.58↓ \\
\cmidrule{2-6}     & \multicolumn{1}{c}{ins3} & \multicolumn{1}{c}{1381.30±6.27} & \multicolumn{1}{c}{1366.83±8.86↓} & 1350.30±12.91↓ & 1363.00±5.13↓ \\
    \midrule
    \multicolumn{3}{c}{W-D-L} & 9-2-1 & 11-1-0 & 12-0-0 \\
    \bottomrule
    \end{tabular}}
\end{table*}%

	\subsection{Acceleration Ratios of MEGO compared to the Baselines}
	
	We record the \#FEs required by the baselines to achieve the objective values of MEGO, and based on this, calculate the acceleration ratio of MEGO relative to baselines across various problem classes and various problem dimensions.
	The \#FEs used for a specific problem class with a particular dimension were averaged over 30 runs on the three test instances.
	The comparison results are shown in Table~\ref{tab: acc_train}.
	
\begin{table*}[htbp]
  \centering
  \caption{Acceleration Ratios of MEGO compared to the Baselines on the problem classes seen in the training set.}
  \label{tab: acc_train}
    \begin{tabular}{ccccccccc}
    \toprule
    \multirow{2}[4]{*}{Problem Class} & \multicolumn{1}{c}{\multirow{2}[4]{*}{Dim}} & \multicolumn{1}{c}{MEGO} & \multicolumn{2}{c}{GA} & \multicolumn{2}{c}{HC} & \multicolumn{2}{c}{BO} \\
\cmidrule(lr){3-3}\cmidrule(lr){4-5}\cmidrule(lr){6-7}\cmidrule(lr){8-9}
&       & \multicolumn{1}{c}{\#FEs} & \multicolumn{1}{c}{\#FEs} & \multicolumn{1}{c}{Ratio} & \multicolumn{1}{c}{\#FEs} & \multicolumn{1}{c}{Ratio} & \multicolumn{1}{c}{\#FEs} & \multicolumn{1}{c}{Ratio} \\
    \midrule
    \multirow{4}[8]{*}{OM} & 40    & 94.28 & 540.36 & 5.73  & 440.98 & 4.68  & 660.24 & 7 \\
\cmidrule{2-9}     & 60    & 127.1 & 495.21 & 3.9   & 735.14 & 5.78  & 598.69 & 4.71 \\
\cmidrule{2-9}     & 80    & 106.68 & 499.29 & 4.68  & 830.94 & 7.79  & 662.28 & 6.21 \\
\cmidrule{2-9}     & 100   & 94.97 & 466.58 & 4.91  & 884.12 & 9.31  & 621.62 & 6.55 \\
    \midrule
    \multirow{4}[8]{*}{KP} & 40    & 132.76 & 583.12 & 4.39  & 685.64 & 5.16  & 701.36 & 5.28 \\
\cmidrule{2-9}     & 60    & 119.93 & 324.49 & 2.71  & 636.63 & 5.31  & 477.37 & 3.98 \\
\cmidrule{2-9}     & 80    & 138.07 & 401.6 & 2.91  & 692.24 & 5.01  & 595.98 & 4.32 \\
\cmidrule{2-9}     & 100   & 131.2 & 381.66 & 2.91  & 757.44 & 5.77  & 509.6 & 3.88 \\
    \midrule
    \multirow{4}[8]{*}{MC} & 40    & 130.28 & 384.72 & 2.95  & 290.71 & 2.23  & 512.94 & 3.94 \\
\cmidrule{2-9}     & 60    & 112.21 & 250.31 & 2.23  & 343.86 & 3.06  & 355.39 & 3.17 \\
\cmidrule{2-9}     & 80    & 133.18 & 383.9 & 2.88  & 431.14 & 3.24  & 555.53 & 4.17 \\
\cmidrule{2-9}     & 100   & 116.28 & 331.52 & 2.85  & 492.92 & 4.24  & 491.82 & 4.23 \\
    \midrule
    \multicolumn{3}{c}{Avg. Ratio} & \multicolumn{2}{c}{3.59} & \multicolumn{2}{c}{5.13} & \multicolumn{2}{c}{4.79} \\
    \bottomrule
    \end{tabular}%
\end{table*}%

	\subsection{Objective Values of MEGO+X and the Baselines}
	MEGO+X uses MEGO as the initial solution generator for X, and X is the baseline. 
	We compare the objective values achieved by MEGO+X and the baselines on the training problem classes while the budget of \#FEs is 800.
	Each problem instance is optimized 30 times.
	The results’ statistical significance was tested through the Wilcoxon rank-sum statistic.
	Table~\ref{tab: result_x_om},~\ref{tab: result_x_kp},~\ref{tab: result_x_mc} compare the objective values achieved by MEGO+X and baselines on the OM problem, KP, and MC problem, respectively.
	
\begin{table*}[htbp]
  \centering
  \caption{\textbf{Objective values achieved by MEGO+X and the Baselines on the OM Problem.} ``↑, ↓, →'' represents that the corresponding method is significantly better, worse, or not significantly different than the MEGO, respectively.}
    \label{tab: result_x_om}
    \begin{tabular}{crcccccc}
    \toprule
    \multicolumn{2}{c}{Problem Instance} & MEGO+GA & \multicolumn{1}{c}{GA} & MEGO+HC & \multicolumn{1}{c}{HC} & MEGO+BO & \multicolumn{1}{c}{BO} \\
    \midrule
    \multirow{3}[6]{*}{Dim=40} & \multicolumn{1}{c}{ins1} & 33.93±1.55 & \multicolumn{1}{c}{32.17±2.03↓} & 40.00±0.00 & \multicolumn{1}{c}{38.73±1.44↓} & 35.27±1.34 & \multicolumn{1}{c}{33.23±1.43↓} \\
\cmidrule{2-8}     & \multicolumn{1}{c}{ins2} & 33.63±1.40 & \multicolumn{1}{c}{32.13±1.67↓} & 40.00±0.00 & \multicolumn{1}{c}{38.87±1.71↓} & 34.67±1.30 & \multicolumn{1}{c}{32.73±1.31↓} \\
\cmidrule{2-8}     & \multicolumn{1}{c}{ins3} & 34.10±1.51 & \multicolumn{1}{c}{31.83±1.55↓} & 40.00±0.00 & \multicolumn{1}{c}{38.37±2.01↓} & 35.17±0.86 & \multicolumn{1}{c}{32.77±1.12↓} \\
    \midrule
    \multirow{3}[6]{*}{Dim=60} & \multicolumn{1}{c}{ins1} & 47.67±1.62 & \multicolumn{1}{c}{44.77±1.96↓} & 55.63±1.20 & \multicolumn{1}{c}{42.57±3.45↓} & 47.23±1.45 & \multicolumn{1}{c}{46.23±1.73↓} \\
\cmidrule{2-8}     & \multicolumn{1}{c}{ins2} & 47.30±1.51 & \multicolumn{1}{c}{44.27±2.03↓} & 54.50±1.36 & \multicolumn{1}{c}{42.90±3.52↓} & 47.30±1.51 & \multicolumn{1}{c}{46.10±1.87↓} \\
\cmidrule{2-8}     & \multicolumn{1}{c}{ins3} & 46.83±2.16 & \multicolumn{1}{c}{44.77±2.28↓} & 53.83±1.27 & \multicolumn{1}{c}{42.07±3.44↓} & 47.70±1.46 & \multicolumn{1}{c}{45.27±1.36↓} \\
    \midrule
    \multirow{3}[6]{*}{Dim=80} & \multicolumn{1}{c}{ins1} & 59.33±2.05 & \multicolumn{1}{c}{56.97±2.17↓} & 62.87±2.01 & \multicolumn{1}{c}{49.83±3.64↓} & 60.77±2.36 & \multicolumn{1}{c}{57.83±1.67↓} \\
\cmidrule{2-8}     & \multicolumn{1}{c}{ins2} & 61.13±2.32 & \multicolumn{1}{c}{57.33±2.51↓} & 65.80±1.66 & \multicolumn{1}{c}{50.70±4.66↓} & 61.13±1.91 & \multicolumn{1}{c}{57.97±1.87↓} \\
\cmidrule{2-8}     & \multicolumn{1}{c}{ins3} & 60.33±2.07 & \multicolumn{1}{c}{57.43±2.39↓} & 65.13±1.52 & \multicolumn{1}{c}{51.47±4.83↓} & 60.90±1.80 & \multicolumn{1}{c}{57.27±2.10↓} \\
    \midrule
    \multirow{3}[6]{*}{Dim=100} & \multicolumn{1}{c}{ins1} & 73.77±2.88 & \multicolumn{1}{c}{68.50±2.78↓} & 75.60±3.28 & \multicolumn{1}{c}{58.07±4.25↓} & 74.30±2.53 & \multicolumn{1}{c}{69.33±2.41↓} \\
\cmidrule{2-8}     & \multicolumn{1}{c}{ins2} & 71.57±2.60 & \multicolumn{1}{c}{68.90±2.31↓} & 73.07±2.19 & \multicolumn{1}{c}{57.50±3.96↓} & 73.70±2.22 & \multicolumn{1}{c}{69.33±2.26↓} \\
\cmidrule{2-8}     & \multicolumn{1}{c}{ins3} & 70.97±3.05 & \multicolumn{1}{c}{68.07±2.79↓} & 70.83±1.24 & \multicolumn{1}{c}{57.63±4.85↓} & 73.07±2.05 & \multicolumn{1}{c}{69.40±2.12↓} \\
    \midrule
    \multicolumn{2}{c}{W-D-L} & \multicolumn{2}{c}{12-0-0} & \multicolumn{2}{c}{12-0-0} & \multicolumn{2}{c}{12-0-0} \\
    \bottomrule
    \end{tabular}
\end{table*}%

\begin{table*}[htbp]
  \centering
  \caption{\textbf{Objective values achieved by MEGO+X and the Baselines on the KP.} ''↑, ↓, →'' represents that the corresponding method is significantly better, worse, or not significantly different than the MEGO, respectively.}
    \label{tab: result_x_kp}
    \begin{tabular}{crcccccc}
    \toprule
    \multicolumn{2}{c}{Problem Instance} & MEGO+GA & \multicolumn{1}{c}{GA} & MEGO+HC & \multicolumn{1}{c}{HC} & MEGO+BO & \multicolumn{1}{c}{BO} \\
    \midrule
    \multirow{3}[6]{*}{Dim=40} & \multicolumn{1}{c}{ins1} & 7.62±0.05 & \multicolumn{1}{c}{7.48±0.13↓} & 7.73±0.03 & \multicolumn{1}{c}{7.38±0.17↓} & 7.65±0.08 & \multicolumn{1}{c}{7.50±0.15↓} \\
\cmidrule{2-8}     & \multicolumn{1}{c}{ins2} & 13.35±0.11 & \multicolumn{1}{c}{13.25±0.16↓} & 13.42±0.08 & \multicolumn{1}{c}{13.15±0.13↓} & 13.36±0.12 & \multicolumn{1}{c}{13.26±0.09↓} \\
\cmidrule{2-8}     & \multicolumn{1}{c}{ins3} & 5.98±0.05 & \multicolumn{1}{c}{5.89±0.08↓} & 6.04±0.03 & \multicolumn{1}{c}{5.83±0.10↓} & 5.98±0.05 & \multicolumn{1}{c}{5.87±0.08↓} \\
    \midrule
    \multirow{3}[6]{*}{Dim=60} & \multicolumn{1}{c}{ins1} & 8.89±0.07 & \multicolumn{1}{c}{8.86±0.09→} & 8.95±0.08 & \multicolumn{1}{c}{8.78±0.14↓} & 8.96±0.10 & \multicolumn{1}{c}{8.89±0.08↓} \\
\cmidrule{2-8}     & \multicolumn{1}{c}{ins2} & 7.55±0.11 & \multicolumn{1}{c}{7.56±0.16→} & 7.56±0.13 & \multicolumn{1}{c}{7.38±0.10↓} & 7.59±0.12 & \multicolumn{1}{c}{7.51±0.11↓} \\
\cmidrule{2-8}     & \multicolumn{1}{c}{ins3} & 13.07±0.06 & \multicolumn{1}{c}{13.04±0.08↓} & 13.11±0.05 & \multicolumn{1}{c}{12.93±0.08↓} & 13.05±0.06 & \multicolumn{1}{c}{13.03±0.07→} \\
    \midrule
    \multirow{3}[6]{*}{Dim=80} & \multicolumn{1}{c}{ins1} & 19.32±0.09 & \multicolumn{1}{c}{19.21±0.10↓} & 19.32±0.09 & \multicolumn{1}{c}{19.04±0.13↓} & 19.34±0.09 & \multicolumn{1}{c}{19.23±0.10↓} \\
\cmidrule{2-8}     & \multicolumn{1}{c}{ins2} & 27.59±0.24 & \multicolumn{1}{c}{27.09±0.70↓} & 27.68±0.11 & \multicolumn{1}{c}{26.52±1.42↓} & 27.60±0.22 & \multicolumn{1}{c}{27.34±0.48↓} \\
\cmidrule{2-8}     & \multicolumn{1}{c}{ins3} & 23.38±0.13 & \multicolumn{1}{c}{23.41±0.19→} & 23.33±0.13 & \multicolumn{1}{c}{22.88±0.33↓} & 23.43±0.15 & \multicolumn{1}{c}{23.36±0.17→} \\
    \midrule
    \multirow{3}[6]{*}{Dim=100} & \multicolumn{1}{c}{ins1} & 18.45±0.05 & \multicolumn{1}{c}{18.45±0.05→} & 18.42±0.03 & \multicolumn{1}{c}{18.35±0.07↓} & 18.44±0.05 & \multicolumn{1}{c}{18.41±0.04↓} \\
\cmidrule{2-8}     & \multicolumn{1}{c}{ins2} & 38.34±0.32 & \multicolumn{1}{c}{37.45±0.93↓} & 38.71±0.14 & \multicolumn{1}{c}{33.69±2.52↓} & 38.32±0.39 & \multicolumn{1}{c}{37.68±0.82↓} \\
\cmidrule{2-8}     & \multicolumn{1}{c}{ins3} & 15.34±0.06 & \multicolumn{1}{c}{15.32±0.06→} & 15.36±0.05 & \multicolumn{1}{c}{15.20±0.08↓} & 15.38±0.07 & \multicolumn{1}{c}{15.34±0.07↓} \\
    \midrule
    \multicolumn{2}{c}{W-D-L} & \multicolumn{2}{c}{7-5-0} & \multicolumn{2}{c}{12-0-0} & \multicolumn{2}{c}{10-2-0} \\
    \bottomrule
    \end{tabular}%
\end{table*}%

\begin{table*}[htbp]
  \centering
  \caption{\textbf{Objective values achieved by MEGO+X and the Baselines on the MC Problem.} ``↑, ↓, →'' represents that the corresponding method is significantly better, worse, or not significantly different than the MEGO, respectively.}
    \label{tab: result_x_mc}
    \begin{tabular}{crcccccc}
    \toprule
    \multicolumn{2}{c}{Problem Instance} & \multicolumn{1}{c}{MEGO+GA} & \multicolumn{1}{c}{GA} & MEGO+HC & \multicolumn{1}{c}{HC} & MEGO+BO & \multicolumn{1}{c}{BO} \\
    \midrule
    \multirow{3}[6]{*}{Dim=40} & \multicolumn{1}{c}{ins1} & \multicolumn{1}{c}{191.73±3.55} & \multicolumn{1}{c}{186.97±3.03↓} & 196.07±0.36 & \multicolumn{1}{c}{192.27±4.88↓} & 192.50±3.13 & \multicolumn{1}{c}{186.47±4.18↓} \\
\cmidrule{2-8}     & \multicolumn{1}{c}{ins2} & \multicolumn{1}{c}{270.80±2.97} & \multicolumn{1}{c}{270.70±3.41→} & 279.03±1.87 & \multicolumn{1}{c}{275.27±3.84↓} & 271.90±2.34 & \multicolumn{1}{c}{271.90±2.86→} \\
\cmidrule{2-8}     & \multicolumn{1}{c}{ins3} & \multicolumn{1}{c}{140.33±1.14} & \multicolumn{1}{c}{138.83±2.27↓} & 142.57±1.86 & \multicolumn{1}{c}{139.97±1.92↓} & 140.73±1.88 & \multicolumn{1}{c}{139.83±2.49↓} \\
    \midrule
    \multirow{3}[6]{*}{Dim=60} & \multicolumn{1}{c}{ins1} & \multicolumn{1}{c}{447.20±3.11} & \multicolumn{1}{c}{446.63±4.43→} & 456.37±2.95 & \multicolumn{1}{c}{446.53±4.43↓} & 450.00±2.32 & \multicolumn{1}{c}{448.70±3.22→} \\
\cmidrule{2-8}     & \multicolumn{1}{c}{ins2} & \multicolumn{1}{c}{630.97±4.00} & \multicolumn{1}{c}{633.17±4.04↑} & 640.80±4.45 & \multicolumn{1}{c}{637.40±7.57↓} & 634.93±3.20 & \multicolumn{1}{c}{633.50±4.01→} \\
\cmidrule{2-8}     & \multicolumn{1}{c}{ins3} & \multicolumn{1}{c}{472.53±3.37} & \multicolumn{1}{c}{471.77±4.19→} & 479.87±3.65 & \multicolumn{1}{c}{476.87±7.67↓} & 477.13±3.24 & \multicolumn{1}{c}{473.40±4.04↓} \\
    \midrule
    \multirow{3}[6]{*}{Dim=80} & \multicolumn{1}{c}{ins1} & \multicolumn{1}{c}{1062.00±6.61} & \multicolumn{1}{c}{1055.00±6.81↓} & 1075.33±7.59 & \multicolumn{1}{c}{1061.87±10.22↓} & 1063.53±5.94 & \multicolumn{1}{c}{1056.73±6.45↓} \\
\cmidrule{2-8}     & \multicolumn{1}{c}{ins2} & \multicolumn{1}{c}{1077.57±4.88} & \multicolumn{1}{c}{1073.43±6.64↓} & 1091.07±5.64 & \multicolumn{1}{c}{1082.27±10.70↓} & 1078.77±4.39 & \multicolumn{1}{c}{1074.57±5.59↓} \\
\cmidrule{2-8}     & \multicolumn{1}{c}{ins3} & \multicolumn{1}{c}{718.97±5.60} & \multicolumn{1}{c}{713.43±7.29↓} & 740.73±6.86 & \multicolumn{1}{c}{723.40±11.53↓} & 718.77±6.05 & \multicolumn{1}{c}{715.90±5.04↓} \\
    \midrule
    \multirow{3}[6]{*}{Dim=100} & \multicolumn{1}{c}{ins1} & \multicolumn{1}{c}{1446.80±7.90} & \multicolumn{1}{c}{1446.23±8.85→} & 1476.00±7.99 & \multicolumn{1}{c}{1446.47±13.35↓} & 1451.23±8.92 & \multicolumn{1}{c}{1440.93±8.30↓} \\
\cmidrule{2-8}     & \multicolumn{1}{c}{ins2} & \multicolumn{1}{c}{1132.00±7.58} & \multicolumn{1}{c}{1129.23±9.50→} & 1159.43±9.75 & \multicolumn{1}{c}{1137.13±13.86↓} & 1130.37±7.26 & \multicolumn{1}{c}{1127.17±8.11→} \\
\cmidrule{2-8}     & \multicolumn{1}{c}{ins3} & \multicolumn{1}{c}{1392.60±7.94} & \multicolumn{1}{c}{1382.73±8.50↓} & 1409.50±7.73 & \multicolumn{1}{c}{1388.30±9.46↓} & 1395.20±6.88 & \multicolumn{1}{c}{1384.50±6.38↓} \\
    \midrule
    \multicolumn{2}{c}{W-D-L} & \multicolumn{2}{c}{6-5-1} & \multicolumn{2}{c}{12-0-0} & \multicolumn{2}{c}{8-4-0} \\
    \bottomrule
    \end{tabular}
\end{table*}%

	\section{Detailed Experimental Results on Unseen Problem Classes}
	\subsection{Objective Values of MEGO and the Baselines}
	Table~\ref{tab: result_caop},~\ref{tab: result_cimp},~\ref{tab: result_as} compare the objective values achieved by MEGO and baselines on the CA problem, CIM problem, and AS problem, respectively, while the baselines use the same \#FEs as MEGO.
	
\begin{table*}[htbp]
  \centering
  \caption{\textbf{Objective values achieved by MEGO and the Baselines on the CA Problem.} ``↑, ↓, →'' represents that the corresponding method is significantly better, worse, or not significantly different than the MEGO, respectively.}
  \label{tab: result_caop}
    \begin{tabular}{cccccc}
    \toprule
    \multicolumn{2}{c}{Problem Instance} & \multicolumn{1}{c}{MEGO} & GA    & HC    & BO \\
    \midrule
    \multirow{3}[6]{*}{Dim=40} & \multicolumn{1}{c}{ins1} & \multicolumn{1}{c}{5576.00±0.00} & 5588.53±16.35↓ & 5585.33±17.66→ & 5589.60±13.88↓ \\
\cmidrule{2-6}     & \multicolumn{1}{c}{ins2} & \multicolumn{1}{c}{6848.00±0.00} & 6854.13±6.09↓ & 6881.87±85.76↓ & 6854.40±3.20↓ \\
\cmidrule{2-6}     & \multicolumn{1}{c}{ins3} & \multicolumn{1}{c}{5672.00±0.00} & 5697.07±87.60→ & 5755.20±172.59↓ & 5974.13±229.91↓ \\
    \midrule
    \multirow{3}[6]{*}{Dim=60} & \multicolumn{1}{c}{ins1} & \multicolumn{1}{c}{6366.93±2.72} & 6396.80±14.98↓ & 6422.40±30.80↓ & 6400.80±12.79↓ \\
\cmidrule{2-6}     & \multicolumn{1}{c}{ins2} & \multicolumn{1}{c}{9580.27±7.08} & 9598.40±18.29↓ & 9641.07±61.27↓ & 9604.53±14.71↓ \\
\cmidrule{2-6}     & \multicolumn{1}{c}{ins3} & \multicolumn{1}{c}{6264.00±0.00} & 6277.33±12.45↓ & 6299.73±39.99↓ & 6280.53±12.03↓ \\
    \midrule
    \multirow{3}[6]{*}{Dim=80} & \multicolumn{1}{c}{ins1} & \multicolumn{1}{c}{5244.80±14.69} & 5317.60±75.45↓ & 5557.07±274.31↓ & 5318.40±44.42↓ \\
\cmidrule{2-6}     & \multicolumn{1}{c}{ins2} & \multicolumn{1}{c}{6136.00±0.00} & 6218.40±47.87↓ & 6346.67±182.97↓ & 6218.67±48.72↓ \\
\cmidrule{2-6}     & \multicolumn{1}{c}{ins3} & \multicolumn{1}{c}{9156.00±17.00} & 9217.33±46.42↓ & 9361.33±146.79↓ & 9225.87±51.38↓ \\
    \midrule
    \multirow{3}[6]{*}{Dim=100} & \multicolumn{1}{c}{ins1} & \multicolumn{1}{c}{5579.47±14.85} & 5601.07±54.21→ & 6030.67±454.58↓ & 5596.00±56.48→ \\
\cmidrule{2-6}     & \multicolumn{1}{c}{ins2} & \multicolumn{1}{c}{4293.33±11.75} & 4344.27±73.48↓ & 5258.93±658.59↓ & 4356.80±29.55↓ \\
\cmidrule{2-6}     & \multicolumn{1}{c}{ins3} & \multicolumn{1}{c}{61038.93±130.05} & 61375.20±331.23↓ & 62328.00±784.37↓ & 61545.60±418.96↓ \\
    \midrule
    \multicolumn{3}{c}{W-D-L} & 10-2-0 & 11-1-0 & 11-1-0 \\
    \bottomrule
    \end{tabular}%
\end{table*}%

\begin{table*}[htbp]
  \centering
  \caption{\textbf{Objective values achieved by MEGO and the Baselines on the CIM Problem.} ``↑, ↓, →'' represents that the corresponding method is significantly better, worse, or not significantly different than the MEGO, respectively. }
  \label{tab: result_cimp}
    {\begin{tabular}{cccccc}
    \toprule
    \multicolumn{2}{c}{Problem Instance} & \multicolumn{1}{c}{MEGO} & GA    & HC    & BO \\
    \midrule
    \multirow{3}[6]{*}{Dim=40} & \multicolumn{1}{c}{ins1} & \multicolumn{1}{c}{29.99±0.09} & 29.37±0.43↓ & 29.23±1.16↓ & 28.82±0.70↓ \\
\cmidrule{2-6}     & \multicolumn{1}{c}{ins2} & \multicolumn{1}{c}{33.06±0.36} & 31.92±0.77↓ & 31.51±1.44↓ & 31.39±0.69↓ \\
\cmidrule{2-6}     & \multicolumn{1}{c}{ins3} & \multicolumn{1}{c}{34.88±0.32} & 34.15±0.64↓ & 33.28±1.66↓ & 33.86±0.65↓ \\
    \midrule
    \multirow{3}[6]{*}{Dim=60} & \multicolumn{1}{c}{ins1} & \multicolumn{1}{c}{41.95±0.75} & 40.22±1.13↓ & 38.03±1.75↓ & 39.91±1.35↓ \\
\cmidrule{2-6}     & \multicolumn{1}{c}{ins2} & \multicolumn{1}{c}{56.03±0.67} & 54.85±1.00↓ & 52.95±2.98↓ & 54.36±1.13↓ \\
\cmidrule{2-6}     & \multicolumn{1}{c}{ins3} & \multicolumn{1}{c}{49.22±0.12} & 47.89±1.01↓ & 46.37±2.83↓ & 47.70±0.66↓ \\
    \midrule
    \multirow{3}[6]{*}{Dim=80} & \multicolumn{1}{c}{ins1} & \multicolumn{1}{c}{36.19±0.44} & 35.40±0.59↓ & 32.95±1.67↓ & 35.50±0.96↓ \\
\cmidrule{2-6}     & \multicolumn{1}{c}{ins2} & \multicolumn{1}{c}{82.89±1.10} & 83.25±1.48→ & 73.50±7.43↓ & 81.74±1.40↓ \\
\cmidrule{2-6}     & \multicolumn{1}{c}{ins3} & \multicolumn{1}{c}{35.08±0.38} & 34.79±0.51↓ & 34.07±1.00↓ & 34.64±0.51↓ \\
    \midrule
    \multirow{3}[6]{*}{Dim=100} & \multicolumn{1}{c}{ins1} & \multicolumn{1}{c}{62.02±0.40} & 60.94±0.66↓ & 57.07±3.08↓ & 60.57±0.50↓ \\
\cmidrule{2-6}     & \multicolumn{1}{c}{ins2} & \multicolumn{1}{c}{44.31±0.60} & 42.53±0.89↓ & 40.79±1.54↓ & 42.01±0.96↓ \\
\cmidrule{2-6}     & \multicolumn{1}{c}{ins3} & \multicolumn{1}{c}{103.53±1.02} & 101.65±1.59↓ & 95.40±5.98↓ & 100.94±1.35↓ \\
    \midrule
    \multicolumn{3}{c}{W-D-L} & 11-1-0 & 12-0-0 & 12-0-0 \\
    \bottomrule
    \end{tabular}}
\end{table*}%

\begin{table*}[htbp]
    \centering
    \caption{\textbf{Objective values achieved by MEGO and the Baselines on the AS  Problem.} ``↑, ↓, →'' represents that the corresponding method is significantly better, worse, or not significantly different than the MEGO, respectively.}
    \label{tab: result_as}
      \begin{tabular}{cccccc}
      \toprule
      \multicolumn{2}{c}{Problem Instance} & \multicolumn{1}{c}{MEGO} & GA    & HC    & BO \\
      \midrule
      \multirow{3}[6]{*}{Dim=40} & \multicolumn{1}{c}{ins1} & \multicolumn{1}{c}{9815.50±128.94} & 9558.63±383.73↓ & 9225.20±468.86↓ & 9269.17±362.24↓ \\
  \cmidrule{2-6}     & \multicolumn{1}{c}{ins2} & \multicolumn{1}{c}{31081.43±573.42} & 29898.00±584.94↓ & 29304.37±1106.26↓ & 29489.83±553.34↓ \\
  \cmidrule{2-6}     & \multicolumn{1}{c}{ins3} & \multicolumn{1}{c}{21284.20±732.88} & 20040.63±721.47↓ & 18845.70±1630.83↓ & 19775.47±641.39↓ \\
      \midrule
      \multirow{3}[6]{*}{Dim=60} & \multicolumn{1}{c}{ins1} & \multicolumn{1}{c}{15851.67±208.03} & 15085.93±328.54↓ & 15036.77±587.77↓ & 15006.57±413.89↓ \\
  \cmidrule{2-6}     & \multicolumn{1}{c}{ins2} & \multicolumn{1}{c}{21706.80±391.56} & 20660.53±573.92↓ & 19015.37±1235.88↓ & 20483.87±558.73↓ \\
  \cmidrule{2-6}     & \multicolumn{1}{c}{ins3} & \multicolumn{1}{c}{41145.00±426.39} & 39396.90±808.88↓ & 37288.10±2194.27↓ & 38741.37±852.26↓ \\
      \midrule
      \multirow{3}[6]{*}{Dim=80} & \multicolumn{1}{c}{ins1} & \multicolumn{1}{c}{42027.97±277.02} & 41116.67±629.09↓ & 38407.73±2398.84↓ & 40948.20±536.36↓ \\
  \cmidrule{2-6}     & \multicolumn{1}{c}{ins2} & \multicolumn{1}{c}{34867.27±426.11} & 33651.47±759.20↓ & 31964.40±1527.77↓ & 33362.63±777.73↓ \\
  \cmidrule{2-6}     & \multicolumn{1}{c}{ins3} & \multicolumn{1}{c}{38326.30±743.44} & 37328.57±919.31↓ & 34838.53±1708.84↓ & 36809.03±960.53↓ \\
      \midrule
      \multirow{3}[6]{*}{Dim=100} & \multicolumn{1}{c}{ins1} & \multicolumn{1}{c}{50003.80±203.51} & 49250.83±577.65↓ & 46700.20±1655.62↓ & 48749.13±404.31↓ \\
  \cmidrule{2-6}     & \multicolumn{1}{c}{ins2} & \multicolumn{1}{c}{34579.93±836.99} & 34986.33±1205.57→ & 32089.50±1813.94↓ & 34446.40±1063.70→ \\
  \cmidrule{2-6}     & \multicolumn{1}{c}{ins3} & \multicolumn{1}{c}{38252.53±442.20} & 37462.07±671.42↓ & 35738.07±1471.09↓ & 37249.10±516.72↓ \\
      \midrule
      \multicolumn{3}{c}{W-D-L} & 11-1-0 & 12-0-0 & 11-1-0 \\
      \bottomrule
      \end{tabular}%
  \end{table*}%

	\subsection{Acceleration Ratios of MEGO compared to the Baselines}
	The comparison results in terms of acceleration ratios on the unseen problem classes are shown in Table~\ref{tab: acc_unseen}.
	
\begin{table*}[htbp]
    \centering
    \caption{Acceleration Ratios of MEGO compared to the Baselines on Unseen Problem classes.}
    \label{tab: acc_unseen}
      {\begin{tabular}{ccccccccc}
      \toprule
      \multirow{2}[4]{*}{Problem Class} & \multicolumn{1}{c}{\multirow{2}[4]{*}{Dim}} & \multicolumn{1}{c}{MEGO} & \multicolumn{2}{c}{GA} & \multicolumn{2}{c}{HC} & \multicolumn{2}{c}{BO} \\
  \cmidrule{3-9}     &       & \multicolumn{1}{c}{\#FEs} & \multicolumn{1}{c}{\#FEs} & \multicolumn{1}{c}{Ratio} & \multicolumn{1}{c}{\#FEs} & \multicolumn{1}{c}{Ratio} & \multicolumn{1}{c}{\#FEs} & \multicolumn{1}{c}{Ratio} \\
      \midrule
      \multirow{4}[8]{*}{CA} & 40    & 108.4 & 185.16 & 1.71  & 141.62 & 1.31  & 216.8 & 2 \\
  \cmidrule{2-9}     & 60    & 124.76 & 488.23 & 3.91  & 356.16 & 2.85  & 499.23 & 4 \\
  \cmidrule{2-9}     & 80    & 106.24 & 571.84 & 5.38  & 408.91 & 3.85  & 580.84 & 5.47 \\
  \cmidrule{2-9}     & 100   & 99.98 & 304.7 & 3.05  & 399.04 & 3.99  & 412.5 & 4.13 \\
      \midrule
      \multirow{4}[8]{*}{CIM} & 40    & 134.41 & 521.99 & 3.88  & 348.19 & 2.59  & 650.46 & 4.84 \\
  \cmidrule{2-9}     & 60    & 122.53 & 449.29 & 3.67  & 540.87 & 4.41  & 618.59 & 5.05 \\
  \cmidrule{2-9}     & 80    & 110.38 & 299.79 & 2.72  & 615.16 & 5.57  & 405.09 & 3.67 \\
  \cmidrule{2-9}     & 100   & 135.17 & 560.06 & 4.14  & 726.86 & 5.38  & 644.77 & 4.77 \\
      \midrule
      \multirow{4}[8]{*}{AS} & 40    & 114.46 & 470.99 & 4.11  & 402.3 & 3.51  & 573.77 & 5.01 \\
  \cmidrule{2-9}     & 60    & 120.18 & 577.58 & 4.81  & 514   & 4.28  & 656.51 & 5.46 \\
  \cmidrule{2-9}     & 80    & 123.02 & 508.94 & 4.14  & 586.01 & 4.76  & 581.57 & 4.73 \\
  \cmidrule{2-9}     & 100   & 121.01 & 356.53 & 2.95  & 650.8 & 5.38  & 463.99 & 3.83 \\
      \midrule
      \multicolumn{3}{c}{Avg. Ratio} & \multicolumn{2}{c}{3.71} & \multicolumn{2}{c}{3.99} & \multicolumn{2}{c}{4.41} \\
      \bottomrule
      \end{tabular}}
  \end{table*}%

	\subsection{Objective Values of MEGO+X and the Baselines}
	Table~\ref{tab: result_x_caop},~\ref{tab: result_x_cimp},~\ref{tab: result_x_as} compare the objective values of MEGO+X and baselines on the CA problem, the CIM problem, and the AS problem, respectively, with the budget of \#FEs being 800.
	
\begin{table*}[htbp]
    \centering
    \caption{\textbf{Objective values achieved by MEGO+X and the Baselines on the CA Problem.} ``↑, ↓, →'' represents that the corresponding method is significantly better, worse, or not significantly different than the MEGO, respectively.}
    \label{tab: result_x_caop}
      \begin{tabular}{crcccccc}
      \toprule
      \multicolumn{2}{c}{Problem Instance} & MEGO+GA & \multicolumn{1}{c}{GA} & MEGO+HC & \multicolumn{1}{c}{HC} & MEGO+BO & \multicolumn{1}{c}{BO} \\
      \midrule
      \multirow{3}[6]{*}{Dim=40} & \multicolumn{1}{c}{ins1} & 5568.80±2.40 & \multicolumn{1}{c}{5569.87±6.09→} & 5568.00±0.00 & \multicolumn{1}{c}{5568.00±0.00→} & 5568.00±0.00 & \multicolumn{1}{c}{5568.27±1.44→} \\
  \cmidrule{2-8}     & \multicolumn{1}{c}{ins2} & 6848.00±0.00 & \multicolumn{1}{c}{6848.53±2.00→} & 6848.00±0.00 & \multicolumn{1}{c}{6848.00±0.00→} & 6848.00±0.00 & \multicolumn{1}{c}{6848.00±0.00→} \\
  \cmidrule{2-8}     & \multicolumn{1}{c}{ins3} & 5672.00±0.00 & \multicolumn{1}{c}{5672.00±0.00→} & 5672.00±0.00 & \multicolumn{1}{c}{5672.00±0.00→} & 5672.00±0.00 & \multicolumn{1}{c}{5672.00±0.00→} \\
      \midrule
      \multirow{3}[6]{*}{Dim=60} & \multicolumn{1}{c}{ins1} & 6366.13±3.38 & \multicolumn{1}{c}{6374.13±15.13→} & 6361.60±3.20 & \multicolumn{1}{c}{6370.13±14.00↓} & 6363.73±3.99 & \multicolumn{1}{c}{6367.20±12.28→} \\
  \cmidrule{2-8}     & \multicolumn{1}{c}{ins2} & 9572.00±4.50 & \multicolumn{1}{c}{9577.33±10.55↓} & 9569.60±3.20 & \multicolumn{1}{c}{9573.07±5.26↓} & 9569.33±2.98 & \multicolumn{1}{c}{9572.27±3.99↓} \\
  \cmidrule{2-8}     & \multicolumn{1}{c}{ins3} & 6264.00±0.00 & \multicolumn{1}{c}{6267.20±7.62→} & 6264.00±0.00 & \multicolumn{1}{c}{6264.00±0.00→} & 6264.00±0.00 & \multicolumn{1}{c}{6264.00±0.00→} \\
      \midrule
      \multirow{3}[6]{*}{Dim=80} & \multicolumn{1}{c}{ins1} & 5229.87±8.75 & \multicolumn{1}{c}{5243.73±21.34↓} & 5224.00±0.00 & \multicolumn{1}{c}{5264.53±121.32↓} & 5225.33±4.17 & \multicolumn{1}{c}{5230.93±13.82↓} \\
  \cmidrule{2-8}     & \multicolumn{1}{c}{ins2} & 6136.00±0.00 & \multicolumn{1}{c}{6155.47±29.70↓} & 6136.00±0.00 & \multicolumn{1}{c}{6136.00±0.00→} & 6136.00±0.00 & \multicolumn{1}{c}{6137.33±2.98→} \\
  \cmidrule{2-8}     & \multicolumn{1}{c}{ins3} & 9133.87±7.43 & \multicolumn{1}{c}{9149.87±27.24↓} & 9126.40±4.33 & \multicolumn{1}{c}{9128.27±9.12→} & 9120.27±4.84 & \multicolumn{1}{c}{9134.13±9.39↓} \\
      \midrule
      \multirow{3}[6]{*}{Dim=100} & \multicolumn{1}{c}{ins1} & 5532.27±18.33 & \multicolumn{1}{c}{5532.27±25.61→} & 5505.07±3.41 & \multicolumn{1}{c}{5553.60±177.15↓} & 5501.07±7.00 & \multicolumn{1}{c}{5512.53±12.72↓} \\
  \cmidrule{2-8}     & \multicolumn{1}{c}{ins2} & 4261.07±17.83 & \multicolumn{1}{c}{4259.73±26.02→} & 4240.53±25.29 & \multicolumn{1}{c}{4244.00±27.07→} & 4213.60±12.76 & \multicolumn{1}{c}{4240.00±12.73↓} \\
  \cmidrule{2-8}     & \multicolumn{1}{c}{ins3} & 60736.27±159.78 & \multicolumn{1}{c}{60801.33±257.36→} & 60474.93±18.53 & \multicolumn{1}{c}{60598.67±178.37↓} & 60519.73±45.79 & \multicolumn{1}{c}{60583.73±63.98↓} \\
      \midrule
      \multicolumn{2}{c}{W-D-L} & \multicolumn{2}{c}{4-8-0} & \multicolumn{2}{c}{5-7-0} & \multicolumn{2}{c}{6-6-0} \\
      \bottomrule
      \end{tabular}%
  \end{table*}%
  
\begin{table*}[htbp]
    \centering
    \caption{\textbf{Objective values achieved by MEGO+X and the Baselines on the CIM Problem.} ``↑, ↓, →'' represents that the corresponding method is significantly better, worse, or not significantly different than the MEGO, respectively.}
    \label{tab: result_x_cimp}
      \begin{tabular}{crcccccc}
      \toprule
      \multicolumn{2}{c}{Problem Instance} & MEGO+GA & \multicolumn{1}{c}{GA} & MEGO+HC & \multicolumn{1}{c}{HC} & MEGO+BO & \multicolumn{1}{c}{BO} \\
      \midrule
      \multirow{3}[6]{*}{Dim=40} & \multicolumn{1}{c}{ins1} & 30.08±0.08 & \multicolumn{1}{c}{29.94±0.28↓} & 30.25±0.09 & \multicolumn{1}{c}{30.22±0.16→} & 30.25±0.13 & \multicolumn{1}{c}{30.08±0.14↓} \\
  \cmidrule{2-8}     & \multicolumn{1}{c}{ins2} & 33.59±0.33 & \multicolumn{1}{c}{33.19±0.46↓} & 34.23±0.18 & \multicolumn{1}{c}{33.70±0.44↓} & 33.82±0.26 & \multicolumn{1}{c}{33.31±0.38↓} \\
  \cmidrule{2-8}     & \multicolumn{1}{c}{ins3} & 35.92±0.59 & \multicolumn{1}{c}{35.54±0.62↓} & 37.26±0.55 & \multicolumn{1}{c}{35.91±1.03↓} & 36.44±0.53 & \multicolumn{1}{c}{36.11±0.58↓} \\
      \midrule
      \multirow{3}[6]{*}{Dim=60} & \multicolumn{1}{c}{ins1} & 43.92±1.47 & \multicolumn{1}{c}{43.36±1.61→} & 44.48±0.97 & \multicolumn{1}{c}{42.05±2.05↓} & 45.30±1.62 & \multicolumn{1}{c}{43.63±1.89↓} \\
  \cmidrule{2-8}     & \multicolumn{1}{c}{ins2} & 57.83±1.05 & \multicolumn{1}{c}{57.29±1.06↓} & 59.52±1.22 & \multicolumn{1}{c}{57.31±1.90↓} & 58.86±0.65 & \multicolumn{1}{c}{57.86±1.07↓} \\
  \cmidrule{2-8}     & \multicolumn{1}{c}{ins3} & 49.43±0.20 & \multicolumn{1}{c}{49.21±0.50→} & 49.74±0.18 & \multicolumn{1}{c}{49.46±0.38↓} & 49.70±0.20 & \multicolumn{1}{c}{49.30±0.29↓} \\
      \midrule
      \multirow{3}[6]{*}{Dim=80} & \multicolumn{1}{c}{ins1} & 38.86±1.17 & \multicolumn{1}{c}{38.50±1.52→} & 39.54±1.25 & \multicolumn{1}{c}{35.22±1.92↓} & 39.86±0.58 & \multicolumn{1}{c}{38.16±1.37↓} \\
  \cmidrule{2-8}     & \multicolumn{1}{c}{ins2} & 86.28±0.88 & \multicolumn{1}{c}{85.51±1.18↓} & 86.56±1.05 & \multicolumn{1}{c}{84.11±2.67↓} & 86.50±1.19 & \multicolumn{1}{c}{85.91±0.96↓} \\
  \cmidrule{2-8}     & \multicolumn{1}{c}{ins3} & 36.56±0.80 & \multicolumn{1}{c}{36.02±0.80↓} & 36.49±0.73 & \multicolumn{1}{c}{35.49±0.84↓} & 37.02±0.45 & \multicolumn{1}{c}{36.49±0.63↓} \\
      \midrule
      \multirow{3}[6]{*}{Dim=100} & \multicolumn{1}{c}{ins1} & 62.65±0.49 & \multicolumn{1}{c}{62.03±0.63↓} & 63.07±0.33 & \multicolumn{1}{c}{62.01±0.94↓} & 62.95±0.52 & \multicolumn{1}{c}{62.66±0.46↓} \\
  \cmidrule{2-8}     & \multicolumn{1}{c}{ins2} & 45.22±0.68 & \multicolumn{1}{c}{44.06±0.94↓} & 46.16±0.33 & \multicolumn{1}{c}{43.26±1.25↓} & 45.67±0.49 & \multicolumn{1}{c}{44.56±0.58↓} \\
  \cmidrule{2-8}     & \multicolumn{1}{c}{ins3} & 105.91±1.11 & \multicolumn{1}{c}{104.53±1.43↓} & 106.81±1.14 & \multicolumn{1}{c}{103.03±1.91↓} & 106.65±1.29 & \multicolumn{1}{c}{105.44±0.97↓} \\
      \midrule
      \multicolumn{2}{c}{W-D-L} & \multicolumn{2}{c}{9-3-0} & \multicolumn{2}{c}{11-1-0} & \multicolumn{2}{c}{12-0-0} \\
      \bottomrule
      \end{tabular}%
  \end{table*}%
  
\begin{table*}[htbp]
    \centering
    \caption{\textbf{Objective values achieved by MEGO+X and the Baselines on the AS Problem.} ``↑, ↓, →'' represents that the corresponding method is significantly better, worse, or not significantly different than the MEGO, respectively.}
    \label{tab: result_x_as}
      \begin{tabular}{crcccccc}
      \toprule
      \multicolumn{2}{c}{Problem Instance} & MEGO+GA & \multicolumn{1}{c}{GA} & MEGO+HC & \multicolumn{1}{c}{HC} & MEGO+BO & \multicolumn{1}{c}{BO} \\
      \midrule
      \multirow{3}[6]{*}{Dim=40} & \multicolumn{1}{c}{ins1} & 10122.6±308.0 & \multicolumn{1}{c}{10132.0±442.0→} & 10250.2±235.6 & \multicolumn{1}{c}{9904.1±375.6↓} & 10483.1±522.9 & \multicolumn{1}{c}{10284.1±410.2→} \\
  \cmidrule{2-8}     & \multicolumn{1}{c}{ins2} & 32099.3±488.9 & \multicolumn{1}{c}{31534.5±757.5↓} & 33702.8±134.5 & \multicolumn{1}{c}{32850.9±1185.0↓} & 32401.5±386.1 & \multicolumn{1}{c}{31927.1±721.7↓} \\
  \cmidrule{2-8}     & \multicolumn{1}{c}{ins3} & 22981.1±673.8 & \multicolumn{1}{c}{22233.3±690.0↓} & 24181.8±681.5 & \multicolumn{1}{c}{22918.7±837.3↓} & 23416.9±691.8 & \multicolumn{1}{c}{22619.5±765.5↓} \\
      \midrule
      \multirow{3}[6]{*}{Dim=60} & \multicolumn{1}{c}{ins1} & 16068.2±543.8 & \multicolumn{1}{c}{15669.4±493.9↓} & 16873.6±1073.3 & \multicolumn{1}{c}{15968.9±396.5↓} & 16429.7±717.1 & \multicolumn{1}{c}{15832.6±66.7↓} \\
  \cmidrule{2-8}     & \multicolumn{1}{c}{ins2} & 22506.8±416.6 & \multicolumn{1}{c}{22016.3±711.8↓} & 23339.7±187.6 & \multicolumn{1}{c}{22569.0±953.0↓} & 23161.4±285.2 & \multicolumn{1}{c}{22510.6±443.0↓} \\
  \cmidrule{2-8}     & \multicolumn{1}{c}{ins3} & 41975.6±682.8 & \multicolumn{1}{c}{41364.7±964.3↓} & 44223.5±369.3 & \multicolumn{1}{c}{43191.0±751.3↓} & 42448.1±506.1 & \multicolumn{1}{c}{41578.0±739.0↓} \\
      \midrule
      \multirow{3}[6]{*}{Dim=80} & \multicolumn{1}{c}{ins1} & 42871.1±507.8 & \multicolumn{1}{c}{42485.0±729.6↓} & 43753.7±405.1 & \multicolumn{1}{c}{41911.1±863.8↓} & 43087.7±357.1 & \multicolumn{1}{c}{42653.1±395.2↓} \\
  \cmidrule{2-8}     & \multicolumn{1}{c}{ins2} & 35942.4±689.0 & \multicolumn{1}{c}{35279.8±574.8↓} & 37487.7±410.3 & \multicolumn{1}{c}{36237.9±888.3↓} & 36383.1±463.8 & \multicolumn{1}{c}{35764.2±523.1↓} \\
  \cmidrule{2-8}     & \multicolumn{1}{c}{ins3} & 39910.1±772.4 & \multicolumn{1}{c}{39326.5±959.3↓} & 41780.9±797.0 & \multicolumn{1}{c}{39112.6±1387.1↓} & 40444.6±536.5 & \multicolumn{1}{c}{39927.0±715.5↓} \\
      \midrule
      \multirow{3}[6]{*}{Dim=100} & \multicolumn{1}{c}{ins1} & 50766.4±341.6 & \multicolumn{1}{c}{50292.8±427.3↓} & 51768.9±219.3 & \multicolumn{1}{c}{49399.4±894.6↓} & 50967.9±283.4 & \multicolumn{1}{c}{50570.9±439.7↓} \\
  \cmidrule{2-8}     & \multicolumn{1}{c}{ins2} & 37678.3±1118.5 & \multicolumn{1}{c}{37481.3±1258.8→} & 38316.5±1242.2 & \multicolumn{1}{c}{35748.0±1389.8↓} & 38641.3±893.5 & \multicolumn{1}{c}{37619.1±874.4↓} \\
  \cmidrule{2-8}     & \multicolumn{1}{c}{ins3} & 39525.2±707.4 & \multicolumn{1}{c}{39238.8±806.2→} & 40665.6±685.8 & \multicolumn{1}{c}{38369.1±1309.5↓} & 39772.6±679.6 & \multicolumn{1}{c}{39267.1±794.8↓} \\
      \midrule
      \multicolumn{2}{c}{W-D-L} & \multicolumn{2}{c}{9-3-0} & \multicolumn{2}{c}{12-0-0} & \multicolumn{2}{c}{11-1-0} \\
      \bottomrule
      \end{tabular}
  \end{table*}%

	\section{Detailed Experimental Results for Component Study}
	
	\subsection{Comparison with Other Expert Selection Strategies}
	We compare our proposed MEGO method with several alternative expert selection strategies. The methods included in the comparison are:
	
	\begin{itemize}
		\item MAX: The maximum value among the optimal solutions obtained by all experts~(The best-case scenario while select one expert).
		\item MEGO: The best solution from the set of experts selected by the MEGO strategy.
		\item SAMPLE: The best solution from a randomly sampled set of experts, where the sample size matches the number selected by MEGO. The reported result is the average over 10,000 independent sampling trials~(Randomly select multiple experts).
		\item MEAN: The mean value of the optimal solutions obtained by all experts~(Randomly select only one expert).
		\item MIN: The minimum value among the optimal solutions obtained by all experts~(The worst-case scenario while select one expert).
	\end{itemize}
	
	The experimental settings are configured as follows. 
	For MEGO and SAMPLE, 64 randomly sampled solutions are used for fine-tuning. Experts are selected based on similarity, with each selected expert generating 4 new solutions. On average, MEGO selects 14 experts.
	For MAX, MEAN, and MIN, 116 randomly sampled solutions are used for fine-tuning, with each expert also generating 4 new solutions.
	
	All comparative results are presented in Table~\ref{tab: strategies_result_om}-\ref{tab: strategies_result_ca}. For a robust evaluation, each method was executed for 30 independent runs on every problem instance. The table lists the mean and standard deviation of the performance across these runs. Furthermore, we conducted the Wilcoxon rank-sum test~(at a \(0.05\) significance level) to statistically compare the results of MEGO against each of the other strategies per instance. The outcomes of these significance tests are annotated within the table.

\begin{table*}[htbp]
\centering
\caption{\textbf{Comparative Performance of Expert Selection Strategies on the OM Problem} ``↑, ↓, →'' represents that the corresponding strategy is significantly better, worse, or not significantly different than the MEGO, respectively.}
\label{tab: strategies_result_om}
	{\begin{tabular}{ccccccc}
	\toprule
	\multicolumn{2}{c}{Problem Instance} & MEGO & MAX & SAMPLE & MEAN & MIN \\
	\midrule
	\multirow{3}[4]{*}{Dim=40} & ins1 & 31.17±1.04 & 34.60±0.84↑ & 30.77±0.46↓ & 26.60±0.51↓ & 13.73±1.61↓ \\
	\cmidrule{2-7}  & ins2 & 32.57±1.12 & 33.73±0.77↑ & 31.66±0.68↓ & 27.01±0.33↓ & 15.23±0.99↓ \\
	\cmidrule{2-7}  & ins3 & 29.77±1.12 & 31.97±1.30↑ & 29.61±0.43→ & 24.96±0.57↓ & 12.80±1.14↓ \\
	\midrule
	\multirow{3}[4]{*}{Dim=60} & ins1 & 44.17±1.29 & 45.83±1.46↑ & 44.47±0.94→ & 37.88±0.43↓ & 25.77±1.89↓ \\
	\cmidrule{2-7}  & ins2 & 44.60±1.36 & 47.70±0.97↑ & 44.13±1.08→ & 38.99±0.56↓ & 22.53±2.28↓ \\
	\cmidrule{2-7}  & ins3 & 43.07±1.15 & 46.87±1.73↑ & 42.83±0.83→ & 37.67±0.61↓ & 22.33±1.78↓ \\
	\midrule
	\multirow{3}[4]{*}{Dim=80} & ins1 & 55.37±2.47 & 57.10±1.30↑ & 56.22±1.30↑ & 46.34±0.64↓ & 33.20±2.87↓ \\
	\cmidrule{2-7}  & ins2 & 52.57±2.65 & 57.03±1.58↑ & 53.29±0.81→ & 47.24±0.73↓ & 30.70±4.17↓ \\
	\cmidrule{2-7}  & ins3 & 56.73±1.95 & 56.60±1.69→ & 57.13±0.79→ & 46.61±0.70↓ & 31.20±1.70↓ \\
	\midrule
	\multirow{3}[4]{*}{Dim=100} & ins1 & 69.93±2.06 & 70.23±1.80→ & 68.64±1.53↓ & 59.25±0.61↓ & 42.50±2.74↓ \\
	\cmidrule{2-7}  & ins2 & 66.80±2.07 & 68.10±1.19↑ & 65.32±1.09↓ & 60.02±0.51↓ & 45.20±2.59↓ \\
	\cmidrule{2-7}  & ins3 & 63.93±1.82 & 71.20±2.57↑ & 64.14±0.85→ & 59.57±0.51↓ & 40.80±2.41↓ \\
	\midrule
\multicolumn{3}{c}{W-D-L}  & 0-2-10 & 4-7-1 & 12-0-0 & 12-0-0\\
	\bottomrule
	\end{tabular}}
\end{table*}

\begin{table*}[htbp]
\centering
\caption{\textbf{Comparative Performance of Expert Selection Strategies on the MC Problem} ``↑, ↓, →'' represents that the corresponding strategy is significantly better, worse, or not significantly different than the MEGO, respectively.}
\label{tab: strategies_result_mc}
	{\begin{tabular}{ccccccc}
	\toprule
	\multicolumn{2}{c}{Problem Instance} & MEGO & MAX & SAMPLE & MEAN & MIN \\
	\midrule
	\multirow{3}[4]{*}{Dim=40} & ins1 & 138.43±1.09 & 138.23±0.62→ & 138.41±0.94→ & 131.50±0.61↓ & 115.90±2.43↓ \\
	\cmidrule{2-7}  & ins2 & 185.63±3.86 & 181.70±1.16↓ & 185.45±2.45→ & 173.00±0.69↓ & 158.10±2.10↓ \\
	\cmidrule{2-7}  & ins3 & 265.70±2.24 & 267.43±2.38↑ & 264.35±1.18↓ & 258.75±0.80↓ & 243.43±1.84↓ \\
	\midrule
	\multirow{3}[4]{*}{Dim=60} & ins1 & 466.30±1.59 & 472.17±2.66↑ & 466.98±1.46→ & 459.38±1.07↓ & 441.67±4.57↓ \\
	\cmidrule{2-7}  & ins2 & 441.47±1.98 & 444.67±2.36↑ & 441.20±1.16→ & 435.40±0.73↓ & 417.43±3.69↓ \\
	\cmidrule{2-7}  & ins3 & 623.77±3.15 & 627.93±2.74↑ & 624.27±2.00→ & 613.58±0.80↓ & 598.97±2.48↓ \\
	\midrule
	\multirow{3}[4]{*}{Dim=80} & ins1 & 1067.53±4.39 & 1077.80±1.62↑ & 1066.25±2.87→ & 1057.57±1.10↓ & 1033.50±4.22↓ \\
	\cmidrule{2-7}  & ins2 & 708.80±5.54 & 713.27±4.07↑ & 708.30±2.82→ & 694.79±1.38↓ & 675.33±4.29↓ \\
	\cmidrule{2-7}  & ins3 & 1050.10±4.72 & 1064.77±4.86↑ & 1049.59±3.56→ & 1040.42±1.37↓ & 1010.27±5.60↓ \\
	\midrule
	\multirow{3}[4]{*}{Dim=100} & ins1 & 1383.23±6.64 & 1380.93±4.49→ & 1382.11±5.20→ & 1361.50±1.55↓ & 1329.40±5.96↓ \\
	\cmidrule{2-7}  & ins2 & 1114.13±7.01 & 1115.03±5.28→ & 1113.81±5.12→ & 1092.66±1.82↓ & 1048.53±16.30↓ \\
	\cmidrule{2-7}  & ins3 & 1422.03±5.65 & 1446.90±7.44↑ & 1423.82±4.48→ & 1420.46±1.49↓ & 1388.37±7.16↓ \\
	\midrule
\multicolumn{3}{c}{W-D-L}  & 1-3-8 & 1-11-0 & 12-0-0 & 12-0-0\\
	\bottomrule
	\end{tabular}}
\end{table*}

\begin{table*}[htbp]
\centering
\caption{\textbf{Comparative Performance of Expert Selection Strategies on the KP Problem} ``↑, ↓, →'' represents that the corresponding strategy is significantly better, worse, or not significantly different than the MEGO, respectively.}
\label{tab: strategies_result_kp}
	{\begin{tabular}{ccccccc}
	\toprule
	\multicolumn{2}{c}{Problem Instance} & MEGO & MAX & SAMPLE & MEAN & MIN \\
	\midrule
	\multirow{3}[4]{*}{Dim=40} & ins1 & 13.14±0.09 & 13.18±0.11→ & 13.13±0.07→ & 12.58±0.10↓ & 10.38±0.99↓ \\
	\cmidrule{2-7}  & ins2 & 7.47±0.11 & 7.44±0.10→ & 7.44±0.08→ & 6.99±0.04↓ & 6.45±0.15↓ \\
	\cmidrule{2-7}  & ins3 & 5.91±0.03 & 5.88±0.07↓ & 5.90±0.02↓ & 5.59±0.02↓ & 5.16±0.11↓ \\
	\midrule
	\multirow{3}[4]{*}{Dim=60} & ins1 & 7.42±0.08 & 7.55±0.08↑ & 7.43±0.07→ & 7.19±0.03↓ & 6.68±0.13↓ \\
	\cmidrule{2-7}  & ins2 & 12.97±0.07 & 12.98±0.06→ & 12.96±0.04→ & 12.68±0.04↓ & 11.72±0.50↓ \\
	\cmidrule{2-7}  & ins3 & 8.81±0.09 & 8.94±0.05↑ & 8.76±0.05↓ & 8.62±0.03↓ & 8.16±0.13↓ \\
	\midrule
	\multirow{3}[4]{*}{Dim=80} & ins1 & 25.92±0.66 & 26.84±0.50↑ & 25.72±0.52→ & 23.30±0.22↓ & 18.68±1.10↓ \\
	\cmidrule{2-7}  & ins2 & 19.22±0.06 & 19.16±0.07↓ & 19.18±0.05↓ & 18.84±0.03↓ & 18.16±0.43↓ \\
	\cmidrule{2-7}  & ins3 & 23.09±0.11 & 23.22±0.11↑ & 23.11±0.09→ & 22.49±0.14↓ & 18.43±1.34↓ \\
	\midrule
	\multirow{3}[4]{*}{Dim=100} & ins1 & 36.08±1.09 & 38.34±0.40↑ & 36.34±0.53→ & 33.09±0.35↓ & 27.42±2.36↓ \\
	\cmidrule{2-7}  & ins2 & 18.37±0.05 & 18.44±0.05↑ & 18.37±0.04→ & 18.23±0.02↓ & 17.88±0.11↓ \\
	\cmidrule{2-7}  & ins3 & 15.29±0.06 & 15.36±0.01↑ & 15.28±0.05→ & 15.07±0.02↓ & 14.63±0.11↓ \\
	\midrule
\multicolumn{3}{c}{W-D-L}  & 2-3-7 & 3-9-0 & 12-0-0 & 12-0-0\\
	\bottomrule
	\end{tabular}}
\end{table*}

\begin{table*}[htbp]
\centering
\caption{\textbf{Comparative Performance of Expert Selection Strategies on the AS Problem} ``↑, ↓, →'' represents that the corresponding strategy is significantly better, worse, or not significantly different than the MEGO, respectively.}
\label{tab: strategies_result_as}
	{\begin{tabular}{ccccccc}
	\toprule
	\multicolumn{2}{c}{Problem Instance} & MEGO & MAX & SAMPLE & MEAN & MIN \\
	\midrule
	\multirow{3}[4]{*}{Dim=40} & ins1 & 30853.00±574.56 & 31482.30±444.39↑ & 30919.37±369.48→ & 29155.88±118.00↓ & 25392.37±1172.58↓ \\
	\cmidrule{2-7}  & ins2 & 9607.90±226.70 & 10318.67±404.00↑ & 9700.72±100.88↑ & 9241.11±47.31↓ & 8074.47±350.89↓ \\
	\cmidrule{2-7}  & ins3 & 20980.47±649.65 & 22438.57±411.74↑ & 20999.54±415.19→ & 20223.49±132.82↓ & 17039.57±763.12↓ \\
	\midrule
	\multirow{3}[4]{*}{Dim=60} & ins1 & 41377.20±566.66 & 40461.13±548.90↓ & 41299.70±548.30→ & 37833.08±249.60↓ & 32013.80±777.93↓ \\
	\cmidrule{2-7}  & ins2 & 15869.67±334.65 & 15874.80±33.71↑ & 15849.54±225.94→ & 14993.89±82.22↓ & 12910.67±370.24↓ \\
	\cmidrule{2-7}  & ins3 & 21906.00±474.12 & 22123.67±302.30↑ & 21627.20±231.31↓ & 20319.40±119.37↓ & 17652.27±512.44↓ \\
	\midrule
	\multirow{3}[4]{*}{Dim=80} & ins1 & 37958.10±627.32 & 39200.50±473.13↑ & 37984.36±375.24→ & 36287.61±155.85↓ & 32447.93±895.56↓ \\
	\cmidrule{2-7}  & ins2 & 42017.37±335.69 & 41976.50±207.68→ & 41961.31±281.80→ & 40249.77±168.64↓ & 35841.70±2290.94↓ \\
	\cmidrule{2-7}  & ins3 & 34970.57±384.71 & 35889.37±356.42↑ & 34903.06±334.73→ & 33319.74±158.18↓ & 30138.20±649.21↓ \\
	\midrule
	\multirow{3}[4]{*}{Dim=100} & ins1 & 49924.53±189.62 & 50589.23±219.52↑ & 49927.39±170.94→ & 49066.97±109.86↓ & 45824.23±1449.81↓ \\
	\cmidrule{2-7}  & ins2 & 34598.37±628.39 & 38340.97±728.83↑ & 35012.55±394.32↑ & 35057.01±204.88↑ & 31694.27±609.80↓ \\
	\cmidrule{2-7}  & ins3 & 38270.53±349.64 & 39739.87±609.75↑ & 38240.96±254.45→ & 37171.64±117.46↓ & 34206.00±531.95↓ \\
	\midrule
\multicolumn{3}{c}{W-D-L}  & 1-1-10 & 1-9-2 & 11-0-1 & 12-0-0\\
	\bottomrule
	\end{tabular}}
\end{table*}

\begin{table*}[htbp]
\centering
\caption{\textbf{Comparative Performance of Expert Selection Strategies on the CIM Problem} ``↑, ↓, →'' represents that the corresponding strategy is significantly better, worse, or not significantly different than the MEGO, respectively.}
\label{tab: strategies_result_cim}
	{\begin{tabular}{ccccccc}
	\toprule
	\multicolumn{2}{c}{Problem Instance} & MEGO & MAX & SAMPLE & MEAN & MIN \\
	\midrule
	\multirow{3}[4]{*}{Dim=40} & ins1 & 32.97±0.33 & 32.97±0.27→ & 32.88±0.26→ & 30.96±0.20↓ & 25.07±1.69↓ \\
	\cmidrule{2-7}  & ins2 & 29.99±0.10 & 29.90±0.08↓ & 30.00±0.08→ & 28.76±0.16↓ & 23.21±2.40↓ \\
	\cmidrule{2-7}  & ins3 & 35.18±0.45 & 35.37±0.34→ & 35.07±0.32→ & 33.46±0.14↓ & 31.30±0.66↓ \\
	\midrule
	\multirow{3}[4]{*}{Dim=60} & ins1 & 56.02±0.67 & 56.73±0.81↑ & 55.79±0.40↓ & 53.84±0.28↓ & 49.80±2.18↓ \\
	\cmidrule{2-7}  & ins2 & 41.62±0.76 & 44.00±1.21↑ & 42.28±0.37↑ & 40.34±0.26↓ & 37.34±0.95↓ \\
	\cmidrule{2-7}  & ins3 & 49.15±0.16 & 48.67±0.37↓ & 49.16±0.11→ & 46.48±0.17↓ & 40.24±2.59↓ \\
	\midrule
	\multirow{3}[4]{*}{Dim=80} & ins1 & 83.31±0.94 & 84.01±0.67↑ & 83.16±0.78→ & 79.92±0.57↓ & 66.56±3.93↓ \\
	\cmidrule{2-7}  & ins2 & 36.67±1.21 & 38.92±0.72↑ & 36.46±0.32↑ & 35.51±0.26↓ & 31.76±0.86↓ \\
	\cmidrule{2-7}  & ins3 & 35.29±0.39 & 36.50±0.44↑ & 35.25±0.23→ & 34.85±0.13↓ & 33.36±0.36↓ \\
	\midrule
	\multirow{3}[4]{*}{Dim=100} & ins1 & 43.99±0.53 & 44.18±0.46→ & 43.95±0.47→ & 41.63±0.28↓ & 37.20±1.34↓ \\
	\cmidrule{2-7}  & ins2 & 61.85±0.25 & 61.69±0.46↓ & 61.79±0.21→ & 59.81±0.17↓ & 56.24±2.70↓ \\
	\cmidrule{2-7}  & ins3 & 103.57±0.90 & 106.40±0.79↑ & 103.39±0.84→ & 101.09±0.42↓ & 93.91±2.42↓ \\
	\midrule
\multicolumn{3}{c}{W-D-L}  & 3-3-6 & 1-9-2 & 12-0-0 & 12-0-0\\
	\bottomrule
	\end{tabular}}
\end{table*}

\begin{table*}[htbp]
\centering
\caption{\textbf{Comparative Performance of Expert Selection Strategies on the CA Problem} ``↑, ↓, →'' represents that the corresponding strategy is significantly better, worse, or not significantly different than the MEGO, respectively.}
\label{tab: strategies_result_ca}
	{\begin{tabular}{ccccccc}
	\toprule
	\multicolumn{2}{c}{Problem Instance} & MEGO & MAX & SAMPLE & MEAN & MIN \\
	\midrule
	\multirow{3}[4]{*}{Dim=40} & ins1 & -5582.13±12.34 & -5576.00±0.00→ & -5576.62±0.96↓ & -5667.26±6.92↓ & -6264.00±0.00↓ \\
	\cmidrule{2-7}  & ins2 & -6848.00±0.00 & -6848.00±0.00→ & -6850.69±1.17↓ & -6996.09±19.69↓ & -7453.87±46.05↓ \\
	\cmidrule{2-7}  & ins3 & -5672.00±0.00 & -5672.00±0.00→ & -5672.10±0.27↓ & -6016.24±48.30↓ & -7496.27±682.51↓ \\
	\midrule
	\multirow{3}[4]{*}{Dim=60} & ins1 & -6367.47±2.00 & -6367.73±1.44→ & -6371.67±4.19↓ & -6428.80±8.51↓ & -6621.07±149.97↓ \\
	\cmidrule{2-7}  & ins2 & -6264.53±2.87 & -6264.00±0.00→ & -6264.19±0.50↓ & -6355.87±26.70↓ & -6897.87±377.92↓ \\
	\cmidrule{2-7}  & ins3 & -9587.20±9.14 & -9584.80±7.55→ & -9585.47±3.50→ & -9723.64±66.56↓ & -11784.00±1523.08↓ \\
	\midrule
	\multirow{3}[4]{*}{Dim=80} & ins1 & -9210.40±47.61 & -9149.87±8.75↑ & -9189.60±10.47→ & -9265.23±26.95↓ & -9712.00±665.96↓ \\
	\cmidrule{2-7}  & ins2 & -6186.93±27.28 & -6155.73±17.74↑ & -6171.74±6.83↑ & -6279.74±16.90↓ & -6581.33±271.25↓ \\
	\cmidrule{2-7}  & ins3 & -5252.27±17.00 & -5261.87±16.12↓ & -5243.46±6.10→ & -5600.49±58.87↓ & -6689.87±458.82↓ \\
	\midrule
	\multirow{3}[4]{*}{Dim=100} & ins1 & -61025.33±167.94 & -60726.13±67.37↑ & -61114.60±92.51↓ & -61753.36±93.29↓ & -64821.87±754.64↓ \\
	\cmidrule{2-7}  & ins2 & -4297.60±13.29 & -4295.47±17.27→ & -4318.57±14.43↓ & -4466.67±24.41↓ & -5090.13±357.45↓ \\
	\cmidrule{2-7}  & ins3 & -5565.33±18.16 & -5534.13±9.39↑ & -5569.89±10.40→ & -5712.57±28.20↓ & -6585.87±283.88↓ \\
	\midrule
\multicolumn{3}{c}{W-D-L}  & 1-7-4 & 7-4-1 & 12-0-0 & 12-0-0\\
	\bottomrule
	\end{tabular}}
\end{table*}

	\subsection{Performance Comparison in Source Space vs. Latent Space}
	
	We compare the performance of our proposed method when searching in two distinct spaces: the source problem solution space and a learned latent space. The two compared strategies are defined as follows:
	\begin{itemize}
		\item Source Space: The MEGO method is applied to search directly in the source problem solution space, where 2,000,000 solutions are randomly sampled and evaluated.
		\item Latent Space: The search is performed directly in the latent space of a pretrained VAE. Initially, 1,024 random latent points are sampled and decoded into solutions. The objective function is defined as min \(-y^\prime\), and gradient descent is run for 1,954 steps, resulting in a total of 2,000,896 solution evaluations.
	\end{itemize}
	
	All comparative results are summarized in Table~\ref{tab: search_space_result_om}-\ref{tab: search_space_result_ca}. Each strategy was run independently 30 times per problem instance. The table reports the mean and standard deviation of the obtained performance. To assess statistical significance, the Wilcoxon rank-sum test (with a significance level of 0.05) was conducted between the results of the two strategies for each instance, and the outcomes are annotated accordingly in the table.

\begin{table}[htbp]
\centering
\caption{\textbf{Comparative Performance of Different Search Space for MEGO on the OM Problem} ``↑, ↓, →'' represents that the Latent Space is significantly better, worse, or not significantly different than the Source Space, respectively.}
\label{tab: search_space_result_om}
	{\begin{tabular}{cccc}
	\toprule
	\multicolumn{2}{c}{Problem Instance} & Source Space & Latent Space \\
	\midrule
	\multirow{3}[4]{*}{Dim=40} & ins1 & 30.93±1.00 & 32.63±0.71↑ \\
	\cmidrule{2-4}  & ins2 & 31.53±1.02 & 30.87±1.09↓ \\
	\cmidrule{2-4}  & ins3 & 31.40±1.31 & 30.30±1.16↓ \\
	\midrule
	\multirow{3}[4]{*}{Dim=60} & ins1 & 44.43±1.31 & 40.63±0.87↓ \\
	\cmidrule{2-4}  & ins2 & 42.53±1.38 & 39.27±1.15↓ \\
	\cmidrule{2-4}  & ins3 & 43.53±1.31 & 40.10±1.27↓ \\
	\midrule
	\multirow{3}[4]{*}{Dim=80} & ins1 & 54.10±2.61 & 50.53±2.32↓ \\
	\cmidrule{2-4}  & ins2 & 55.00±1.51 & 54.23±1.54→ \\
	\cmidrule{2-4}  & ins3 & 56.67±1.89 & 52.67±1.49↓ \\
	\midrule
	\multirow{3}[4]{*}{Dim=100} & ins1 & 67.80±3.26 & 61.40±2.36↓ \\
	\cmidrule{2-4}  & ins2 & 66.03±2.30 & 62.77±0.67↓ \\
	\cmidrule{2-4}  & ins3 & 64.33±1.37 & 62.13±1.91↓ \\
	\midrule
\multicolumn{3}{c}{W-D-L}  & 10-1-1\\
	\bottomrule
	\end{tabular}}
\end{table}

\begin{table}[htbp]
\centering
\caption{\textbf{Comparative Performance of Different Search Space for MEGO on the MC Problem} ``↑, ↓, →'' represents that the Latent Space is significantly better, worse, or not significantly different than the Source Space, respectively.}
\label{tab: search_space_result_mc}
	{\begin{tabular}{cccc}
	\toprule
	\multicolumn{2}{c}{Problem Instance} & Source Space & Latent Space \\
	\midrule
	\multirow{3}[4]{*}{Dim=40} & ins1 & 138.20±1.08 & 138.47±1.20→ \\
	\cmidrule{2-4}  & ins2 & 186.00±3.80 & 188.00±4.27→ \\
	\cmidrule{2-4}  & ins3 & 264.60±2.14 & 264.90±2.41→ \\
	\midrule
	\multirow{3}[4]{*}{Dim=60} & ins1 & 464.97±2.44 & 461.17±4.10↓ \\
	\cmidrule{2-4}  & ins2 & 441.30±1.81 & 442.43±3.38→ \\
	\cmidrule{2-4}  & ins3 & 622.87±2.51 & 622.20±3.12↓ \\
	\midrule
	\multirow{3}[4]{*}{Dim=80} & ins1 & 1067.97±3.90 & 1068.20±4.35→ \\
	\cmidrule{2-4}  & ins2 & 710.57±7.29 & 704.27±6.07↓ \\
	\cmidrule{2-4}  & ins3 & 1045.20±5.01 & 1041.30±6.02↓ \\
	\midrule
	\multirow{3}[4]{*}{Dim=100} & ins1 & 1379.70±5.97 & 1371.40±8.01↓ \\
	\cmidrule{2-4}  & ins2 & 1108.73±5.15 & 1107.37±7.09→ \\
	\cmidrule{2-4}  & ins3 & 1424.13±6.58 & 1421.87±5.38→ \\
	\midrule
\multicolumn{3}{c}{W-D-L}  & 5-7-0\\
	\bottomrule
	\end{tabular}}
\end{table}

\begin{table}[htbp]
\centering
\caption{\textbf{Comparative Performance of Different Search Space for MEGO on the KP Problem} ``↑, ↓, →'' represents that the Latent Space is significantly better, worse, or not significantly different than the Source Space, respectively.}
\label{tab: search_space_result_kp}
	{\begin{tabular}{cccc}
	\toprule
	\multicolumn{2}{c}{Problem Instance} & Source Space & Latent Space \\
	\midrule
	\multirow{3}[4]{*}{Dim=40} & ins1 & 13.11±0.08 & 12.92±0.16↓ \\
	\cmidrule{2-4}  & ins2 & 7.48±0.10 & 7.42±0.16↓ \\
	\cmidrule{2-4}  & ins3 & 5.90±0.05 & 5.73±0.10↓ \\
	\midrule
	\multirow{3}[4]{*}{Dim=60} & ins1 & 7.39±0.08 & 7.34±0.04↓ \\
	\cmidrule{2-4}  & ins2 & 12.96±0.06 & 12.90±0.07↓ \\
	\cmidrule{2-4}  & ins3 & 8.71±0.09 & 8.68±0.07↓ \\
	\midrule
	\multirow{3}[4]{*}{Dim=80} & ins1 & 25.90±0.94 & 23.64±1.13↓ \\
	\cmidrule{2-4}  & ins2 & 19.18±0.07 & 19.09±0.08↓ \\
	\cmidrule{2-4}  & ins3 & 23.16±0.11 & 23.07±0.14↓ \\
	\midrule
	\multirow{3}[4]{*}{Dim=100} & ins1 & 36.42±0.88 & 36.23±1.14→ \\
	\cmidrule{2-4}  & ins2 & 18.36±0.03 & 18.35±0.06→ \\
	\cmidrule{2-4}  & ins3 & 15.27±0.04 & 15.19±0.06↓ \\
	\midrule
\multicolumn{3}{c}{W-D-L}  & 10-2-0\\
	\bottomrule
	\end{tabular}}
\end{table}

\begin{table}[htbp]
\centering
\caption{\textbf{Comparative Performance of Different Search Space for MEGO on the AS Problem} ``↑, ↓, →'' represents that the Latent Space is significantly better, worse, or not significantly different than the Source Space, respectively.}
\label{tab: search_space_result_as}
	{\begin{tabular}{cccc}
	\toprule
	\multicolumn{2}{c}{Problem Instance} & Source Space & Latent Space \\
	\midrule
	\multirow{3}[4]{*}{Dim=40} & ins1 & 30992.93±532.20 & 30458.60±397.02↓ \\
	\cmidrule{2-4}  & ins2 & 9745.73±220.71 & 9446.10±278.16↓ \\
	\cmidrule{2-4}  & ins3 & 21265.60±850.30 & 19130.20±584.08↓ \\
	\midrule
	\multirow{3}[4]{*}{Dim=60} & ins1 & 40968.03±432.25 & 38448.90±993.37↓ \\
	\cmidrule{2-4}  & ins2 & 15914.53±293.09 & 15713.23±448.60↓ \\
	\cmidrule{2-4}  & ins3 & 21832.03±465.00 & 21010.53±127.74↓ \\
	\midrule
	\multirow{3}[4]{*}{Dim=80} & ins1 & 38340.23±698.99 & 37004.30±1320.68↓ \\
	\cmidrule{2-4}  & ins2 & 41911.77±299.68 & 41540.17±432.21↓ \\
	\cmidrule{2-4}  & ins3 & 34882.03±435.98 & 33295.30±751.07↓ \\
	\midrule
	\multirow{3}[4]{*}{Dim=100} & ins1 & 49929.23±227.46 & 49405.23±177.34↓ \\
	\cmidrule{2-4}  & ins2 & 34673.20±834.20 & 34551.93±505.71→ \\
	\cmidrule{2-4}  & ins3 & 38359.80±485.44 & 39045.13±704.90↑ \\
	\midrule
\multicolumn{3}{c}{W-D-L}  & 10-1-1\\
	\bottomrule
	\end{tabular}}
\end{table}

\begin{table}[htbp]
\centering
\caption{\textbf{Comparative Performance of Different Search Space for MEGO on the CIM Problem} ``↑, ↓, →'' represents that the Latent Space is significantly better, worse, or not significantly different than the Source Space, respectively.}
\label{tab: search_space_result_cim}
	{\begin{tabular}{cccc}
	\toprule
	\multicolumn{2}{c}{Problem Instance} & Source Space & Latent Space \\
	\midrule
	\multirow{3}[4]{*}{Dim=40} & ins1 & 32.97±0.36 & 32.29±0.36↓ \\
	\cmidrule{2-4}  & ins2 & 29.96±0.09 & 29.30±0.35↓ \\
	\cmidrule{2-4}  & ins3 & 34.85±0.32 & 34.23±0.30↓ \\
	\midrule
	\multirow{3}[4]{*}{Dim=60} & ins1 & 55.83±0.71 & 54.79±0.82↓ \\
	\cmidrule{2-4}  & ins2 & 41.62±0.76 & 42.45±1.36↑ \\
	\cmidrule{2-4}  & ins3 & 49.10±0.12 & 48.83±0.40↓ \\
	\midrule
	\multirow{3}[4]{*}{Dim=80} & ins1 & 83.27±1.01 & 80.90±1.81↓ \\
	\cmidrule{2-4}  & ins2 & 36.04±0.38 & 35.58±1.06→ \\
	\cmidrule{2-4}  & ins3 & 35.20±0.36 & 34.78±0.66↓ \\
	\midrule
	\multirow{3}[4]{*}{Dim=100} & ins1 & 44.36±0.82 & 42.94±0.73↓ \\
	\cmidrule{2-4}  & ins2 & 61.92±0.32 & 61.25±0.47↓ \\
	\cmidrule{2-4}  & ins3 & 103.35±0.85 & 100.69±1.13↓ \\
	\midrule
\multicolumn{3}{c}{W-D-L}  & 10-1-1\\
	\bottomrule
	\end{tabular}}
\end{table}

\begin{table}[htbp]
\centering
\caption{\textbf{Comparative Performance of Different Search Space for MEGO on the CA Problem} ``↑, ↓, →'' represents that the Latent Space is significantly better, worse, or not significantly different than the Source Space, respectively.}
\label{tab: search_space_result_ca}
	{\begin{tabular}{cccc}
	\toprule
	\multicolumn{2}{c}{Problem Instance} & Source Space & Latent Space \\
	\midrule
	\multirow{3}[4]{*}{Dim=40} & ins1 & -5575.73±1.44 & -5597.60±18.49↓ \\
	\cmidrule{2-4}  & ins2 & -6848.00±0.00 & -6849.87±4.92→ \\
	\cmidrule{2-4}  & ins3 & -5672.00±0.00 & -5675.20±15.81→ \\
	\midrule
	\multirow{3}[4]{*}{Dim=60} & ins1 & -6368.80±4.31 & -6397.33±10.80↓ \\
	\cmidrule{2-4}  & ins2 & -6265.07±3.99 & -6264.00±0.00→ \\
	\cmidrule{2-4}  & ins3 & -9580.53±7.36 & -9608.27±18.87↓ \\
	\midrule
	\multirow{3}[4]{*}{Dim=80} & ins1 & -9165.33±26.64 & -9177.33±22.16↓ \\
	\cmidrule{2-4}  & ins2 & -6204.27±14.57 & -6265.60±53.76↓ \\
	\cmidrule{2-4}  & ins3 & -5242.93±14.23 & -5298.13±32.46↓ \\
	\midrule
	\multirow{3}[4]{*}{Dim=100} & ins1 & -61057.87±101.17 & -61284.00±96.57↓ \\
	\cmidrule{2-4}  & ins2 & -4355.20±41.39 & -4462.40±122.56↓ \\
	\cmidrule{2-4}  & ins3 & -5589.33±11.75 & -5582.93±35.04→ \\
	\midrule
\multicolumn{3}{c}{W-D-L}  & 8-4-0\\
	\bottomrule
	\end{tabular}}
\end{table}
	
	\subsection{Ablation Study on the Solution Mapping Process}
	
	We compare the performance of the MEGO framework under two configurations to assess the contribution of its core solution mapping process. The two compared settings are defined as follows:
	\begin{itemize}
		\item \textbf{Mapping}: The standard MEGO procedure.
		\item \textbf{No Mapping}: An ablated version where the fine-tuning and solution mapping steps are omitted. After experts are selected, new solutions are sampled directly in the source problem's solution space. These solutions are then truncated or zero-padded to match the target problem's dimension. The top-\(k\) solutions are selected, which are ranked by their performance on the expert models, are finally evaluated on the actual target problem instance.
	\end{itemize}
	All comparative results are presented in Table~\ref{tab: mapping_result_om}-\ref{tab: mapping_result_ca}. Each configuration was executed for 30 independent runs on every problem instance. The table reports the mean and standard deviation of the performance. To determine statistical significance, the Wilcoxon rank-sum test (at a 0.05 significance level) was conducted between the results of the two configurations for each instance, and the outcomes are indicated within the table.

\begin{table}[htbp]
\centering
\caption{\textbf{Comparative Performance of Mapping of Not in MEGO on the MC Problem} ``↑, ↓, →'' represents that the No Mapping Result is significantly better, worse, or not significantly different than the Mapping Result, respectively.}
\label{tab: mapping_result_mc}
	{\begin{tabular}{cccc}
	\toprule
	\multicolumn{2}{c}{Problem Instance} & Mapping & No Mapping \\
	\midrule
	\multirow{3}[4]{*}{Dim=40} & ins1 & 138.20±1.08 & 135.77±1.45↓ \\
	\cmidrule{2-4}  & ins2 & 186.00±3.80 & 178.60±2.06↓ \\
	\cmidrule{2-4}  & ins3 & 264.60±2.14 & 263.87±2.47→ \\
	\midrule
	\multirow{3}[4]{*}{Dim=60} & ins1 & 464.97±2.44 & 458.60±3.00↓ \\
	\cmidrule{2-4}  & ins2 & 441.30±1.81 & 437.57±2.49↓ \\
	\cmidrule{2-4}  & ins3 & 622.87±2.51 & 616.37±2.98↓ \\
	\midrule
	\multirow{3}[4]{*}{Dim=80} & ins1 & 1067.97±3.90 & 1042.50±3.93↓ \\
	\cmidrule{2-4}  & ins2 & 710.57±7.29 & 683.37±5.13↓ \\
	\cmidrule{2-4}  & ins3 & 1045.20±5.01 & 1038.03±5.34↓ \\
	\midrule
	\multirow{3}[4]{*}{Dim=100} & ins1 & 1379.70±5.97 & 1346.53±7.05↓ \\
	\cmidrule{2-4}  & ins2 & 1108.73±5.15 & 1009.57±15.53↓ \\
	\cmidrule{2-4}  & ins3 & 1424.13±6.58 & 1238.70±41.32↓ \\
	\midrule
\multicolumn{3}{c}{W-D-L}  & 11-1-0\\
	\bottomrule
	\end{tabular}}
\end{table}

\begin{table}[htbp]
\centering
\caption{\textbf{Comparative Performance of Mapping of Not in MEGO on the OM Problem} ``↑, ↓, →'' represents that the No Mapping Result is significantly better, worse, or not significantly different than the Mapping Result, respectively.}
\label{tab: mapping_result_om}
	{\begin{tabular}{cccc}
	\toprule
	\multicolumn{2}{c}{Problem Instance} & Mapping & No Mapping \\
	\midrule
	\multirow{3}[4]{*}{Dim=40} & ins1 & 30.93±1.00 & 26.90±1.83↓ \\
	\cmidrule{2-4}  & ins2 & 31.53±1.02 & 29.40±1.14↓ \\
	\cmidrule{2-4}  & ins3 & 31.40±1.31 & 27.90±1.11↓ \\
	\midrule
	\multirow{3}[4]{*}{Dim=60} & ins1 & 44.43±1.31 & 38.30±0.90↓ \\
	\cmidrule{2-4}  & ins2 & 42.53±1.38 & 37.67±1.11↓ \\
	\cmidrule{2-4}  & ins3 & 43.53±1.31 & 37.57±1.02↓ \\
	\midrule
	\multirow{3}[4]{*}{Dim=80} & ins1 & 54.10±2.61 & 43.00±1.10↓ \\
	\cmidrule{2-4}  & ins2 & 55.00±1.51 & 47.40±1.38↓ \\
	\cmidrule{2-4}  & ins3 & 56.67±1.89 & 52.37±1.20↓ \\
	\midrule
	\multirow{3}[4]{*}{Dim=100} & ins1 & 67.80±3.26 & 52.70±1.29↓ \\
	\cmidrule{2-4}  & ins2 & 66.03±2.30 & 62.70±1.51↓ \\
	\cmidrule{2-4}  & ins3 & 64.33±1.37 & 63.10±1.14↓ \\
	\midrule
\multicolumn{3}{c}{W-D-L}  & 12-0-0\\
	\bottomrule
	\end{tabular}}
\end{table}

\begin{table}[htbp]
\centering
\caption{\textbf{Comparative Performance of Mapping of Not in MEGO on the CIM Problem} ``↑, ↓, →'' represents that the No Mapping Result is significantly better, worse, or not significantly different than the Mapping Result, respectively.}
\label{tab: mapping_result_cim}
	{\begin{tabular}{cccc}
	\toprule
	\multicolumn{2}{c}{Problem Instance} & Mapping & No Mapping \\
	\midrule
	\multirow{3}[4]{*}{Dim=40} & ins1 & 32.97±0.36 & 31.15±0.93↓ \\
	\cmidrule{2-4}  & ins2 & 29.96±0.09 & 29.83±0.10↓ \\
	\cmidrule{2-4}  & ins3 & 34.85±0.32 & 33.82±0.52↓ \\
	\midrule
	\multirow{3}[4]{*}{Dim=60} & ins1 & 55.83±0.71 & 50.10±0.95↓ \\
	\cmidrule{2-4}  & ins2 & 41.62±0.76 & 39.08±0.77↓ \\
	\cmidrule{2-4}  & ins3 & 49.10±0.12 & 43.67±0.82↓ \\
	\midrule
	\multirow{3}[4]{*}{Dim=80} & ins1 & 83.27±1.01 & 56.89±1.70↓ \\
	\cmidrule{2-4}  & ins2 & 36.04±0.38 & 27.30±1.30↓ \\
	\cmidrule{2-4}  & ins3 & 35.20±0.36 & 32.85±0.65↓ \\
	\midrule
	\multirow{3}[4]{*}{Dim=100} & ins1 & 44.36±0.82 & 41.46±0.55↓ \\
	\cmidrule{2-4}  & ins2 & 61.92±0.32 & 39.11±0.97↓ \\
	\cmidrule{2-4}  & ins3 & 103.35±0.85 & 65.32±1.62↓ \\
	\midrule
\multicolumn{3}{c}{W-D-L}  & 12-0-0\\
	\bottomrule
	\end{tabular}}
\end{table}

\begin{table}[htbp]
\centering
\caption{\textbf{Comparative Performance of Mapping of Not in MEGO on the CA Problem} ``↑, ↓, →'' represents that the No Mapping Result is significantly better, worse, or not significantly different than the Mapping Result, respectively.}
\label{tab: mapping_result_ca}
	{\begin{tabular}{cccc}
	\toprule
	\multicolumn{2}{c}{Problem Instance} & Mapping & No Mapping \\
	\midrule
	\multirow{3}[4]{*}{Dim=40} & ins1 & -5575.73±1.44 & -5580.27±10.88→ \\
	\cmidrule{2-4}  & ins2 & -6848.00±0.00 & -6854.13±9.39↓ \\
	\cmidrule{2-4}  & ins3 & -5672.00±0.00 & -5672.00±0.00→ \\
	\midrule
	\multirow{3}[4]{*}{Dim=60} & ins1 & -6368.80±4.31 & -6522.13±11.49↓ \\
	\cmidrule{2-4}  & ins2 & -6265.07±3.99 & -6314.13±5.44↓ \\
	\cmidrule{2-4}  & ins3 & -9580.53±7.36 & -9620.53±21.33↓ \\
	\midrule
	\multirow{3}[4]{*}{Dim=80} & ins1 & -9165.33±26.64 & -12887.47±33.05↓ \\
	\cmidrule{2-4}  & ins2 & -6204.27±14.57 & -6179.20±43.31↑ \\
	\cmidrule{2-4}  & ins3 & -5242.93±14.23 & -5528.80±133.14↓ \\
	\midrule
	\multirow{3}[4]{*}{Dim=100} & ins1 & -61057.87±101.17 & -66536.00±0.00↓ \\
	\cmidrule{2-4}  & ins2 & -4355.20±41.39 & -5939.20±19.33↓ \\
	\cmidrule{2-4}  & ins3 & -5589.33±11.75 & -5987.73±6.77↓ \\
	\midrule
\multicolumn{3}{c}{W-D-L}  & 9-2-1\\
	\bottomrule
	\end{tabular}}
\end{table}

\begin{table}[htbp]
\centering
\caption{\textbf{Comparative Performance of Mapping of Not in MEGO on the AS Problem} ``↑, ↓, →'' represents that the No Mapping Result is significantly better, worse, or not significantly different than the Mapping Result, respectively.}
\label{tab: mapping_result_as}
	{\begin{tabular}{cccc}
	\toprule
	\multicolumn{2}{c}{Problem Instance} & Mapping & No Mapping \\
	\midrule
	\multirow{3}[4]{*}{Dim=40} & ins1 & 30992.93±532.20 & 29233.20±575.12↓ \\
	\cmidrule{2-4}  & ins2 & 9745.73±220.71 & 9374.97±407.48↓ \\
	\cmidrule{2-4}  & ins3 & 21265.60±850.30 & 20593.90±686.01↓ \\
	\midrule
	\multirow{3}[4]{*}{Dim=60} & ins1 & 40968.03±432.25 & 32584.50±885.95↓ \\
	\cmidrule{2-4}  & ins2 & 15914.53±293.09 & 15149.37±269.51↓ \\
	\cmidrule{2-4}  & ins3 & 21832.03±465.00 & 19855.93±522.65↓ \\
	\midrule
	\multirow{3}[4]{*}{Dim=80} & ins1 & 38340.23±698.99 & 29163.37±1490.43↓ \\
	\cmidrule{2-4}  & ins2 & 41911.77±299.68 & 33200.30±833.80↓ \\
	\cmidrule{2-4}  & ins3 & 34882.03±435.98 & 30520.87±527.08↓ \\
	\midrule
	\multirow{3}[4]{*}{Dim=100} & ins1 & 49929.23±227.46 & 37319.30±725.90↓ \\
	\cmidrule{2-4}  & ins2 & 34673.20±834.20 & 18924.40±915.72↓ \\
	\cmidrule{2-4}  & ins3 & 38359.80±485.44 & 35121.73±670.46↓ \\
	\midrule
\multicolumn{3}{c}{W-D-L}  & 12-0-0\\
	\bottomrule
	\end{tabular}}
\end{table}

\begin{table}[htbp]
\centering
\caption{\textbf{Comparative Performance of Mapping of Not in MEGO on the KP Problem} ``↑, ↓, →'' represents that the No Mapping Result is significantly better, worse, or not significantly different than the Mapping Result, respectively.}
\label{tab: mapping_result_kp}
	{\begin{tabular}{cccc}
	\toprule
	\multicolumn{2}{c}{Problem Instance} & Mapping & No Mapping \\
	\midrule
	\multirow{3}[4]{*}{Dim=40} & ins1 & 13.11±0.08 & 13.09±0.16→ \\
	\cmidrule{2-4}  & ins2 & 7.48±0.10 & 7.18±0.12↓ \\
	\cmidrule{2-4}  & ins3 & 5.90±0.05 & 5.67±0.07↓ \\
	\midrule
	\multirow{3}[4]{*}{Dim=60} & ins1 & 7.39±0.08 & 7.29±0.06↓ \\
	\cmidrule{2-4}  & ins2 & 12.96±0.06 & 12.75±0.05↓ \\
	\cmidrule{2-4}  & ins3 & 8.71±0.09 & 8.55±0.08↓ \\
	\midrule
	\multirow{3}[4]{*}{Dim=80} & ins1 & 25.90±0.94 & 16.81±0.66↓ \\
	\cmidrule{2-4}  & ins2 & 19.18±0.07 & 17.65±0.57↓ \\
	\cmidrule{2-4}  & ins3 & 23.16±0.11 & 17.71±0.56↓ \\
	\midrule
	\multirow{3}[4]{*}{Dim=100} & ins1 & 36.42±0.88 & 17.82±0.75↓ \\
	\cmidrule{2-4}  & ins2 & 18.36±0.03 & 16.49±0.66↓ \\
	\cmidrule{2-4}  & ins3 & 15.27±0.04 & 15.14±0.08↓ \\
	\midrule
\multicolumn{3}{c}{W-D-L}  & 11-1-0\\
	\bottomrule
	\end{tabular}}
\end{table}

{\color{black}
\section{Further Discussion on Non-Arbitrary Regularities and an Additional Experiment on Arbitrary Random Binary Functions}

\subsection{Non-Arbitrary Regularities Across Problem Classes}

In the main text, we argue that transfer in MEGO is plausible because practical binary optimization problems often exhibit some coarse-grained reusable structure (regularities) in their solution-quality landscapes.
Here we further elaborate on what such structure may look like in the present experiments.

First, variable importance is often highly heterogeneous: some binary decisions are much more influential than others.
In generalized OM, the problem provides a clear bit-level signal, because matching certain positions of the hidden reference vector directly improves the objective value.
In KP, different items have very different value/weight trade-offs, so their usefulness is far from uniform.
In CA, compiler flags also differ greatly in impact: a small number of flags can be highly influential, while many others have weak or even harmful effects.
In CIM and AS, different nodes or anchors likewise have very different marginal utilities.
Thus, good solutions are typically not formed by treating all variables as equally important.

Second, the effect of one binary decision often depends strongly on the others, rather than being simply additive.
In MC, whether selecting a vertex is useful depends on how it relates to the other selected and unselected vertices.
In KP, whether an item is worth selecting depends not only on its own value, but also on the remaining budget after other items have been chosen.
In CA, compiler flags often interact strongly, so the effect of enabling one flag depends on which others are enabled.
In CIM, two seed nodes may either overlap heavily in influence or complement each other.
In AS, two anchors may cover highly overlapping subsets of problems or provide complementary coverage.
Therefore, the quality contribution of a single bit often cannot be understood in isolation.

Third, good solutions often correspond to selecting a limited compatible subset rather than turning on bits indiscriminately.
This is explicit in KP, constrained MC, CIM, and AS, all of which involve size or budget constraints.
Although CA has no explicit cardinality constraint, it empirically also exhibits the pattern that only a relatively small compatible subset of flags is useful, and indiscriminately enabling many flags does not continue to improve performance.
In this sense, good solutions are sparse but coordinated subsets of decisions.

These regularities are more naturally described at the level of coarse structural tendencies in the solution-quality landscape, rather than as explicit recurring 0-1 patterns that can be directly interpreted.
Accordingly, what MEGO appears to exploit is a transferable bias toward certain kinds of structured solution landscapes, such as non-uniform variable importance, interaction-sensitive decisions, and sparse but compatible decision subsets.

\subsection{Additional Experiment on Arbitrary Random Binary Functions}

We additionally conducted an experiment in which the target problem instance is an arbitrary random binary function.
For each binary solution, we convert its 0-1 string into a hash value using the BLAKE2b function, interpret the hash value as an integer random seed, initialize a pseudo-random number generator with this seed, and define the objective value as the first random number generated from it.
Thus, each solution deterministically corresponds to one pseudo-random objective value, while nearby solutions need not bear any meaningful structural relation to each other.
Note this experiment is not intended to model a practical application problem.
Rather, it serves as a boundary-case test in which the target objective lacks reusable structure that can be shared with the available expert pool.

We compare MEGO with GA, HC, BO, and an additional RANDOM baseline.
The RANDOM baseline samples candidate solutions uniformly at random.
For a fair comparison, all methods are run with the same \#FEs as MEGO.
Each method is repeated for 30 independent runs, and the mean and standard deviation of the best objective values are reported.

\begin{table*}[tbp]
\centering
\caption{\color{black}Performance comparison on an arbitrary random binary function. All methods are repeated for 30 independent runs. Mean and STD report the best objective values obtained over these runs, where larger values are better. The last column reports the Wilcoxon rank-sum test against MEGO at the 0.05 significance level.}
\label{tab:random_binary_function_result}
\begin{tabular}{lccc}
\toprule
Method & Mean & STD & Wilcoxon Sign-Rank Test (vs. MEGO) \\
\midrule
MEGO   & 0.988660 & 0.005585 & -- \\
GA     & 0.991193 & 0.007097 & no significant difference \\
HC     & 0.991391 & 0.009887 & no significant difference \\
BO     & 0.981913 & 0.015662 & no significant difference \\
RANDOM & 0.992004 & 0.007142 & MEGO is worse \\
\bottomrule
\end{tabular}
\end{table*}

As shown in Table~\ref{tab:random_binary_function_result}, MEGO is not significantly better than GA, HC, or BO in this experiment, and is even significantly worse than RANDOM.
This result is consistent with the interpretation in the main manuscript: when reusable structure is absent, transfer in MEGO is no longer meaningful and negative transfer may occur.
}

\ifdefined\supplementaryinput
\let\next 
\else
\let\next\relax
\fi
\next

\end{document}